\documentclass[letterpaper]{article} % DO NOT CHANGE THIS
\usepackage{aaai24}  % DO NOT CHANGE THIS
\usepackage{times}  % DO NOT CHANGE THIS
\usepackage{helvet}  % DO NOT CHANGE THIS
\usepackage{courier}  % DO NOT CHANGE THIS
\usepackage[hyphens]{url}  % DO NOT CHANGE THIS
\usepackage{graphicx} % DO NOT CHANGE THIS
\usepackage{natbib}  % DO NOT CHANGE THIS AND DO NOT ADD ANY OPTIONS TO IT
\usepackage{caption} % DO NOT CHANGE THIS AND DO NOT ADD ANY OPTIONS TO IT
\usepackage{algorithm}
\usepackage{algorithmic}

\usepackage{amsmath}
\usepackage{amssymb}
\usepackage{mathtools}
\usepackage{amsthm}
\usepackage{subfigure}
\usepackage{multirow}
\usepackage[figuresright]{rotating}
\usepackage{amsfonts}	
\usepackage{bm}
\usepackage{enumitem}
\usepackage{bbold}
\usepackage{array}
\newcolumntype{x}[1]{>{\centering\arraybackslash}p{#1}}

\usepackage[export]{adjustbox}
\usepackage[capitalize,noabbrev]{cleveref}
\usepackage{booktabs}
\usepackage{makecell}

\newtheorem{theorem}{Theorem}[section]

\usepackage{newfloat}
\usepackage{listings}
\DeclareCaptionStyle{ruled}{labelfont=normalfont,labelsep=colon,strut=off} % DO NOT CHANGE THIS
\floatstyle{ruled}
\newfloat{listing}{tb}{lst}{}
\floatname{listing}{Listing}
\title{Levenshtein Distance Embedding \\with Poisson Regression for DNA Storage}
\author{
    Xiang Wei\equalcontrib\textsuperscript{\rm 1},
	Alan J.X. Guo\equalcontrib\thanks{Corresponding Author.}\textsuperscript{\rm 1},
	Sihan Sun\textsuperscript{\rm 1},
	Mengyi Wei\textsuperscript{\rm 1},
    Wei Yu\textsuperscript{\rm 2}
}
\affiliations{
    \textsuperscript{\rm 1} Center for Applied Mathematics, Tianjin University,\\
	No. 92, Weijin Road, Tianjin, 300072, China\\
	\textsuperscript{\rm 2} China Mobile Research Institute, \\
    No. 32, Xuanwumen West Street, Beijing, 100053, China\\
    \{weixiang, jiaxiang.guo, sihansun, mengyi.wei\}@tju.edu.cn,\\
	YuweiNKU@outlook.com\\
}

\usepackage{bibentry}
\begin{document}

\maketitle

\begin{abstract}
	Efficient computation or approximation of Levenshtein distance, 
	a widely-used metric for evaluating sequence similarity, 
	has attracted significant attention with the emergence of DNA storage and other biological applications. 
	Sequence embedding, which maps Levenshtein distance to a conventional distance between embedding vectors, 
	has emerged as a promising solution. 
	In this paper, a novel neural network-based sequence embedding technique using Poisson regression is proposed. 
	We first provide a theoretical analysis of the impact of embedding dimension on model performance 
	and present a criterion for selecting an appropriate embedding dimension. 
	Under this embedding dimension, the Poisson regression is introduced by assuming the Levenshtein distance 
	between sequences of fixed length following a Poisson distribution, 
	which naturally aligns with the definition of Levenshtein distance. 
	Moreover, from the perspective of the distribution of embedding distances, 
	Poisson regression approximates the negative log likelihood 
	of the chi-squared distribution and offers advancements in removing the skewness. 
	Through comprehensive experiments on real DNA storage data, 
	we demonstrate the superior performance of the proposed method compared to state-of-the-art approaches. 
\end{abstract}

\section{Introduction}\label{sec:introduction}
    The Levenshtein distance~\cite{levenshtein1966binary} (also known as the edit distance) 
    between two sequences is defined as the minimum number of insertions, deletions, or substitutions 
    required to modify one sequence into another. 
    The dynamic programming algorithm introduced in~\cite{wagner1974string} is commonly 
    employed for accurate calculation of the Levenshtein distance. 
    However, this method has a computational complexity of $O(mn)$ for two strings of length $m$ and $n$.
    According to \cref{thm:complexity} from~\cite{backurs2015edit}, 
    \begin{theorem}\label{thm:complexity}
    Given two sequences of length $n$, 
    the Levenshtein distance can't be computed in time $O(n^{2-\delta}), \forall \: \delta >0$, 
    otherwise the Strong Exponential Time Hypothesis would be violated.
    \end{theorem}
    \noindent a method of calculating Levenshtein distance in linear complexity is not reachable. 
    The computational complexity of the Levenshtein distance poses limitations on its application at a large scale, 
    particularly in the field of DNA storage~\cite{rashtchian2017clustering, dong2020dna}. 
    
    The applications of the Levenshtein distance encompass a wide range of domains, 
    including optical character recognition~\cite{haldar2011levenshtein}, 
    text plagiarism detection~\cite{su2008plagiarism}, 
    entity linking~\cite{jiang2014string}, \emph{etc.}
    In the field of bioinformatics, where nucleic acids and proteins are represented as sequences of their basic building blocks, 
    the Levenshtein distance finds widespread applications.
    It is utilized in tasks such as
    multiple sequence alignment~\cite{li2010survey}, biological database retrieval~\cite{berger2020levenshtein}, 
    sequence hierarchical clustering~\cite{sberro2019large}, \emph{etc.} 
    In recent years, the rapid development of DNA storage~\cite{goldman2013towards,church2012next,grass2015robust}
    has introduced numerous applications of the Levenshtein distance, 
    including sequence clustering~\cite{rashtchian2017clustering, zorita2015starcode, qu2022clover, logan20223gold}, 
    sequence alignment~\cite{li2010survey, corso2021neural}, 
    synchronization channel coding~\cite{press2020hedges,bar2023size,welzel2023dna}, \emph{etc.} 
    However, as the scale of information stored by DNA molecules continues to grow, 
    the computational complexity of the Levenshtein distance becomes a significant challenge for the aforementioned applications. 
    
    To address the challenge of high computational complexity, 
    various methods have been proposed to approximate the Levenshtein distance. 
    For example, in~\cite{ostrovsky2007low}, 
    the authors utilized $\ell_1$ distance to approximate the Levenshtein distance of ${0,1}$-words with low distortion. 
    Another approach, the CGK algorithm~\cite{chakraborty2016streaming}, 
    maps the Levenshtein distance to Hamming distance through a randomized injective embedding. 
    Both of these methods use low-complexity distances as approximations for the Levenshtein distance. 
    
    Neural network-based methods have also been explored. 
    The gated recurrent unit (GRU)~\cite{cho2014learning} has been adopted to embed the Levenshtein distance into 
    Euclidean space~\cite{zhang2019neural}, 
    preserving the relative order between the sequences. 
    In~\cite{dai2020convolutional}, a CNN-based embedding pipeline for the Levenshtein distance was proposed. 
    The authors provided theoretical evidence for employing a convolutional neural network (CNN) 
    as the embedding model. 
    In~\cite{corso2021neural}, a framework for embedding biological sequences in geometric vector spaces was proposed.
    In their work, the DNA sequences are embedded into hyperbolic space instead of Euclidean space
    to capture the implicit hierarchical structure from biological relations between the sequences~\cite{chami2020trees}.
    Additionally, squared Euclidean distance has also been used to embed the Levenshtein distance \cite{guo2022deep}. 
    This work established connections between the Levenshtein distance and degree of freedom with the output distribution 
    of the embedding model, 
    and proposed the so-called chi-square regression. 
    
    In this paper, a neural network-based Levenshtein distance embedding algorithm is proposed with the Poisson regression. 
    We first demonstrate that the choice of embedding dimension has a theoretical impact on the distribution of the approximation, 
    which in turn affects the approximation precision. 
    Leveraging these theoretical analyses, 
    we introduce a method to determine the appropriate embedding dimension for a given dataset and embedding network. 
    Once the embedding dimension is determined, we employ Poisson regression to train the embedding model. 
    The Poisson regression offers two significant advantages. 
    Firstly, it naturally aligns with the definition of Levenshtein distance. 
    When the Levenshtein distance is small, it can be approximately interpreted as
    counting the event of edit operations from a fixed interval. 
    Secondly, it approximates the negative log likelihood of the chi-squared distribution 
    and shows advances in removing the skewness. 
    Experimental results in real-world scenarios validate the efficacy of the theoretical analysis on the choice of 
    embedding dimension.
    Through experiments conducted on DNA storage data, it is demonstrated that the proposed method outperforms 
    state-of-the-art approaches. 

\section{Framework of the Embedding Method}\label{sec:dsee}
    In this section, we provide a brief overview of the embedding method framework and highlight some of 
    the assumptions that have been previously validated in~\cite{guo2022deep}. 
    
    Most of the sequence embedding methods explictly~\cite{guo2022deep, zheng2019sense} 
    or implicitly~\cite{cho2014learning,dai2020convolutional} adopt the framework of 
    Siamese neural network~\cite{bromley1993signature, he2020momentum}. 
    In this framework, the embedding network $f(\cdot;\theta)$ maps a sequence $\bm{s}$ to its corresponding embedding vector 
    $\bm{u}=f(\bm{s};\theta)$ using parameter $\theta$. 
    The embedding network $f(\cdot;\theta)$ is optimized by training two identical branches in the Siamese neural network. 
    Let $((\bm{s},\bm{t}),d)$ be a sample from the training data, where the $\bm{s}$ and $\bm{t}$ are two sequences, 
    and $d$ represents the groundtruth Levenshtein distance between them. 
    The parameter $\theta$ is trained by minimizing the difference between the groundtruth Levenshtein distance and the 
    approximation, as shown in the following equation:
    \begin{align}
    \hat{\theta} &= \mathop{\arg\min}_{\theta} \mathcal{L}(\bm{d},\hat{\bm{d}};\theta) \nonumber\\
    &= \mathop{\arg\min}_{\theta} 
    \sum\mathcal{L}(d,\Delta(f(\bm{s};\theta),f(\bm{t};\theta))). \label{optimization}
    \end{align}
    Here, the $\mathcal{L}$ is a predefined loss function, and the $\Delta(\bm{u},\bm{v})$ denotes the distance between 
    embedding vectors $\bm{u}$ and $\bm{v}$ in the embedding space.
    During the testing phase, the sequences $\bm{s}$ and $\bm{t}$ are mapped to their respective embedding vectors
    $\bm{u} = f(\bm{s};\hat{\theta})$ and $\bm{v} = f(\bm{t};\hat{\theta})$ with the optimized parameter $\hat{\theta}$. 
    The distance $\Delta(\bm{u},\bm{v})$ between the embedding vectors is then used as an approximation of the Levenshtein distance 
    between the sequences $\bm{s}$ and $\bm{t}$.
    
    In previous work, it has been shown that the squared Euclidean distance, defined as:
    \begin{equation}\label{eq:SED}
    d_{\ell_2^2}(\bm{u},\bm{v})=||\bm{u}-\bm{v}||_2^2 = \sum_{i}(u_i-v_i)^2,
    \end{equation}
    can effectively approximate the Levenshtein distance~\cite{guo2022deep}. 
    Although the squared Euclidean distance is not a true metric, 
    the application of squared Euclidean distance not only offers good approximation precision, 
    but also establishes a connection between the Levenshtein distance 
    and the degree of freedom with the difference of the embedding vectors. 
    Specifically, under certain assumptions, it was derived that for pairs of sequences with a consistent Levenshtein distance $d$, 
    the distribution of the embedding vector $\bm{u}-\bm{v}$ has a degree $d$ of freedom. 
    To align with their work, we briefly describe these assumptions as follows:
    \begin{enumerate}[label=A\arabic*,ref=A\arabic*,leftmargin=*,align=left]
    \item\label{ass:n01} By deploying a batch normalization layer, each element $u_{i}$ of the embedding vector $\bm{u}$ 
    follows the standard normal distribution $N(0,1)$; 
    \item\label{ass:indi} The embedding vector $\bm{u}$ has no information redundancy. 
    In other words, if $i\neq j$, the embedding elements $u_{i}$ and $u_{j}$ 
    of $\bm{u}$ are independent of each other; 
    \item\label{ass:globalindi} Embedding vectors of non-related sequences are independent of each other.
    \end{enumerate}
    These assumptions allow for theoretical analysis and facilitate the connection between the Levenshtein distance 
    and the squared Euclidean distance.
    
\section{Embedding Dimension and Approximation Precision}\label{sec:var}
    \subsection{Why large embedding dimension improve the approximation precision?}\label{subsec:var}
    The average of the groundtruth Levenshtein distance between the independent sequences is a statistic feature of the dataset. 
    A good embedding method should ensure that the expectation of predicted distance between independent embedding vectors 
    matches the average of groundtruth 
    Levenshtein distance on pairs of independent sequences. 
    To achieve this, a scale factor can be utilized to help the model satisfy this requirement.
    Let $\bm{s}$ and $\bm{t}$ be two independent sequences from a fixed dataset, 
    and $\bm{u}=f(\bm{s})$ and $\bm{v}=f(\bm{t})$ be their corresponding embedding vectors of length $n$. 
    The embedding method~\cite{guo2022deep}empirically selects $80$ as the length of embedding vectors, 
    and the corresponding scale factor is set to be $\sqrt{2}/2$. 
    To adapt this framework to arbitrary dimension $n$, we use a scale factor $r(n)$ with respect to 
    the embedding dimension $n$
    to calculate the approximated distance as follows:
    \begin{equation}\label{eqn:dist}
    \hat{d}(\bm{s},\bm{t}) = d_{\ell_2^2}(r(n)\bm{u},r(n)\bm{v})=r^2(n)\sum_{i=1}^{n}(u_i-v_i)^2.
    \end{equation}
    Based on the three assumptions mentioned earlier, where $u_i$ and $v_i$ independently follow $N(0,1)$, 
    it can be inferred that $u_i-v_i$ follows $N(0,2)$, and $\frac{1}{2}(u_i-v_i)^2$ follows $\chi^2(1)$. 
    Therefore, the approximated distance between the independent sequences $\bm{s}$ and $\bm{t}$ follows a chi-squared distribution 
    with $n$ degree of freedom
    \begin{equation}
    \frac{1}{2r^2(n)}\hat{d}(\bm{s},\bm{t}) \sim \chi^2(n).
    \end{equation} 
    The expectation of $\hat{d}(\bm{s},\bm{t})$ can be easily calculated as: 
    \begin{equation}\label{eqn:mean}
    \mathbb{E}[\hat{d}(\bm{s},\bm{t})] = 2nr^2(n).
    \end{equation} 
    Let $M$ be the average Levenshtein distance between independent sequences over the specific dataset. 
    In order to satisfy \cref{eqn:mean}, the scale factor can be set as:
    \begin{equation}\label{eqn:scalefactor}
    r(n) = \sqrt{\frac{M}{2n}}.
    \end{equation}
    % In summary, by using the scale factor $r(n) = \sqrt{{M}/{2n}}$ with respect to the embedding dimension $n$, we can 
    % embed the sequences into arbitrary dimension $n$, 
    % while ensuring that the expected distance between independent embedding vectors aligns with the average Levenshtein distance 
    % between independent sequences in the dataset. 
    Thus ensuring that the expected distance between independent embedding vectors aligns with the 
    average Levenshtein distance between independent sequences in the dataset. 
    
    An embedding method primarily focuses on approximating the Levenshtein distance between correlated sequences. 
    We will now demonstrate how the embedding dimension $n$ affects the distribution of the approximation for such pairs of sequences. 
    Consider two correlated sequences $\bm{s}$ and $\bm{t}$ with a groundtruth Levenshtein distance of $d$. 
    Using a slight abuse of notation, 
    let $\tilde{\bm{u}}=r(n)\bm{u}$ and $\tilde{\bm{v}}=r(n)\bm{v}$ denote the respective scaled embedding vectors. 
    It is assumed that the difference $\tilde{\bm{u}}-\tilde{\bm{v}}$ has a degree of freedom 
    proportional to the groundtruth distance $d$, which can be expressed as follows:
    \begin{equation} \label{eqn:corralted}
    \tilde{\bm{u}}-\tilde{\bm{v}}=\bm{y}\bm{P}=(y_{1},y_{2}...,y_{m},0...,0) \sqrt{\frac{M}{n}}\bm{P}
    \end{equation}
    where the non-zero elements $y_{i}$ of $\bm{y}$ are independently and identically distributed (i.i.d.) random variables 
    following the standard normal distribution $N(0,1)$, 
    the matrix $\bm{P}$ is an orthogonal matrix, 
    and $m$ is called the degree of freedom with $\tilde{\bm{u}}-\tilde{\bm{v}}$ and equals $nd/M$. 
    Note that under assumptions \ref{ass:indi} and \ref{ass:globalindi}, 
    when $d=M$, we have $m$ equal to the embedding dimension $n$. 
    Although $m$ is rarely an integer, \cref{eqn:corralted} is considered acceptable if the model can learn to 
    approximate integers close to $m$. 
    
    Now, let's calculate the squared Euclidean distance between the embedding vectors $\tilde{\bm{u}}$ and $\tilde{\bm{v}}$: 
    \begin{align}
    d_{l_{2}^{2}}(\tilde{\bm{u}},\tilde{\bm{v}})&=(\tilde{\bm{u}}-\tilde{\bm{v}})(\tilde{\bm{u}}-\tilde{\bm{v}})^{T} 
    = \frac{M}{n}y\bm{P}\bm{P}^{T}y^{T}\nonumber\\
    &=\frac{M}{n}yy^{T}
    =\frac{M}{n}\sum_{i=1}^{m}y_{i}^{2}.
    \end{align}
    This formula shows that the distribution of the approximated distance $d_{l_{2}^{2}}(\tilde{\bm{u}},\tilde{\bm{v}})$ 
    for a correlated pair of sequences with Levenshtein distance $d$ 
    follows the chi-squared distribution: 
    \begin{equation}
    \frac{n}{M}d_{l_{2}^{2}}(\tilde{\bm{u}},\tilde{\bm{v}})\sim \chi^{2}(m) = \chi^{2}(\frac{n}{M}d).
    \label{eqn:v-u}
    \end{equation}
    It can be observed that the expected value of the approximated distribution for correlated pairs 
    is equal to their Levenshtein distance,
    which serves as the learning target for the model. 
    In order to simplify the notations, let's introduce a new variable $k=n/M$. 
    Consequently, the \cref{eqn:v-u} is rewrited as 
    \begin{equation}
    k d_{l_{2}^{2}}(\tilde{\bm{u}},\tilde{\bm{v}})\sim \chi^{2}(kd),\quad k=\frac{n}{M}.
    \label{eqn:final}
    \end{equation}
    
    Given a sequence pair $(\bm{s},\bm{t})$ with a groundtruth Levenshtein distance $d$, it is evident that 
    the precision of the approximation is affected by the variance in the distribution. 
    Referring to \cref{eqn:final}, the variance of approximations can be calculated as: 
    \begin{equation}\label{eqn:var}
    \mathrm{Var}[d_{l_{2}^{2}}(\tilde{\bm{u}},\tilde{\bm{v}})]=\frac{2d}{k}=\frac{2dM}{n}.
    \end{equation}
    When an embedding network $f(\cdot,\hat{\theta})$ is well-trained, one can deduce from \cref{eqn:var} 
    that a larger embedding dimension $n$ leads to smaller variance in the approximations,  
    resulting in higher precision. 
    It is worth noting that, based on \cref{eqn:var}, regardless of the choice of $k$ or the embedding dimension $n$, 
    sequences with smaller Levenshtein distance exhibit higher approximation precision. 
    
    \subsection{Deciding the appropriate embedding dimension.} \label{subsec:esd}
    
    The embedding dimension $n$ is a critical hyperparameter. 
    As discussed in \cref{subsec:var}, 
    there exists a relationship between the variance of approximated distances and the embedding dimension, 
    indicating that a larger $n$ leads to a decrease in variance and higher theoretical approximation precision for the learned model.  
    Therefore, the learned model has higher theoretical approximation precision with a larger $n$.
    However, it is important to note that this assertion may not hold universally due to a potential violation 
    of assumption~\ref{ass:indi}. 
    Referring to \cref{eqn:final}, 
    the degree of freedom with $\tilde{\bm{u}}-\tilde{\bm{v}}$ is $nd/M$. 
    Consequently, increasing the embedding dimension $n$ results in a higher degree of freedom with $\tilde{\bm{u}}-\tilde{\bm{v}}$. 
    For a pair of sequences from a specific dataset, the information capacity is limited 
    since these sequences are discrete and possess finite lengths. 
    Hence, in the context of non-generative neural networks, 
    the degree of freedom with the difference between the embedding vectors $\tilde{\bm{u}}-\tilde{\bm{v}}$ does not reach infinity. 
    Additionally, different network architectures possess varying abilities in feature extraction, 
    which can further restrict the expressiveness of the degree of freedom. 
    Therefore, the attainable degree of freedom may be constrained by both the dataset's inherent information limitations 
    and the specific network architecture employed. 
    In summary, we propose the following assumption as a supplementary
    \begin{enumerate}[label=A\arabic*,ref=A\arabic*,leftmargin=*,align=left]
    \setcounter{enumi}{3}
    \item\label{ass:bound} 
    On a fixed dataset and network architecture, 
    there is an upper bound $n_0$ on the embedding dimension $n$.  
    When $n \leq n_0$, it is possible to achieve full freedom ($n$ degrees of freedom) 
    in the embedding $\tilde{\bm{u}}-\tilde{\bm{v}}$, 
    when $n > n_0$, the embedding $\tilde{\bm{u}}-\tilde{\bm{v}}$ can have at most $n_0$ degrees of freedom. 
    \end{enumerate}
    The relationship between the embedding dimension $n$ and the variance of approximations in \cref{eqn:var} suggests 
    that a larger embedding dimension can potentially improve the theoretical approximation precision, 
    while according to the assumption~\ref{ass:bound}, 
    an embedding dimension beyond $n_{0}$ would be unnecessary 
    and may even hinder the method's performance. 
    Therefore, it is worth to determine the appropriate upper bound $n_{0}$ and utilize it as the embedding dimension. 
    
    \subsection{Embedding dimension searching.} 
    To determine the appropriate embedding dimension $n_0$, 
    one could start by considering assumption \ref{ass:bound} and \cref{eqn:corralted}. 
    Given an embedding dimension $n$ and a fully trained embedding network $f(\cdot;\hat{\theta})$, 
    collect a set of mutually non-related sequences $\{\bm{s}_i\}$ 
    along with their corresponding embedding vectors $\{\bm{u}_i\} = f(\bm{s}_i;\hat{\theta})$. 
    According to assumption~\ref{ass:bound}, if $n\leq n_0$, the difference between the embedding vectors 
    can be expressed as a linear combination of $n$ independent standard normal variables, 
    regarding the scale factor. This can be expressed as: 
    \begin{equation}\label{eqn:uyp1}
    \bm{u_i} - \bm{u_j} = (y_1,y_2,\ldots,y_n)\bm{P} \quad(i\neq j),
    \end{equation}
    where $y_{i}$s are i.i.d. and follow the standard normal distribution $N(0,1)$, 
    and the matrix $\bm{P}$ is an orthogonal matrix. 
    On the other hand, when $n>n_0$, the expression becomes: 
    \begin{equation}\label{eqn:uyp2}
    \bm{u_i} - \bm{u_j} = (y_1,y_2,\ldots,y_{n_0},0,\ldots,0)\bm{P} \quad(i\neq j). 
    \end{equation}
    This leads to consider that the spectral decomposition of the covariance matrix 
    $\mathrm{cov}(\bm{u_i} - \bm{u_j},\bm{u_i} - \bm{u_j})$, 
    which is also a key step in naive principal component analysis (PCA).
    When $n\leq n_0$, the eigenvalues of the covariance matrix are around $1$, 
    when $n>n_0$, the sorted eigenvalues exhibit a steep decline, reaching close to zero.
    Therefore, by gradually increasing the embedding dimension and monitoring the decreasing pattern of the sorted eigenvalues, 
    we can determine the appropriate embedding dimension $n_0$.
    We refer to this value of $n_0$ as the early-stopping dimension (ESD), 
    and adopt it as our choice for the suitable embedding dimension. 
    
\section{Poisson Regression}
    
    In previous studies on Levenshtein distance embedding, 
    the mean square error (MSE), the mean absolute error (MAE), or a combination of both losses have often been employed 
    to optimize the model. 
    However, these losses penalize the predicted distance $\hat{d}$ symmetrically with respect to its groundtruth $d$. 
    This disregards the fact that the distribution of the predicted distance $\hat{d}$ exhibits a clear bias when $d$ is small. 
    In~\cite{guo2022deep}, they introduced the so-called chi-squared regression to address this issue.
    The negative log-likelihood loss of chi-squared distribution 
    \begin{equation}\label{eqn:rex}
    \mathrm{RE}\chi^2(\hat{d},d) = \hat{d} - (d-2)\ln{\hat{d}}
    \end{equation}
    is used as their loss function. 
    It can be observed from \cref{eqn:rex} that this loss function is minimized at $\hat{d}=d-2$ 
    rather than $\hat{d}=d$. 
    To overcome the drawbacks of these existing losses, we propose the adoption of Poisson regression 
    from two different perspectives. 
    
    \subsection{Poisson regression: interpreting Levenshtein distance with Poisson distribution.}
    The Poisson distribution represents the distribution of events occurring within a fixed interval in constant mean rate. 
    The probability mass function of Poisson distribution with expectation $\lambda$ is:
    \begin{equation}
    P(k;\lambda) = \frac{\lambda^k \mathrm{e}^{-\lambda}}{k!}, \quad k={0,1,2,\ldots}.
    \end{equation} 
    In the context of Levenshtein distance, which measures the minimum number of deletions, insertions, and substitutions 
    needed to transform one sequence into another, 
    we can approximate the Levenshtein distance as a Poisson-distributed random variable when the distance is 
    significantly smaller than the length of the sequences. 
    This approximation treats the sequence as a fixed interval and the operations as random events occurring within this interval. 
    In Poisson regression, 
    the model parameter $\theta$ is estimated based on the maximum likelihood of Poisson distribution. 
    The optimization target is typically the negative log-likelihood loss with the Poisson distribution (PNLL) of the 
    groundtruth: 
    \begin{equation}\label{eqn:poiNLL1}
    \mathrm{PNLL}(\hat{d},d) = \hat{d} -d\ln{\hat{d}} + \ln{d!}. 
    \end{equation}
    Omitting the terms that that do not contribute to the gradients, 
    the PNLL loss can be rewritten as: 
    \begin{equation}\label{eqn:poiNLL}
    \mathrm{PNLL}(\hat{d},d) \overset{\cdot}{=}  \hat{d} - d\ln{\hat{d}}.
    \end{equation}
    In Poisson regression, the loss defined in \cref{eqn:poiNLL} reaches its minimum value when $\hat{d}=d$, 
    and also derives asymmetric penalization on $\hat{d}$ around its groundtruth value $d$. 
    
    \subsection{Poisson regression: an asymptotic chi-squared regression.}
    Recall from \cref{sec:dsee} that the approximation of Levenshtein distance
    theoretically follows a chi-squared distribution in \cref{eqn:final}, under the discussed assumptions. 
    In contrast, the proposed Poisson regression utilizes a Poisson distribution, which differs from 
    the chi-squared distribution in \cref{eqn:final}. 
    This raises concerns about the potential impact on the performance of the proposed method. 
    However, the following discussion aims to address and alleviate such doubts 
    by providing another explaination on Poisson regression. 
    
    The PNLL in \cref{eqn:poiNLL} can be viewed as an approximation of the negative log-likelihood of distribution in 
    \cref{eqn:final} when $k$ is sufficiently large. 
    Recall that the distribution of approximations follows \cref{eqn:final} with a parameter $k$. 
    Note that the probability density function of $\chi^2(d)$ is
    \begin{equation}
    p(x;d) = \frac{1}{2^{\frac{d}{2}}\Gamma\left(\frac{d}{2}\right)} x^{\frac{d}{2}-1}\mathrm{e}^{-\frac{x}{2}}
    \end{equation}
    where $\Gamma(\cdot)$ is the Gamma function. 
    One step further, the probability density function $p(x;k,d)$ of \cref{eqn:final} is calculated as
    \begin{equation}\label{eqn:kdpdf}
    p(x;k,d) = \frac{k}{2^{\frac{kd}{2}}\Gamma\left(\frac{kd}{2}\right)} (kx)^{\frac{kd}{2}-1}\mathrm{e}^{-\frac{kx}{2}}.
    \end{equation}
    By \cref{eqn:kdpdf}, we can compute the negative log-likelihood of a predicted $\hat{d}$ from the chi-squared distribution 
    in \cref{eqn:final} as 
    \begin{align}
    -\ln{p(\hat{d};k,d)} =& \frac{k\hat{d}}{2}-(\frac{kd}{2}-1)\ln(k\hat{d})\nonumber\\ 
    & -\ln{k}+\frac{kd}{2}\ln{2} + \ln{\Gamma\left(\frac{kd}{2}\right)}. \label{eqn:nll1}
    \end{align}
    By omitting terms that contribute zero to the gradients and rescaling the equation by multiplying ${2}/{k}$, 
    the negative log-likelihood loss is
    \begin{equation}\label{eqn:nll}
    -\ln{p(\hat{d};k,d)} \overset{\cdot}{=} \hat{d} -(d-\frac{2}{k})\ln\hat{d}.
    \end{equation}
    It is evident that when $k\rightarrow +\infty$, the PNLL in \cref{eqn:poiNLL} is the limit of \cref{eqn:nll}. 
    This observation suggests that the PNLL is an approximation of the negative log-likelihood of 
    the chi-squared distribution in \cref{eqn:final}. 
    It is worth noting that as $k$ increases, the \cref{eqn:nll} converges to \cref{eqn:poiNLL}, 
    and the minimum point of the loss in \cref{eqn:nll} converges to the groundtruth $d$. 
    
    As mentioned in \cref{sec:var}, the chosen embedding dimension $n$ (proportional to $k$) 
    is selected to be as large as possible within the limit of $n_0$. 
    Considering the above discussion, 
    the choice of Poisson regression preserves the advantage of chi-squared regression in terms of asymmetric penalization 
    and addresses the concern of skewness in the chi-squared distribution. 
    
    \section{Experiments}\label{sec:experiment}
    \paragraph{Dataset} 
    The experiments are conducted using the DNA storage data introduced in~\cite{guo2022deep}. 
    This data\footnote{The data can be accessed 
    through \url{https://github.com/TeamErlich/dna-fountain} and 
    \url{https://www.ebi.ac.uk/ena/data/view/PRJEB19305}.} 
    was originally provided by a research called DNA-Fountain~\cite{erlich2017dna}, 
    which is a milestone in DNA storage. 
    Each data sample from the DNA storage dataset can be represented as a tuple $((\bm{s},\bm{t}),d)$, 
    where $(\bm{s},\bm{t})$ is a pair of DNA sequences 
    and $d$ is the groundtruth Levenshtein distance between these sequences. 
    The DNA sequences in the dataset are strings composed of oligonucleotides and sequencing failure bases, 
    denoted by the alphabet $\{\mathrm{A}, \mathrm{T}, \mathrm{G}, \mathrm{C}, \mathrm{N}\}$. 
    The samples can be categorized into two classes of homologous and non-homologous samples. 
    In a homologous sample $((\bm{s},\bm{t}),d)$, the sequences $\bm{s}$ and $\bm{t}$ belong to the same cluster 
    obtained from the DNA-storage pipeline and have a small Levenshtein distance $d$. 
    On the other hand, in a non-homologous sample $((\bm{s},\bm{t}),d)$, the sequences $\bm{s}$ and $\bm{t}$ 
    belong to different clusters and can be considered as independent sequences with a large Levenshtein distance $d$. 
    The length of the DNA sequences in the dataset is approximately $152$, 
    and they are padded to a fixed length of $160$ before being fed into the embedding network. 
    To ensure the separation of data for training and testing, 
    the training and testing sets of the DNA storage data are divided by a partition on the clusters 
    of the retrieved DNA sequences. 
    
    \paragraph{Metric} 
    Two metrics are used to evaluate the performance of the model, as they are, 
    the global approximation error ($\mathrm{AE}_g$) and the homologous approximation error ($\mathrm{AE}_h$). 
    The $\mathrm{AE}_g$ is calculated as the mean absolute error over the testing set $\mathrm{Te}$
    \begin{equation}
    \mathrm{AE}_g=\frac{1}{\#\mathrm{Te}}\sum_{((\bm{s},\bm{t}),d)\in \mathrm{Te}} |\hat{d}-d|, 
    \end{equation}
    while the $\mathrm{AE}_h$ is the mean absolute error over the homologous samples $\mathrm{Te}_h$ from the testing set
    \begin{equation}
    \mathrm{AE}_h=\frac{1}{\#\mathrm{Te}_{h}}\sum_{((\bm{s},\bm{t}),d)\in \mathrm{Te}_{h}} |d-\hat{d}|,
    \end{equation}
    where the $\#\mathrm{Te}$ and $\#\mathrm{Te}_{h}$ are the number of samples in $\mathrm{Te}$ and $\mathrm{Te}_{h}$, respectively. 
    The values of $\mathrm{AE}_g$ and $\mathrm{AE}_h$ provide insights into the overall approximation error 
    and the specific performance on homologous samples, respectively. 
    In general, the Levenshtein distance between two completely unrelated sequences is considered of lesser importance. 
    Therefore, the $\mathrm{AE}_h$ is a more important metric than $\mathrm{AE}_g$ in this study. 
    
    \paragraph{Models} 
    In the experiments, we employ a variety of neural network structures as the embedding network, 
    including the commonly used CNN-$5$, CNN-$10$, and GRU networks as utilized in previous studies 
    \cite{zhang2019neural,dai2020convolutional,guo2022deep}. 
    Additionally, 
    to explore the impact of network depth and width on the ESD and model performance, 
    we also use wider embedding networks, denoted as CNN-$5$-w and CNN-$10$-w. 
    The only difference between CNN-$\ast$-w and CNN-$\ast$ is that the CNN-$\ast$-w has more convolution channels. 
    These models are the $1$D versions from their original model. 
    Further details can be found in the Appendices.
    % \footnote{Please refer to \url{https://github.com/TeamErlich/dna-fountain} for the Appendices.}. 
    
    \subsection{The early-stopping dimension.} \label{subsec:exp-esd}
    In \cref{subsec:esd}, it is analyzed that there exists an appropriate embedding dimension called the ESD $n_0$, 
    with the fixed dataset and network architecture. 
    The theoretical approximation precision improves as 
    the embedding dimension $n$ increases, up to the point where $n$ reaches the ESD $n_0$. 
    
    \begin{figure*}[htb!]
        \centering
        \subfigure[$n=40$]
        {\includegraphics[width=0.21\textwidth]{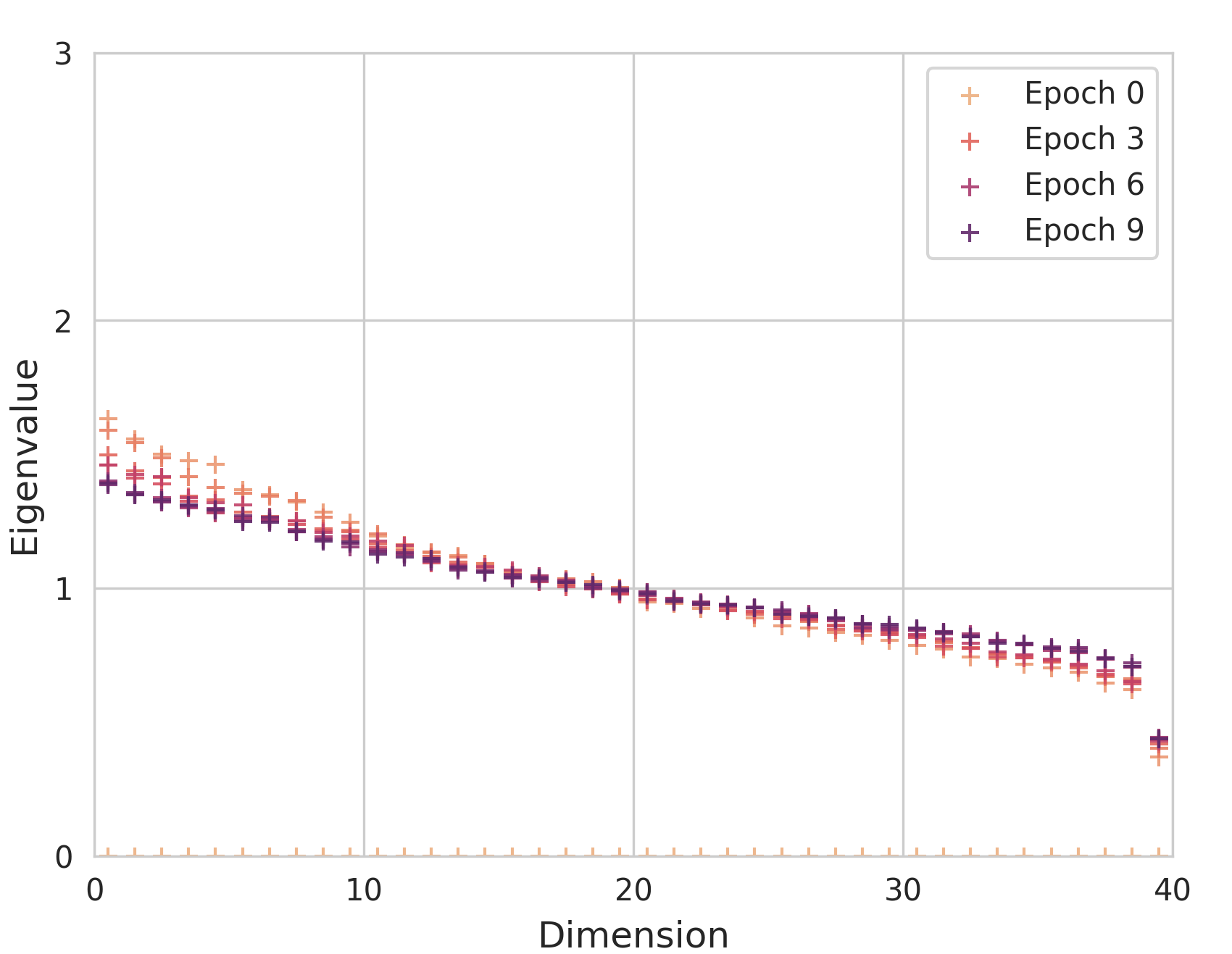}}
        \subfigure[$n=60$]
        {\includegraphics[width=0.21\textwidth]{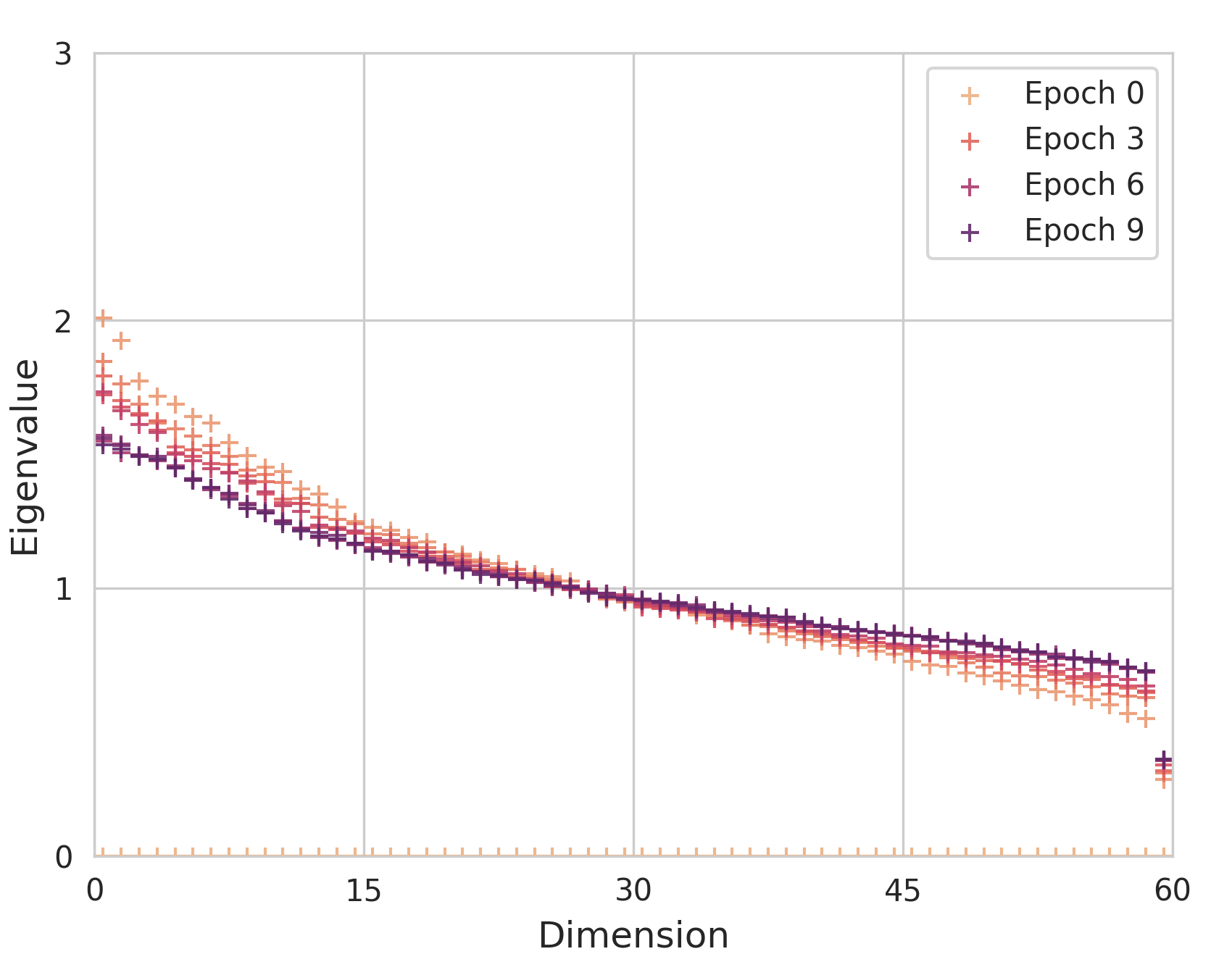}}
        \subfigure[$n=80$]
        {\includegraphics[width=0.21\textwidth]{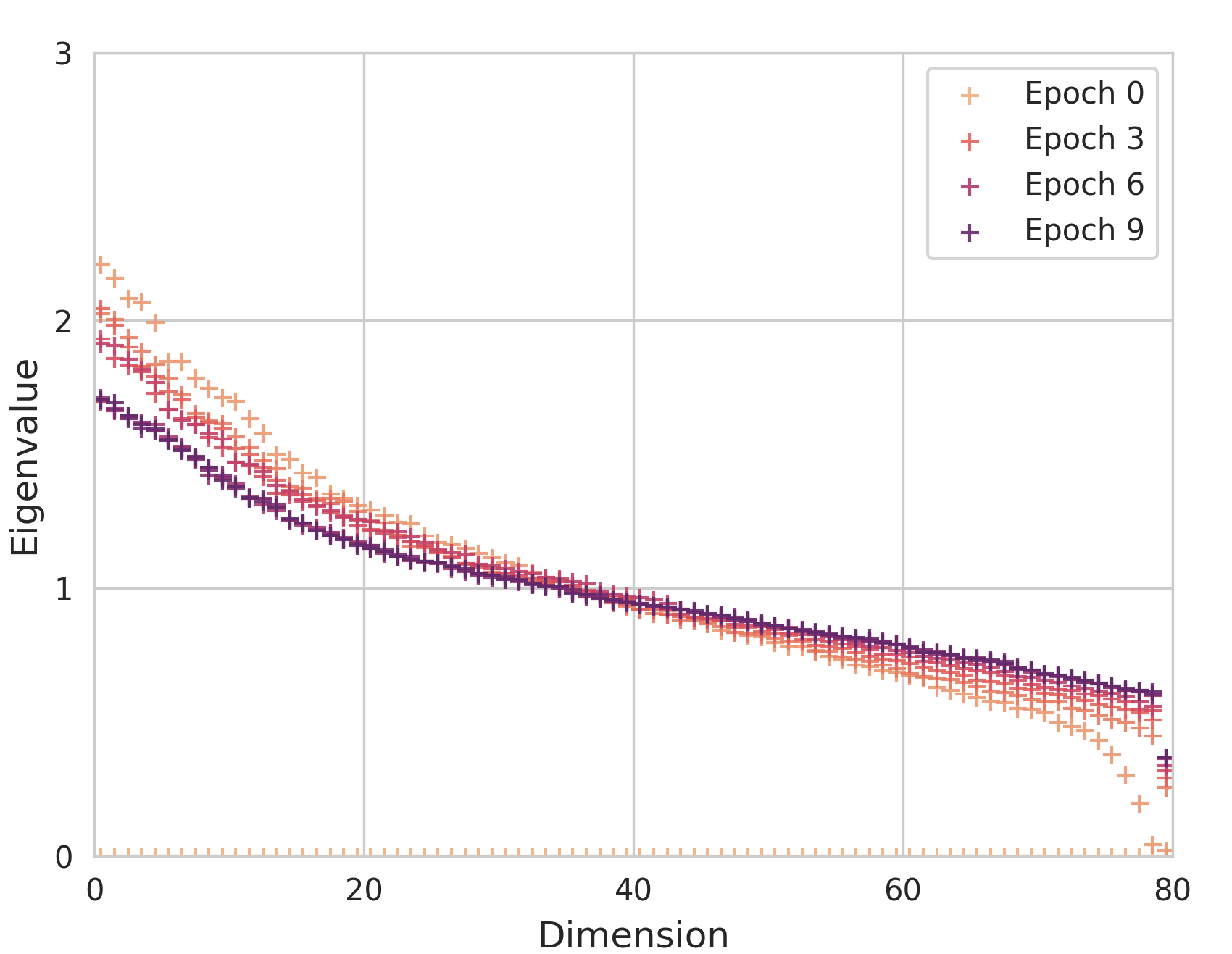}}
        \subfigure[$n=100$]
        {\includegraphics[width=0.21\textwidth]{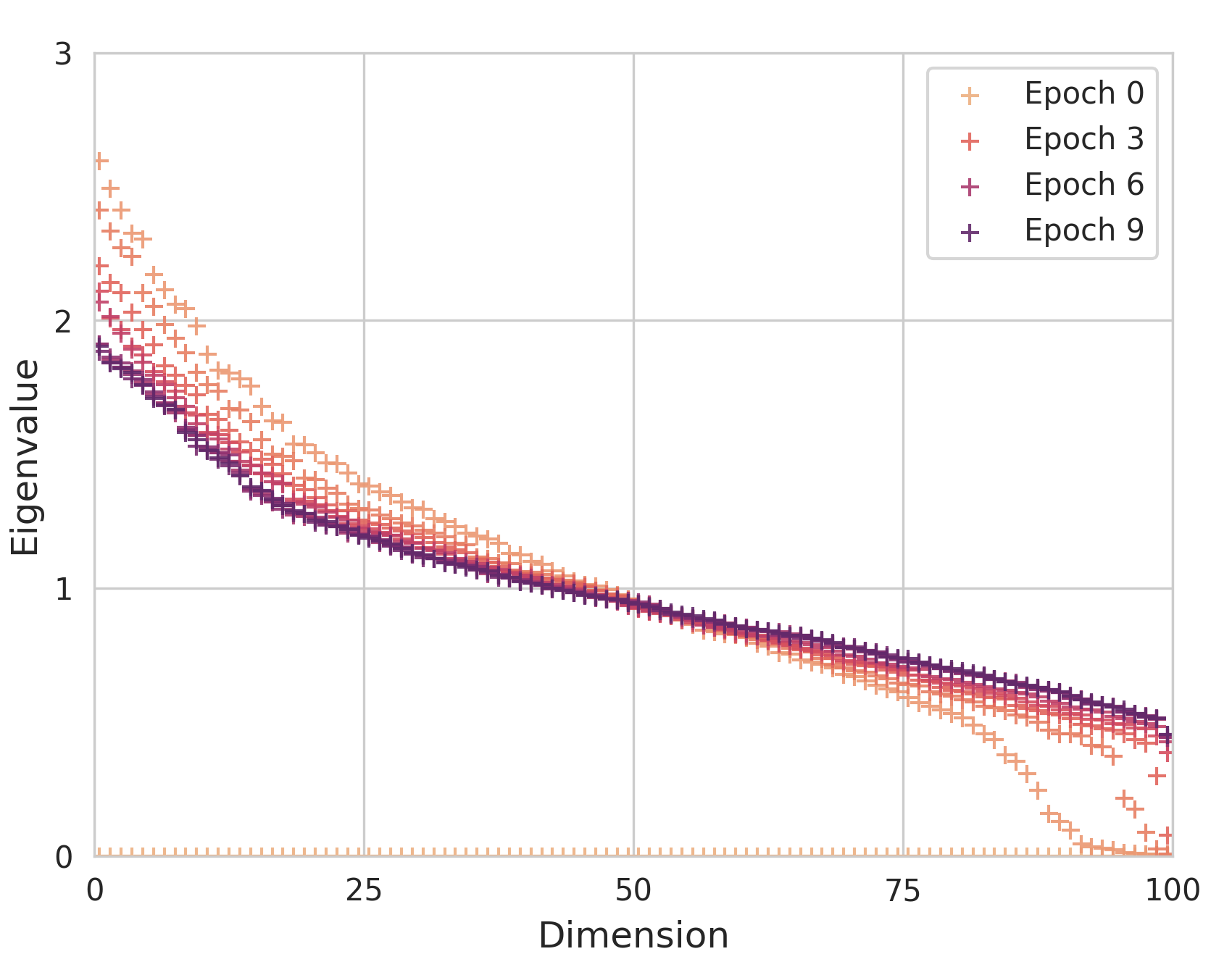}}
        \subfigure[$n=120$] 
        {\includegraphics[width=0.21\textwidth]{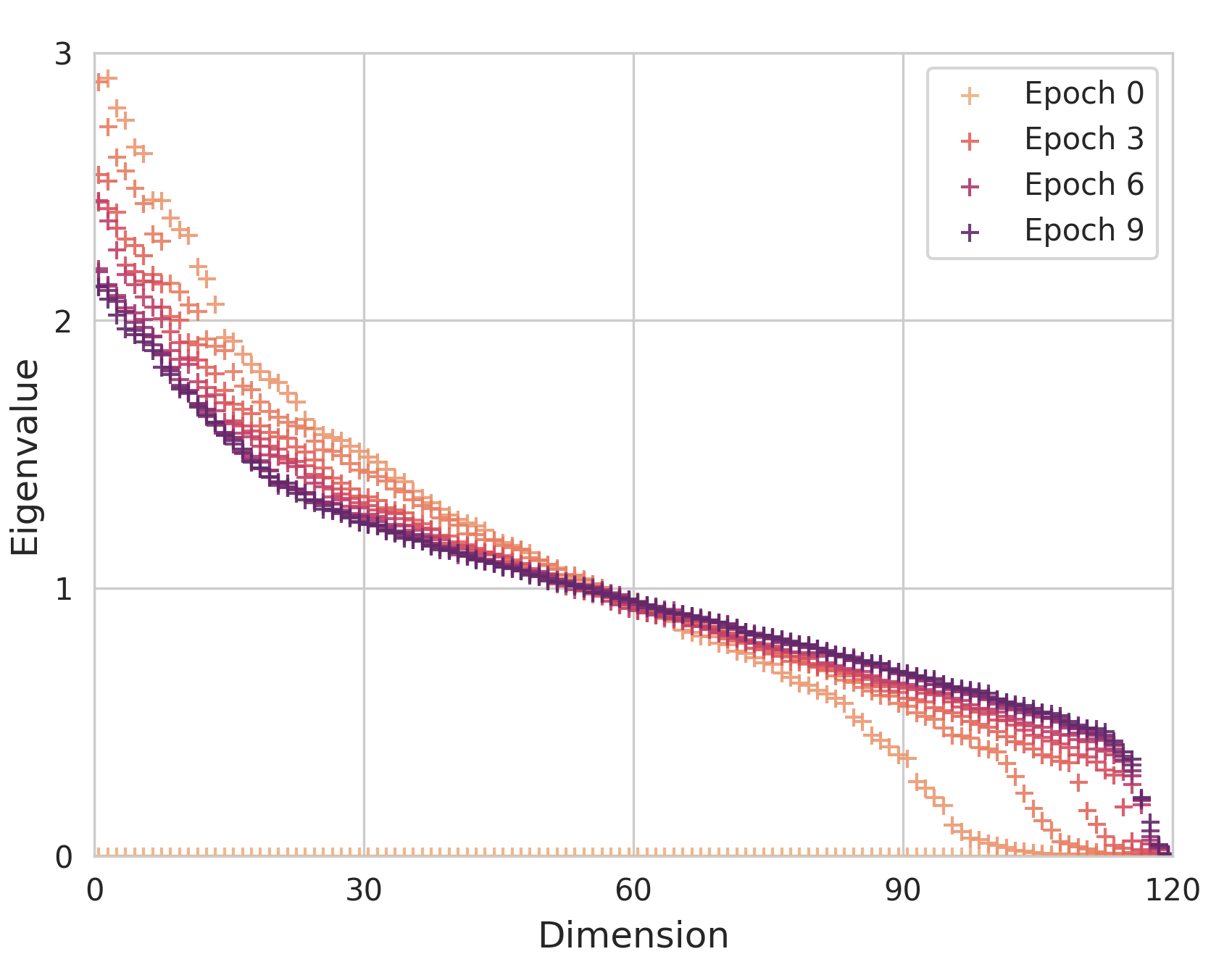}}
        \subfigure[$n=140$]
        {\includegraphics[width=0.21\textwidth]{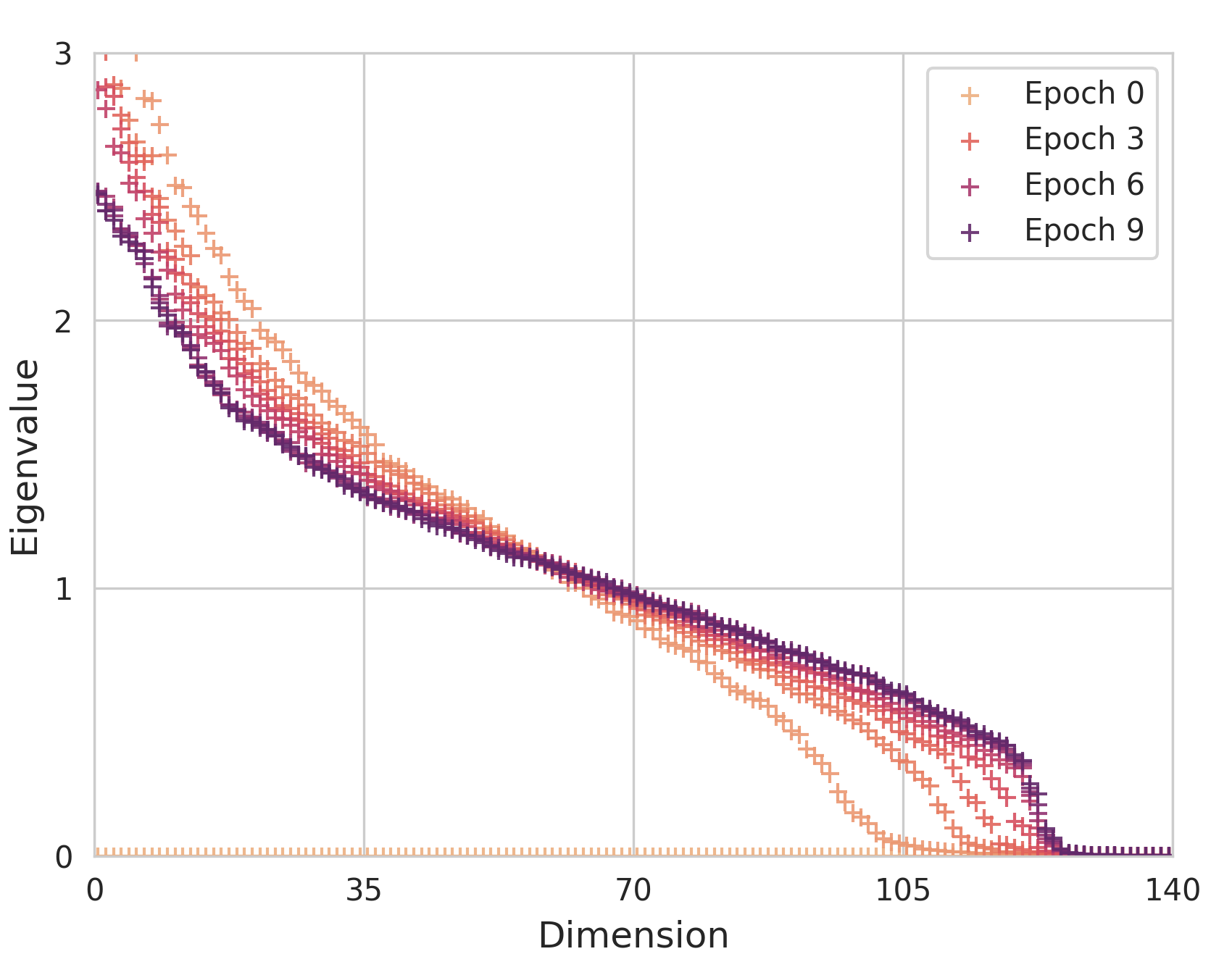}}
        \subfigure[$n=160$]
        {\includegraphics[width=0.21\textwidth]{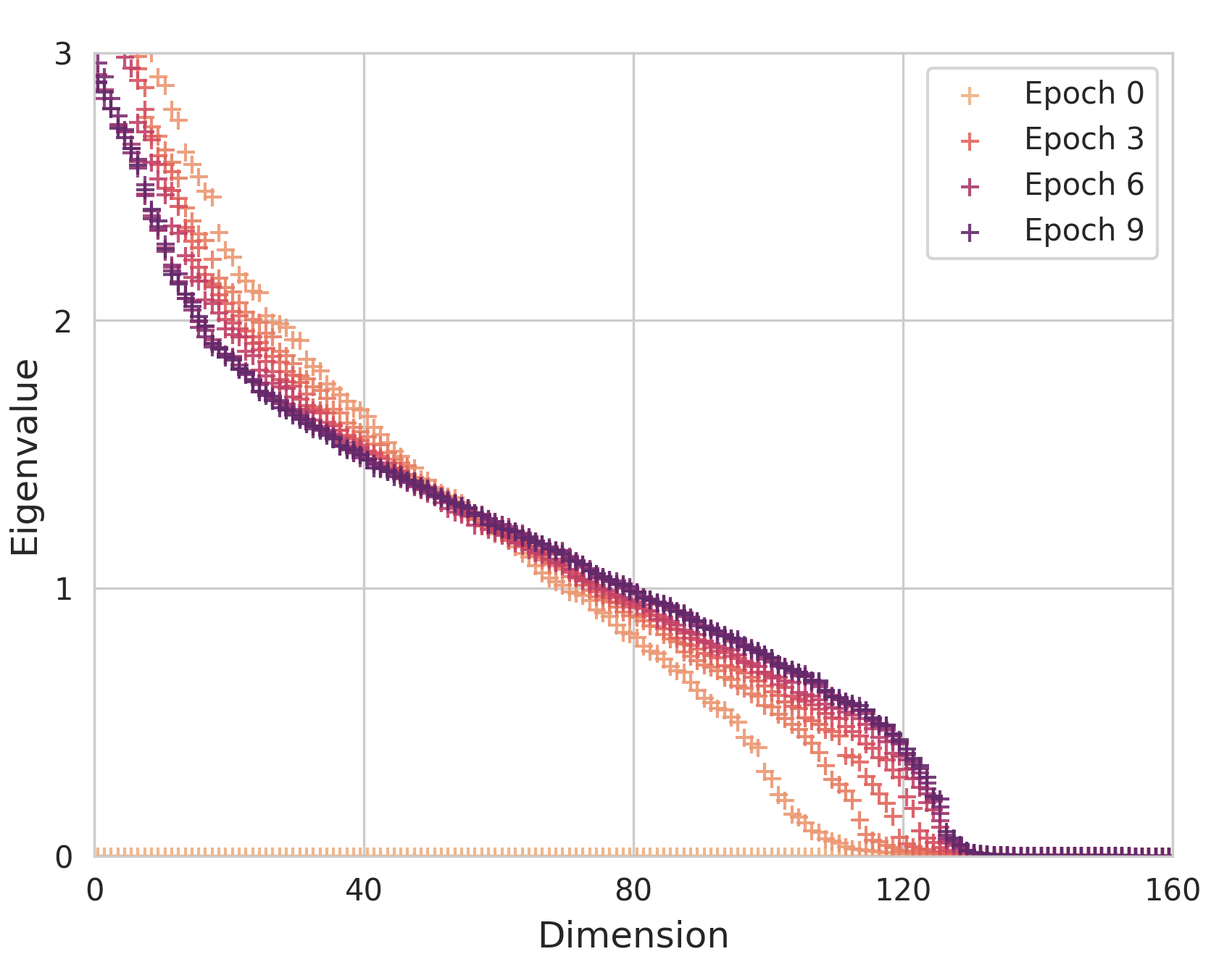}}
        \subfigure[$n=180$]
        {\includegraphics[width=0.21\textwidth]{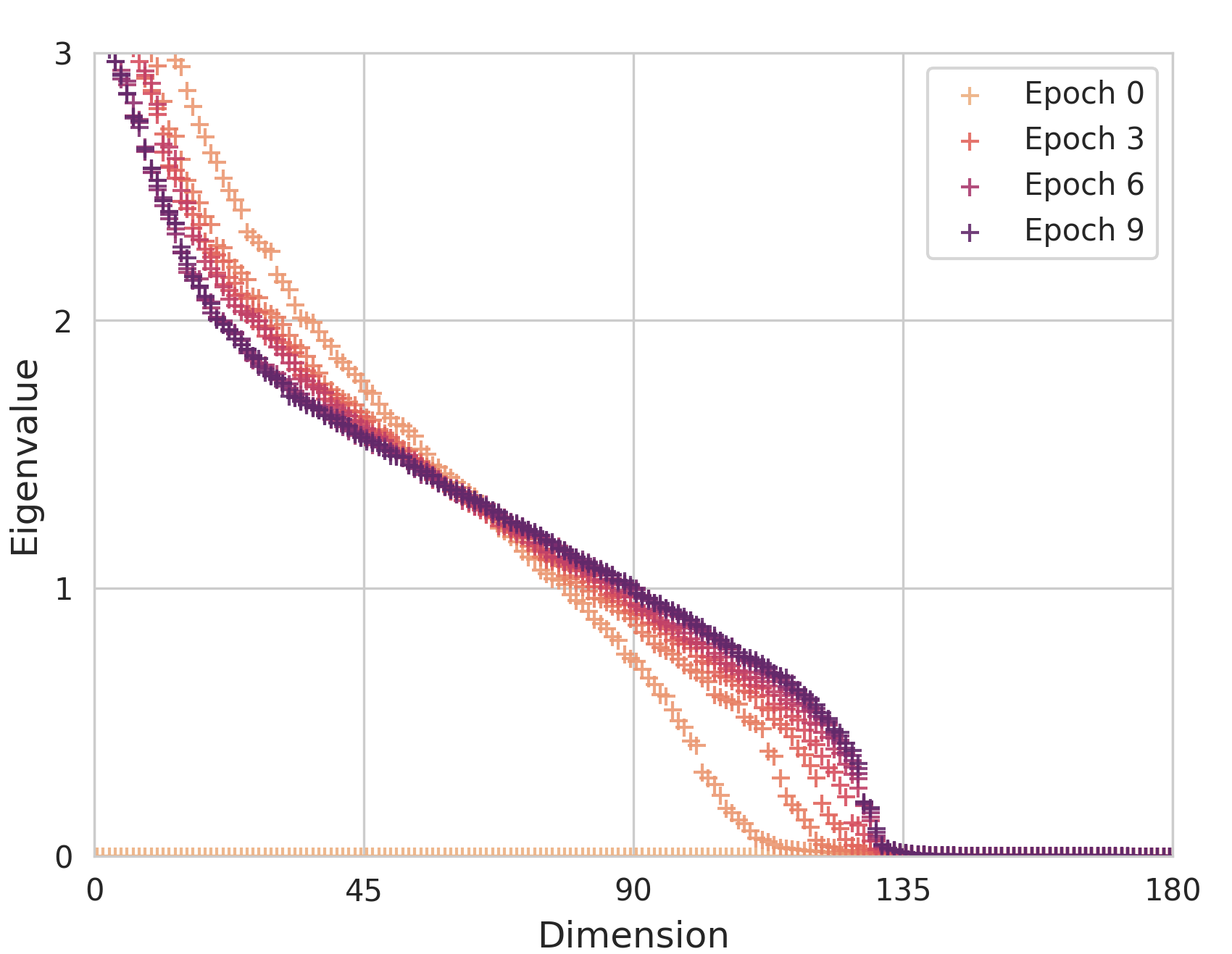}}
        \caption{The sorted eigenvalues of $\mathrm{cov}(\bm{u_i}-\bm{u_j},\bm{u_i}-\bm{u_j}), (i\neq j)$ 
            are plotted for different choices of the embedding dimension $n$ in the CNN-$5$ embedding network. 
            When $n$ is small, the eigenvalues are distributed around $1$, as in (a)--(d). 
            Increasing the embedding dimension $n$, the sorted eigenvalues decrease to $0$ after some dimension, as in (e)--(h). 
        }
        \label{fig:esd}
    \end{figure*}
    
    Specifically, there exists the ESD $n_0$. 
    When the embedding dimension $n$ is less than or equal to $n_0$, 
    the eigenvalues of $\mathrm{cov}(\bm{u_i}-\bm{u_j},\bm{u_i}-\bm{u_j}), (i\neq j)$ are approximately equal to $1$.  
    When $n$ exceeds $n_0$, the sorted eigenvalues gradually decrease to $0$ after reaching $n_0$.
    To verify this phenomenon, 
    we train embedding networks with different embedding dimensions, 
    and plot the sorted eigenvalues of $\mathrm{cov}(\bm{u_i}-\bm{u_j},\bm{u_i}-\bm{u_j}), (i\neq j)$. 
    The results for the embedding network of CNN-$5$ are plotted in \cref{fig:esd}. 
    It is suggested that, when the embedding dimension $n$ is small, 
    the eigenvalues are distributed around $1$, indicating that $n$ is smaller than the ESD $n_0$.
    As the embedding dimension increases, the sorted eigenvalues sharply decrease to $0$ after a certain dimension. 
    And regardless of how large the embedding dimension $n$ is, the number of non-zero eigenvalues 
    remains relatively constant. 
    This observation suggests that the embedding dimension exceeds the value of $n_0$. 
    Based on the results shown in \cref{fig:esd}, we determine that the ESD $n_0$ for the engaged dataset and the CNN-$5$  
    is approximately $n_0 = 120$. 
    
    It is worth noting that, when using the CNN-$5$-w or CNN-$10$-w as the embedding network, 
    the phenomenon of sudden vanishing eigenvalues along increasing embedding dimension becomes more evident. 
    Furthermore, the curves of approximation errors are more stable on the embedding networks CNN-$5$-w and CNN-$10$-w. 
    % For more details, please refer to \cref{app:esd}.
    For more details, please refer to the Appendices.
    
    \begin{figure}[htb!]
        \centering
        \subfigure[AE$_g$]
        {\includegraphics[width=0.43\linewidth]{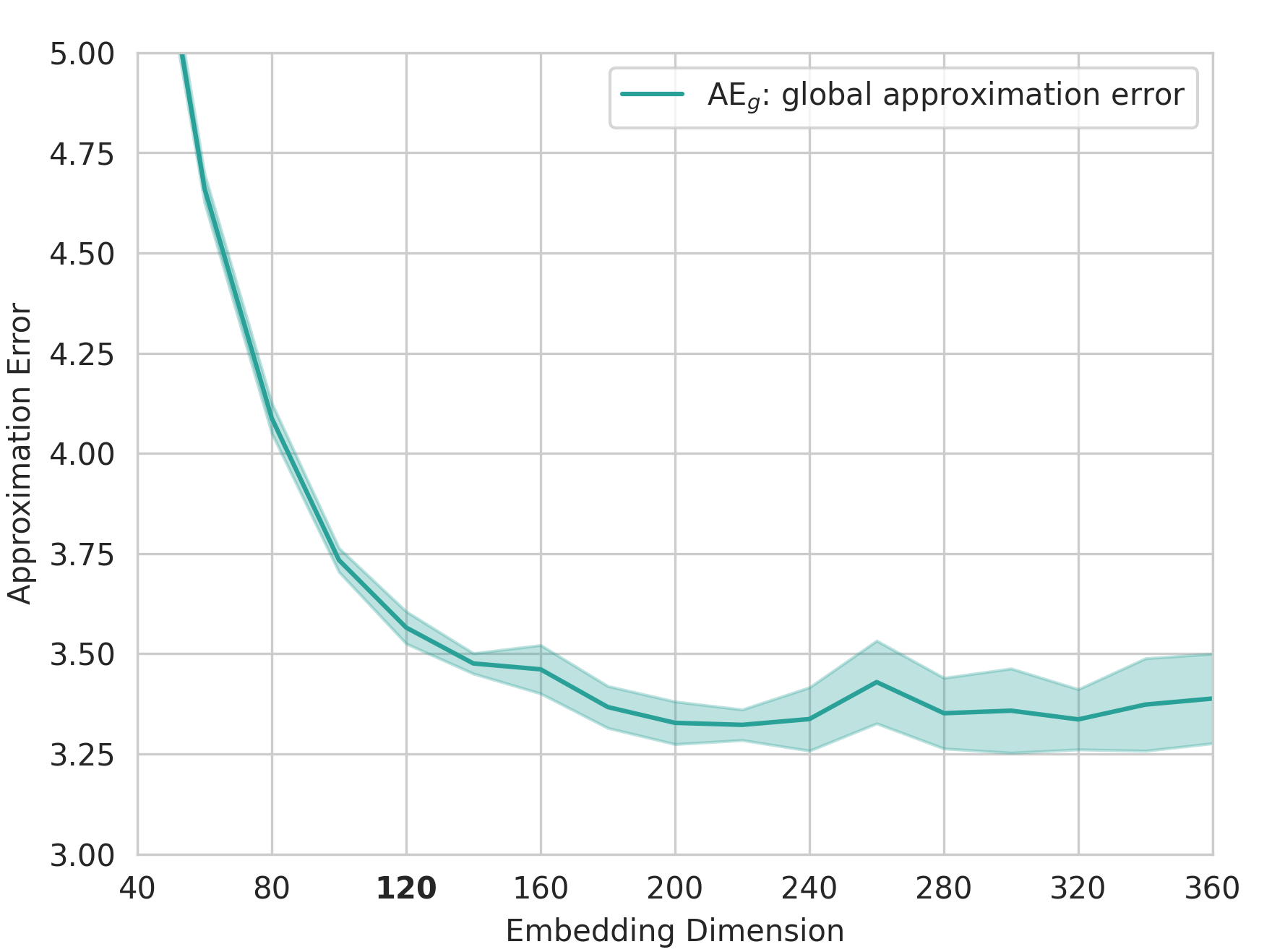}}
        \subfigure[AE$_h$]
        {\includegraphics[width=0.43\linewidth]{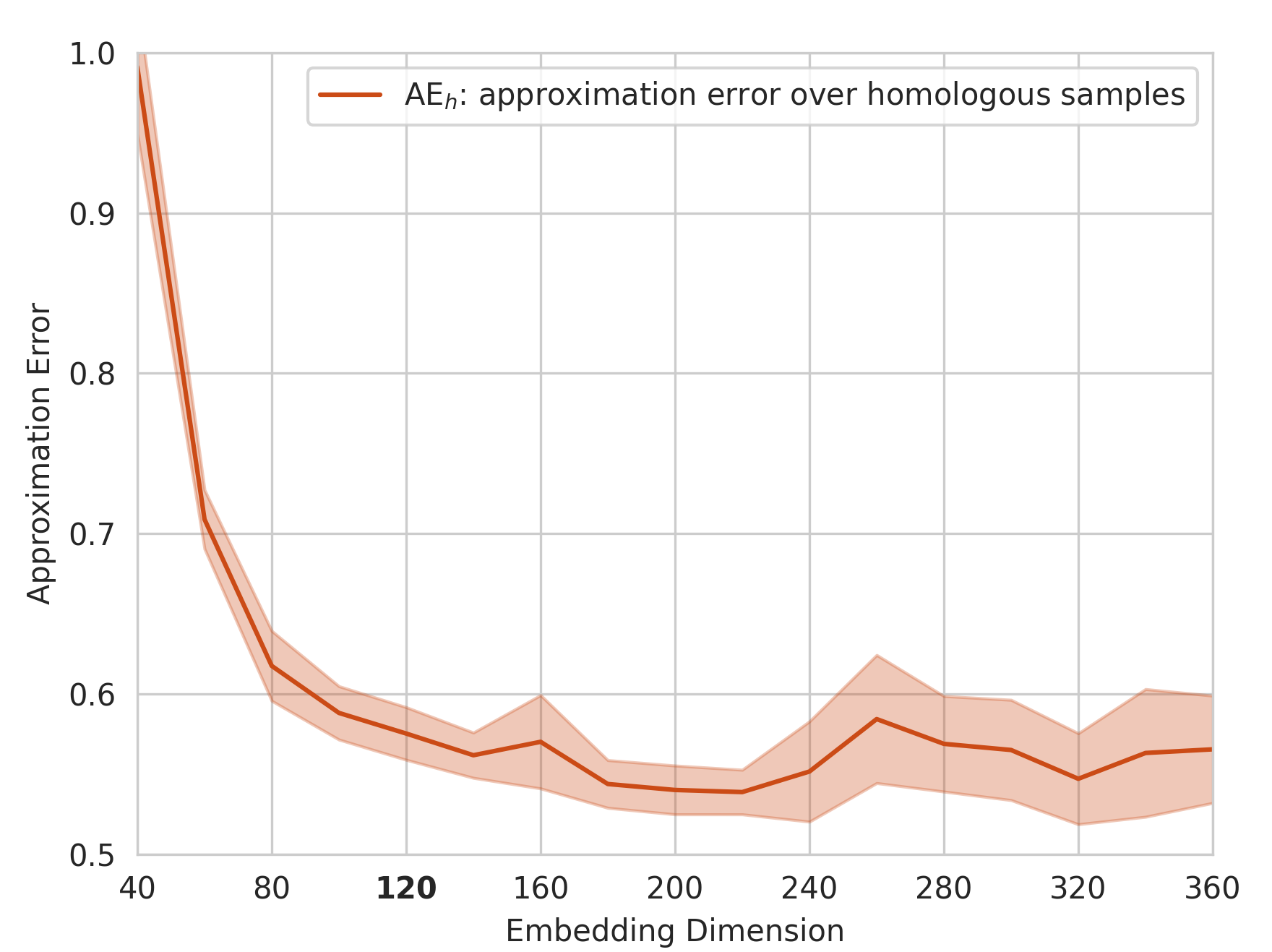}}
        \caption{
        The global approximation error and homologous approximation error are shown against the embedding dimension in (a) and (b), respectively. 
        The curves are plotted based on the mean and standard deviation over 5 runs. 
        The approximation errors decrease alongwith increase the embedding dimension $n$ until the ESD $n_0$, 
        which is $120$ for CNN-$5$. 
        When the $n>n_0$, there is no gain of the performance on a larger $n$, but the model performance becomes unstable with 
        a larger standard deviation.}
        \label{fig:esd-performance}
    \end{figure} 
        
    To verify the claim that the performance improves as the embedding dimension $n$ increases until reaching the ESD $n_0$, 
    the approximation precision of the CNN-$5$ model with different embedding dimension 
    is plotted in \cref{fig:esd-performance}. 
    This figure suggests that increasing the embedding dimension leads to a decrease in the approximation error when the 
    embedding dimension $n$ is less than the ESD $n_0=120$. 
    When the embedding dimension $n$ exceeds the ESD $n_0=120$, the improvement in approximation precision becomes negligible, 
    and the standard deviation of the approximation error across different runs grows. 
    This indicates that the stability of the embedding network in training phase is compromised. 
    
    In the Appendices, we provide similar figures to those in \cref{fig:esd} and \cref{fig:esd-performance} 
    for the CNN-$10$, CNN-$5$-w, CNN-$10$-w, and GRU embedding networks used in this study.
    The ESD values corresponding to these four networks are CNN-$10$: $n_0=120$, CNN-$5$-w: $n_0=140$, CNN-$10$-w: $n_0=140$, 
    and GRU: $n_0=140$. 
    Considering those figures and the results in the comparative experiments, we may speculate that the width of the network plays a more significant role than 
    the depth of the network in improving the approximation precision. 
    
    \begin{table*}[htb]
        \centering
        %   \renewcommand\arraystretch{1.7}
        % \begin{minipage}{\linewidth}
            % \resizebox{\linewidth}{!}{
        {\fontsize{9}{1}\selectfont
        \begin{tabular}{lccccc}
            % \begin{tabular}{x{0.20\linewidth}x{0.15\linewidth}x{0.15\linewidth}x{0.15\linewidth}x{0.15\linewidth}x{0.15\linewidth}}
                \toprule
                \multirow{2}{*}{\begin{tabular}[c]{@{}c@{}}Embedding \\Network\end{tabular}} & \multirow{2}{*}{Metric} & \multicolumn{4}{c}{Objective Function}                      \\\cmidrule{3-6}
                &                         & MSE            & MAE           & RE$\chi^2$    & PNLL          \\\midrule
                \multirow{2}{*}{CNN-$5$($80$)}                                             & AE$_g$                  & 4.06$\pm$0.07  & \textbf{4.00$\pm$0.02} & 4.32$\pm$0.01 & 4.09$\pm$0.04 \\
                & AE$_h$                  & 0.83$\pm$0.02  & 0.68$\pm$0.02 & 0.85$\pm$0.00 & \textbf{0.62$\pm$0.02} \\\midrule
                \multirow{2}{*}{CNN-$5$($120$)}                                            & AE$_g$                  & \textbf{3.48$\pm$0.02}  & 3.52$\pm$0.07 & 4.15$\pm$0.07 & 3.57$\pm$0.04 \\
                & AE$_h$                  & 0.82$\pm$0.02  & 0.64$\pm$0.03 & 0.83$\pm$0.01 & \textbf{0.58$\pm$0.02} \\\midrule
                \multirow{2}{*}{CNN-$10$($80$)}                                            & AE$_g$                  & 3.78$\pm$0.02  & \textbf{3.76$\pm$0.08} & 4.24$\pm$0.05 & 3.90$\pm$0.02 \\
                & AE$_h$                  & 0.72$\pm$0.02  & 0.61$\pm$0.01 & 0.87$\pm$0.00 & \textbf{0.58$\pm$0.01} \\\midrule
                \multirow{2}{*}{CNN-$10$($120$)}                                           & AE$_g$                  & \textbf{3.27$\pm$0.04}  & 3.44$\pm$0.04 & 4.00$\pm$0.04 & 3.35$\pm$0.05 \\
                & AE$_h$                  & 0.72$\pm$0.02  & 0.64$\pm$0.03 & 0.84$\pm$0.01 & \textbf{0.55$\pm$0.01} \\\midrule
                \multirow{2}{*}{CNN-$5$-w($80$)}                                           & AE$_g$                  & \textbf{3.84$\pm$0.01} & 3.85$\pm$0.01 & 4.45$\pm$0.04 & 3.95$\pm$0.02 \\
                & AE$_h$                  & 0.69$\pm$0.01 & 0.59$\pm$0.00 & 0.85$\pm$0.00 & \textbf{0.54$\pm$0.01} \\\midrule
                \multirow{2}{*}{CNN-$5$-w($140$)}                                          & AE$_g$                  & \textbf{2.98$\pm$0.01} & 3.08$\pm$0.02 & 4.17$\pm$0.04 & 3.11$\pm$0.03 \\
                & AE$_h$                  & 0.67$\pm$0.00 & 0.56$\pm$0.00 & 0.82$\pm$0.01 & \textbf{0.50$\pm$0.00} \\\midrule
                \multirow{2}{*}{CNN-$10$-w($80$)}                                          & AE$_g$                  & \textbf{3.70$\pm$0.01} & 3.74$\pm$0.01 & 4.63$\pm$0.03 & 3.76$\pm$0.01 \\
                & AE$_h$                  & 0.65$\pm$0.01 & 0.56$\pm$0.01 & 0.87$\pm$0.01 & \textbf{0.48$\pm$0.01} \\\midrule
                \multirow{2}{*}{CNN-$10$-w($140$)}                                         & AE$_g$                  & \textbf{2.90$\pm$0.01} & 3.14$\pm$0.02 & 4.50$\pm$0.05 & 3.00$\pm$0.02 \\
                & AE$_h$                  & 0.62$\pm$0.00 & 0.54$\pm$0.00 & 0.84$\pm$0.00 & \textbf{0.47$\pm$0.00} \\\midrule
                \multirow{2}{*}{GRU($80$)}                                                 & AE$_g$                  & \textbf{4.08$\pm$0.01} & 4.12$\pm$0.01 & 5.81$\pm$0.10 & 4.34$\pm$0.02 \\
                & AE$_h$                  & 0.79$\pm$0.00 & 0.73$\pm$0.00 & 0.85$\pm$0.00 & \textbf{0.61$\pm$0.00} \\\midrule
                \multirow{2}{*}{GRU($140$)}                                                & AE$_g$                  & \textbf{3.49$\pm$0.02} & 3.55$\pm$0.01 & 6.99$\pm$0.05 & 4.54$\pm$0.04 \\
                & AE$_h$                  & 0.91$\pm$0.01 & 0.80$\pm$0.00 & 0.84$\pm$0.01 & \textbf{0.64$\pm$0.00} \\
                \bottomrule 
        \end{tabular}}
        \caption{Results of the experiments on DNA storage data. Contains all combinations of models, dimensions and losses.
            The results are reported in the format ``$\mathrm{mean} \pm \mathrm{std}$'' of the mean value and the standard deviation over 5 runs of the
            experiments.}\label{tab:results}
                % }
            % \end{minipage}
    \end{table*}
    
    \subsection{The comparative experiments.}\label{subsec:com} 
    We conduct experiments using different combinations of embedding networks, loss functions, and embedding dimensions. 
    The embedding networks include CNN-$5$, CNN-$10$, CNN-$5$-w, CNN-$10$-w, and GRU.  
    The loss functions used in the experiments are MSE, MAE, RE$\chi^2$ from \cite{guo2022deep}, and the proposed PNLL. 
    The choices of embedding dimensions are an empirical fixed dimension of $80$ and the ESDs corresponding to the engaged embedding networks. 
    The results of these experimental settings are presented in \cref{tab:results}. 
    It is worth noting that, unlike in \cite{guo2022deep},
    we utilize a learnable scaling factor instead of a predefined fixed one. 
    As a result, the obtained results may differ from those reported in \cite{guo2022deep}, 
    even though the overall setting appears similar.

    As shown in \cref{tab:results}, the proposed Poisson regression and PNLL achieve the best performance 
    in terms of the metric AE$_h$ across all the embedding networks. 
    Although the performance of PNLL in AE$_g$ is slightly behind the best performance, 
    the relative difference is not significant. 
    Moreover, as mentioned earlier, the AE$_h$ is a more crucial metric of interest than AE$_g$ in the experiments.
    These results indicate that the proposed Poisson regression outperforms other objective functions. 
    
    It can be also observed that increasing the embedding dimension from $80$ to the ESD of the respective embedding network 
    leads to better performance in both AE$_h$ and AE$_g$, especially a significant improvement in AE$_g$. 
    This finding highlights the effectiveness of the proposed ESD. 
    However, for the GRU model, increasing the embedding dimension results in worse performance. 
    It is speculated that the assumption \ref{ass:bound} may not hold for the GRU model, as discussed in the Appendices. 
    Overall, it is evident from \cref{tab:results} that the combination of the proposed PNLL and the proposed ESD 
    yields the best performance. 
    
    Comparing models with different numbers of layers in the convolutional network, 
    the improvement resulting from increasing the number of layers is found to be negligible. 
    However, increasing the number of channels in the model leads to improved performance 
    across most of the convolutional embedding networks. 
    Furthermore, increasing the depth of the model results in a higher ESD, 
    indicating that the model has a greater capacity to capture and express features from its input. 
    Based on these observations, we can infer that width (number of channels) plays a more crucial role 
    than depth (number of layers) in the task of Levenshtein distance embedding. 

    Readers may be concerned about computational overhead. 
    Given that changes in embedding dimensions only affect the neural network's embedding top, 
    the experimental time complexities are similar across different choices of embedding dimensions. 
    Calculating the loss functions, as outlined in \cref{eqn:rex} and \cref{eqn:poiNLL}, is not computationally intensive. 
    Therefore, the differences in time complexities due to the engaged loss functions are negligible. 
    
    In summary, the CNN-$10$-w embedding network, with an ESD embedding dimension of $n_0=140$ and 
    optimized by the Poisson regression reports the highest performance among all the models. 
    
\section{Conclusion}\label{sec:conclution}
    Deep embedding is an effective way for fast approximation of Levenshtein distance. 
    We propose a deep Levenshtein distance embedding algorithm based on Poisson regression. 
    The implementation of our algorithm is based on two key points: embedding vector dimension and model training method. 
    The adaptation of Poisson distribution to Levenshtein distance was theoretically analyzed under reasonable assumptions. 
    Our method achieves a more accurate approximation on DNA storage data. Moreover, the proposed algorithm is robust to different 
    embedding models and all show positive results. We hope our method will be useful with other
    approximate tasks that involve discrete labels. 

\section*{Acknowledgements}
The authors are grateful to the reviewers for their valuable comments. 
This work was supported by the National Key Research and Development Program of China under Grant 
2020YFA0712100 and the National Natural Science Foundation of China.

\bibliography{Xiang-Wei} 

\begin{thebibliography}{32}
\providecommand{\natexlab}[1]{#1}

\bibitem[{Backurs and Indyk(2015)}]{backurs2015edit}
Backurs, A.; and Indyk, P. 2015.
\newblock Edit distance cannot be computed in strongly subquadratic time
  (unless SETH is false).
\newblock In \emph{Proceedings of the forty-seventh annual ACM symposium on
  Theory of computing}, 51--58.

\bibitem[{Bar-Lev, Etzion, and Yaakobi(2023)}]{bar2023size}
Bar-Lev, D.; Etzion, T.; and Yaakobi, E. 2023.
\newblock On the Size of Balls and Anticodes of Small Diameter Under the
  Fixed-Length Levenshtein Metric.
\newblock \emph{IEEE Transactions on Information Theory}, 69(4): 2324--2340.

\bibitem[{Berger, Waterman, and Yu(2020)}]{berger2020levenshtein}
Berger, B.; Waterman, M.~S.; and Yu, Y.~W. 2020.
\newblock Levenshtein distance, sequence comparison and biological database
  search.
\newblock \emph{IEEE transactions on information theory}, 67(6): 3287--3294.

\bibitem[{Bromley et~al.(1993)Bromley, Guyon, LeCun, S\"{a}ckinger, and
  Shah}]{bromley1993signature}
Bromley, J.; Guyon, I.; LeCun, Y.; S\"{a}ckinger, E.; and Shah, R. 1993.
\newblock {Signature Verification Using a "Siamese" Time Delay Neural Network}.
\newblock In \emph{Proceedings of the 6th International Conference on Neural
  Information Processing Systems}, NIPS'93, 737–744. San Francisco, CA, USA:
  Morgan Kaufmann Publishers Inc.

\bibitem[{Chakraborty, Goldenberg, and
  Kouck{\`y}(2016)}]{chakraborty2016streaming}
Chakraborty, D.; Goldenberg, E.; and Kouck{\`y}, M. 2016.
\newblock Streaming algorithms for embedding and computing edit distance in the
  low distance regime.
\newblock In \emph{Proceedings of the forty-eighth annual ACM symposium on
  Theory of Computing}, 712--725.

\bibitem[{Chami et~al.(2020)Chami, Gu, Chatziafratis, and
  R{\'e}}]{chami2020trees}
Chami, I.; Gu, A.; Chatziafratis, V.; and R{\'e}, C. 2020.
\newblock From trees to continuous embeddings and back: Hyperbolic hierarchical
  clustering.
\newblock \emph{Advances in Neural Information Processing Systems}, 33:
  15065--15076.

\bibitem[{Cho et~al.(2014)Cho, van Merri{\"e}nboer, Gulcehre, Bahdanau,
  Bougares, Schwenk, and Bengio}]{cho2014learning}
Cho, K.; van Merri{\"e}nboer, B.; Gulcehre, C.; Bahdanau, D.; Bougares, F.;
  Schwenk, H.; and Bengio, Y. 2014.
\newblock Learning Phrase Representations using {RNN} Encoder{--}Decoder for
  Statistical Machine Translation.
\newblock In \emph{Proceedings of the 2014 Conference on Empirical Methods in
  Natural Language Processing ({EMNLP})}, 1724--1734. Doha, Qatar: Association
  for Computational Linguistics.

\bibitem[{Church, Gao, and Kosuri(2012)}]{church2012next}
Church, G.~M.; Gao, Y.; and Kosuri, S. 2012.
\newblock Next-generation digital information storage in DNA.
\newblock \emph{Science}, 337(6102): 1628--1628.

\bibitem[{Corso et~al.(2021)Corso, Ying, P{\'a}ndy, Veli{\v{c}}kovi{\'c},
  Leskovec, and Li{\`o}}]{corso2021neural}
Corso, G.; Ying, Z.; P{\'a}ndy, M.; Veli{\v{c}}kovi{\'c}, P.; Leskovec, J.; and
  Li{\`o}, P. 2021.
\newblock Neural distance embeddings for biological sequences.
\newblock \emph{Advances in Neural Information Processing Systems}, 34:
  18539--18551.

\bibitem[{Dai et~al.(2020)Dai, Yan, Zhou, Wang, Yang, and
  Cheng}]{dai2020convolutional}
Dai, X.; Yan, X.; Zhou, K.; Wang, Y.; Yang, H.; and Cheng, J. 2020.
\newblock Convolutional embedding for edit distance.
\newblock In \emph{proceedings of the 43rd international ACM SIGIR conference
  on Research and Development in information retrieval}, 599--608.

\bibitem[{Dong et~al.(2020)Dong, Sun, Ping, Ouyang, and Qian}]{dong2020dna}
Dong, Y.; Sun, F.; Ping, Z.; Ouyang, Q.; and Qian, L. 2020.
\newblock {DNA} storage: research landscape and future prospects.
\newblock \emph{National Science Review}, 7(6): 1092--1107.

\bibitem[{Erlich and Zielinski(2017)}]{erlich2017dna}
Erlich, Y.; and Zielinski, D. 2017.
\newblock {DNA} Fountain enables a robust and efficient storage architecture.
\newblock \emph{science}, 355(6328): 950--954.

\bibitem[{Goldman et~al.(2013)Goldman, Bertone, Chen, Dessimoz, LeProust,
  Sipos, and Birney}]{goldman2013towards}
Goldman, N.; Bertone, P.; Chen, S.; Dessimoz, C.; LeProust, E.~M.; Sipos, B.;
  and Birney, E. 2013.
\newblock Towards practical, high-capacity, low-maintenance information storage
  in synthesized {DNA}.
\newblock \emph{nature}, 494(7435): 77--80.

\bibitem[{Grass et~al.(2015)Grass, Heckel, Puddu, Paunescu, and
  Stark}]{grass2015robust}
Grass, R.~N.; Heckel, R.; Puddu, M.; Paunescu, D.; and Stark, W.~J. 2015.
\newblock Robust chemical preservation of digital information on {DNA} in
  silica with error-correcting codes.
\newblock \emph{Angewandte Chemie International Edition}, 54(8): 2552--2555.

\bibitem[{Guo, Liang, and Hou(2022)}]{guo2022deep}
Guo, A.~J.; Liang, C.; and Hou, Q.-H. 2022.
\newblock Deep Squared Euclidean Approximation to the Levenshtein Distance for
  {DNA} Storage.
\newblock In \emph{International Conference on Machine Learning}, 8095--8108.
  PMLR.

\bibitem[{Haldar and Mukhopadhyay(2011)}]{haldar2011levenshtein}
Haldar, R.; and Mukhopadhyay, D. 2011.
\newblock Levenshtein distance technique in dictionary lookup methods: An
  improved approach.
\newblock \emph{arXiv preprint arXiv:1101.1232}.

\bibitem[{He et~al.(2020)He, Fan, Wu, Xie, and Girshick}]{he2020momentum}
He, K.; Fan, H.; Wu, Y.; Xie, S.; and Girshick, R. 2020.
\newblock Momentum contrast for unsupervised visual representation learning.
\newblock In \emph{Proceedings of the IEEE/CVF conference on computer vision
  and pattern recognition}, 9729--9738.

\bibitem[{Jiang et~al.(2014)Jiang, Li, Feng, and Li}]{jiang2014string}
Jiang, Y.; Li, G.; Feng, J.; and Li, W.-S. 2014.
\newblock String similarity joins: An experimental evaluation.
\newblock \emph{Proceedings of the VLDB Endowment}, 7(8): 625--636.

\bibitem[{Levenshtein et~al.(1966)}]{levenshtein1966binary}
Levenshtein, V.~I.; et~al. 1966.
\newblock Binary codes capable of correcting deletions, insertions, and
  reversals.
\newblock In \emph{Soviet physics doklady}, volume~10, 707--710. Soviet Union.

\bibitem[{Li and Homer(2010)}]{li2010survey}
Li, H.; and Homer, N. 2010.
\newblock {A survey of sequence alignment algorithms for next-generation
  sequencing}.
\newblock \emph{Briefings in Bioinformatics}, 11(5): 473--483.

\bibitem[{Logan et~al.(2022)Logan, Fleischmann, Annis, Wehe, Tilly, Woods, and
  Khrapko}]{logan20223gold}
Logan, R.; Fleischmann, Z.; Annis, S.; Wehe, A.~W.; Tilly, J.~L.; Woods, D.~C.;
  and Khrapko, K. 2022.
\newblock 3GOLD: optimized Levenshtein distance for clustering third-generation
  sequencing data.
\newblock \emph{BMC bioinformatics}, 23(1): 1--18.

\bibitem[{Ostrovsky and Rabani(2007)}]{ostrovsky2007low}
Ostrovsky, R.; and Rabani, Y. 2007.
\newblock Low distortion embeddings for edit distance.
\newblock \emph{Journal of the ACM (JACM)}, 54(5): 23--es.

\bibitem[{Press et~al.(2020)Press, Hawkins, Jones~Jr, Schaub, and
  Finkelstein}]{press2020hedges}
Press, W.~H.; Hawkins, J.~A.; Jones~Jr, S.~K.; Schaub, J.~M.; and Finkelstein,
  I.~J. 2020.
\newblock {HEDGES} error-correcting code for {DNA} storage corrects indels and
  allows sequence constraints.
\newblock \emph{Proceedings of the National Academy of Sciences}, 117(31):
  18489--18496.

\bibitem[{Qu, Yan, and Wu(2022)}]{qu2022clover}
Qu, G.; Yan, Z.; and Wu, H. 2022.
\newblock {Clover: tree structure-based efficient DNA clustering for DNA-based
  data storage}.
\newblock \emph{Briefings in Bioinformatics}, 23(5).
\newblock Bbac336.

\bibitem[{Rashtchian et~al.(2017)Rashtchian, Makarychev, Racz, Ang, Jevdjic,
  Yekhanin, Ceze, and Strauss}]{rashtchian2017clustering}
Rashtchian, C.; Makarychev, K.; Racz, M.; Ang, S.; Jevdjic, D.; Yekhanin, S.;
  Ceze, L.; and Strauss, K. 2017.
\newblock Clustering billions of reads for {DNA} data storage.
\newblock \emph{Advances in Neural Information Processing Systems}, 30.

\bibitem[{Sberro et~al.(2019)Sberro, Fremin, Zlitni, Edfors, Greenfield,
  Snyder, Pavlopoulos, Kyrpides, and Bhatt}]{sberro2019large}
Sberro, H.; Fremin, B.~J.; Zlitni, S.; Edfors, F.; Greenfield, N.; Snyder,
  M.~P.; Pavlopoulos, G.~A.; Kyrpides, N.~C.; and Bhatt, A.~S. 2019.
\newblock Large-scale analyses of human microbiomes reveal thousands of small,
  novel genes.
\newblock \emph{Cell}, 178(5): 1245--1259.

\bibitem[{Su et~al.(2008)Su, Ahn, Eom, Kang, Kim, and Kim}]{su2008plagiarism}
Su, Z.; Ahn, B.-R.; Eom, K.-Y.; Kang, M.-K.; Kim, J.-P.; and Kim, M.-K. 2008.
\newblock Plagiarism detection using the Levenshtein distance and
  Smith-Waterman algorithm.
\newblock In \emph{2008 3rd International Conference on Innovative Computing
  Information and Control}, 569--569. IEEE.

\bibitem[{Wagner and Fischer(1974)}]{wagner1974string}
Wagner, R.~A.; and Fischer, M.~J. 1974.
\newblock The string-to-string correction problem.
\newblock \emph{Journal of the ACM (JACM)}, 21(1): 168--173.

\bibitem[{Welzel et~al.(2023)Welzel, Schwarz, L{\"o}chel, Kabdullayeva,
  Clemens, Becker, Freisleben, and Heider}]{welzel2023dna}
Welzel, M.; Schwarz, P.~M.; L{\"o}chel, H.~F.; Kabdullayeva, T.; Clemens, S.;
  Becker, A.; Freisleben, B.; and Heider, D. 2023.
\newblock {DNA}-Aeon provides flexible arithmetic coding for constraint
  adherence and error correction in {DNA} storage.
\newblock \emph{Nature Communications}, 14(1): 628.

\bibitem[{Zhang, Yuan, and Indyk(2019)}]{zhang2019neural}
Zhang, X.; Yuan, Y.; and Indyk, P. 2019.
\newblock Neural embeddings for nearest neighbor search under edit distance.

\bibitem[{Zheng et~al.(2019)Zheng, Yang, Genco, Wactawski-Wende, Buck, and
  Sun}]{zheng2019sense}
Zheng, W.; Yang, L.; Genco, R.~J.; Wactawski-Wende, J.; Buck, M.; and Sun, Y.
  2019.
\newblock SENSE: Siamese neural network for sequence embedding and
  alignment-free comparison.
\newblock \emph{Bioinformatics}, 35(11): 1820--1828.

\bibitem[{Zorita, Cuscó, and Filion(2015)}]{zorita2015starcode}
Zorita, E.; Cuscó, P.; and Filion, G.~J. 2015.
\newblock {Starcode: sequence clustering based on all-pairs search}.
\newblock \emph{Bioinformatics}, 31(12): 1913--1919.

\end{thebibliography}

\clearpage
\appendix
\section{Details on the embedding networks} \label{app:arch}
Five models, namely CNN-$5$, CNN-$10$, CNN-$5$-w, CNN-$10$-w, and GRU are utilized as the embedding networks in this study, 
following the architecture settings described in~\cite{guo2022deep,dai2020convolutional}. 
The CNN-$5$ consists of five $1$D-convolutional layers, while the CNN-$10$ has ten $1$D-convolutional layers. 
Each $1$D-convolutional layer has output channels of $64$, a kernel size of $3$, a stride of $1$, and a padding size of $1$. 
Average pooling layers with a kernel size of $2$ and ReLU activation function are also applied. 
To generate the embedding vector, two fully connected layers and a final batch normalization layer are employed on top of 
the convolutional layers. 
The CNN-$5$-w and CNN-$10$-w are the wide versions of CNN-$5$ and CNN-$10$, respectively. 
The main difference between the wide versions and their original counterparts is that the number of channels 
in the convolutional layers is four times larger ($256$ vs. $64$). 
The GRU model consists of two bidirectional recurrent layers with a hidden size of $64$.
Similar to the convolutional models, the GRU model also includes two fully connected layers and a final batch normalization layer 
at the top of the model. 

It is important to note that in the original work~\cite{guo2022deep}, 
the final batch normalization layer used a default parameter of $\epsilon = 1\mathrm{e}-5$ in $\mathrm{PyTorch}$. 
This setting did not affect the normality of the elements of the embedding vectors in their study. 
However, by using the learnable scaling factor and increasing the embedding dimension, the normality is compromised. 
We discovered that the value of $\epsilon = 1\mathrm{e}-5$ is too large, 
and the model learns features with the same order of magnitude as $\epsilon$. 
Consequently, the guarantee of the embedding elements following $N(0,1)$ was no longer valid. 
To address this issue, a smaller value of $\epsilon = 1\mathrm{e}-9$ for the final batch normalization layers 
is used in this research. 
This adjustment can be easily verified in the experiments. 
Since it is not our main focus, the details are not presented. 	

\section{The early stopping dimension for the engaged embedding networks} \label{app:esd}
In this appendix, we provide the sorted eigenvalues of $\mathrm{cov}(\bm{u_i}-\bm{u_j},\bm{u_i}-\bm{u_j}), (i\neq j)$ 
for different embedding dimensions. 
We also plot the approximation errors with respect to the embedding dimensions. 
These figures allow us to estimate the ESD values for the embedding networks CNN-$10$, CNN-$5$-w, CNN-$10$-w, and GRU. 

\begin{figure}[htb!]
	\centering
	% \subfigure[$n=40$]
	% {\includegraphics[width=0.24\textwidth, trim=23 5 23 20]{fig/40_True_cnn10.png}}
	\subfigure[$n=60$]
	{\includegraphics[width=0.32\linewidth]{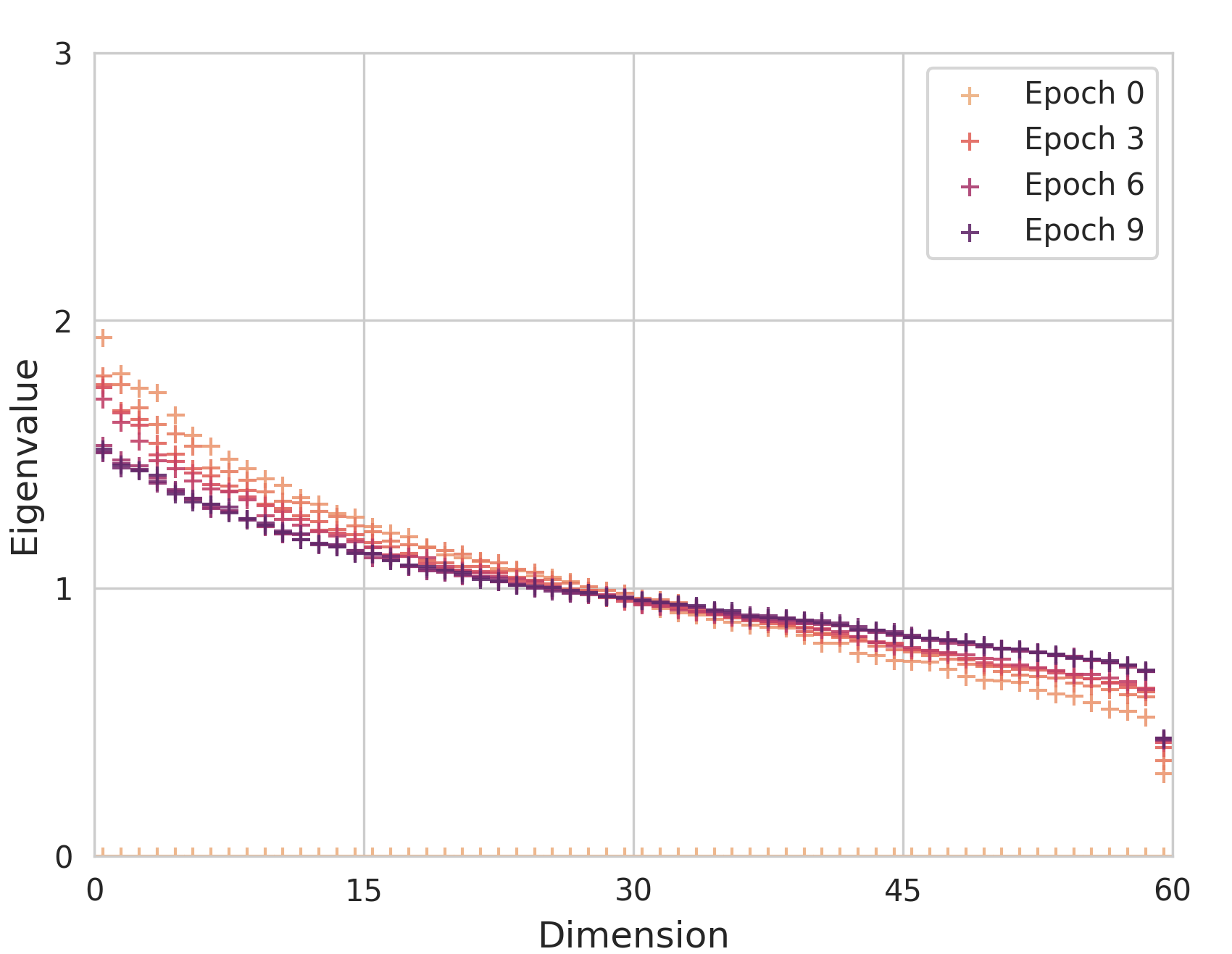}}
	\subfigure[$n=80$]
	{\includegraphics[width=0.32\linewidth]{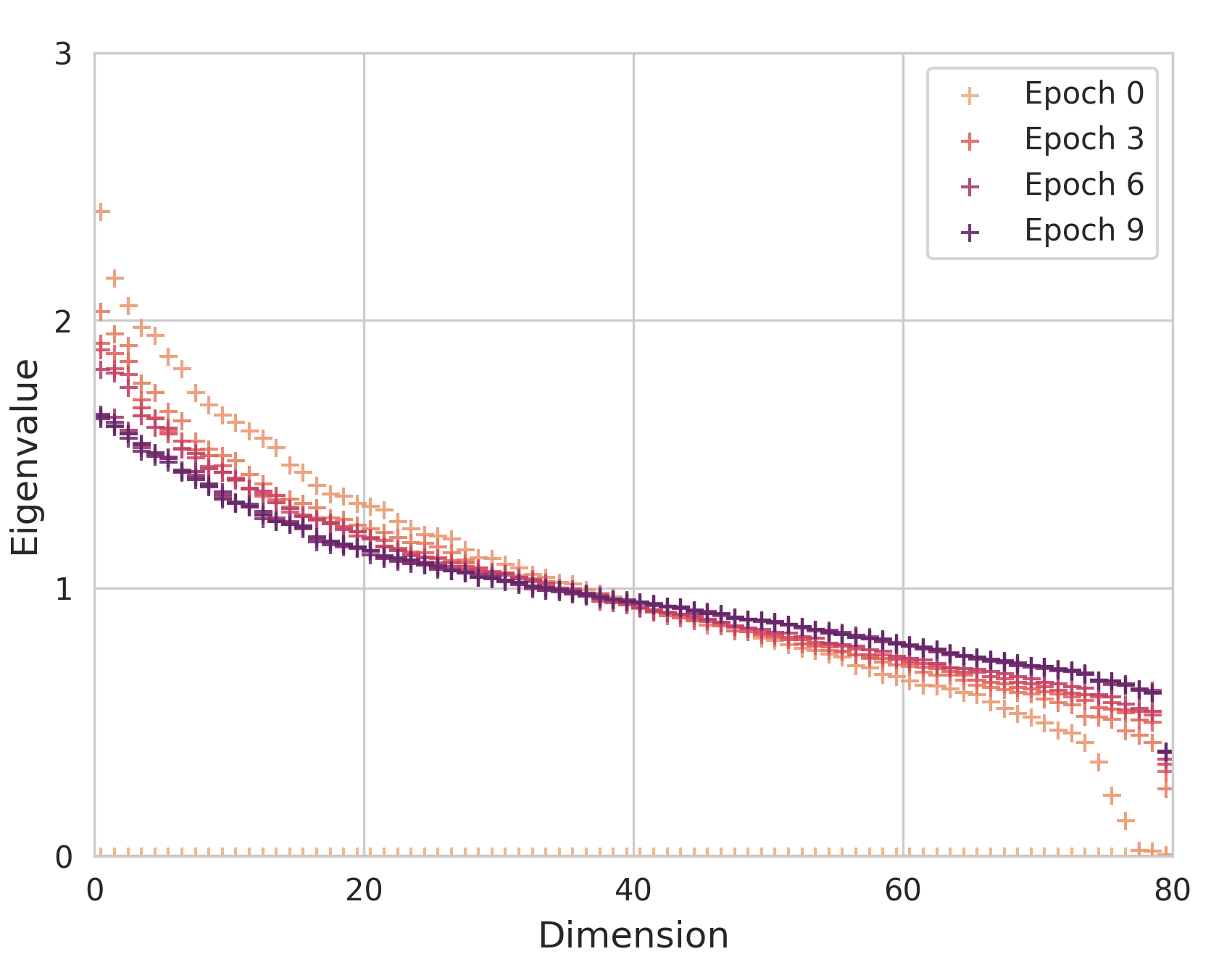}}
	\subfigure[$n=100$]
	{\includegraphics[width=0.32\linewidth]{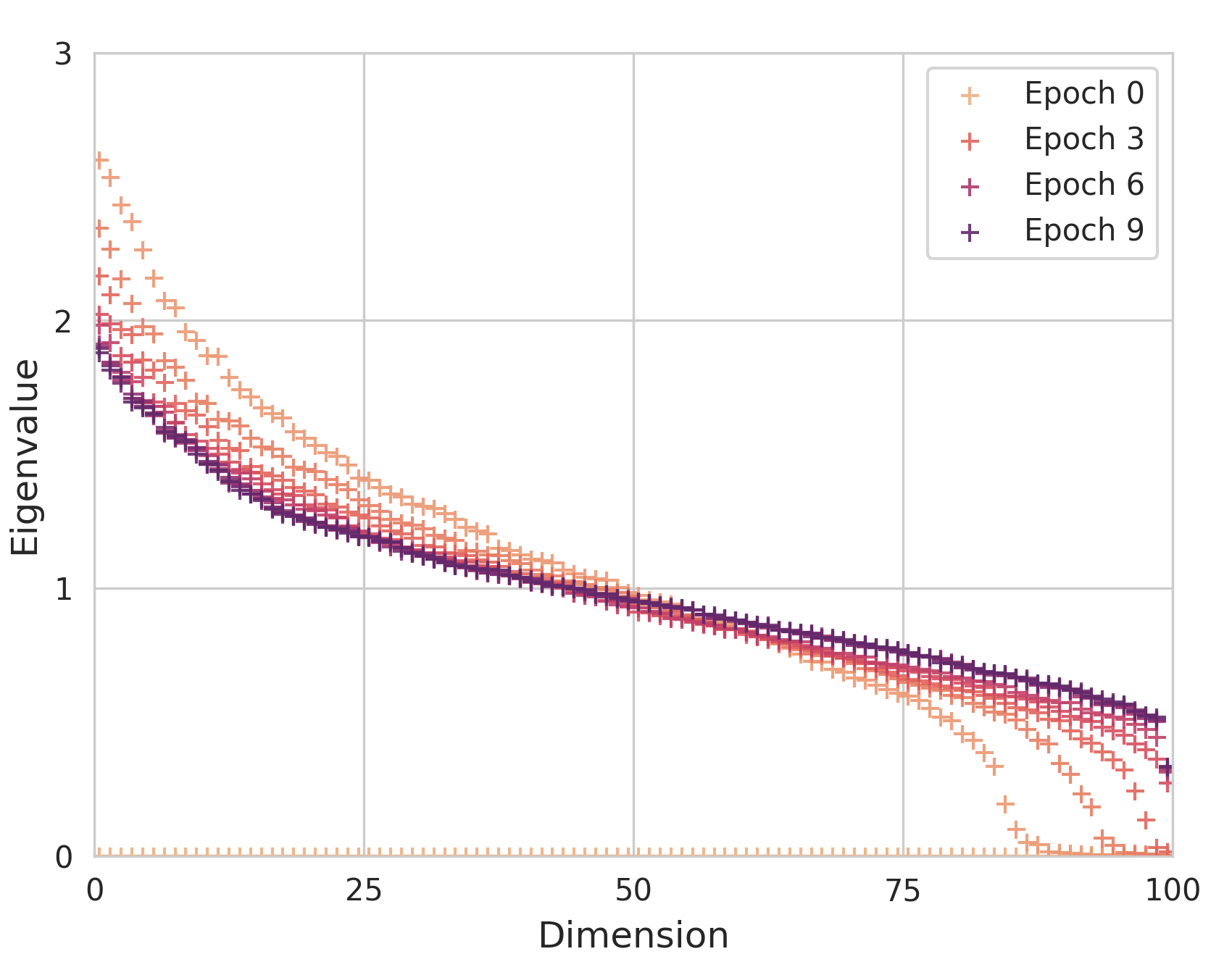}}
	\subfigure[$n=120$] 
	{\includegraphics[width=0.32\linewidth]{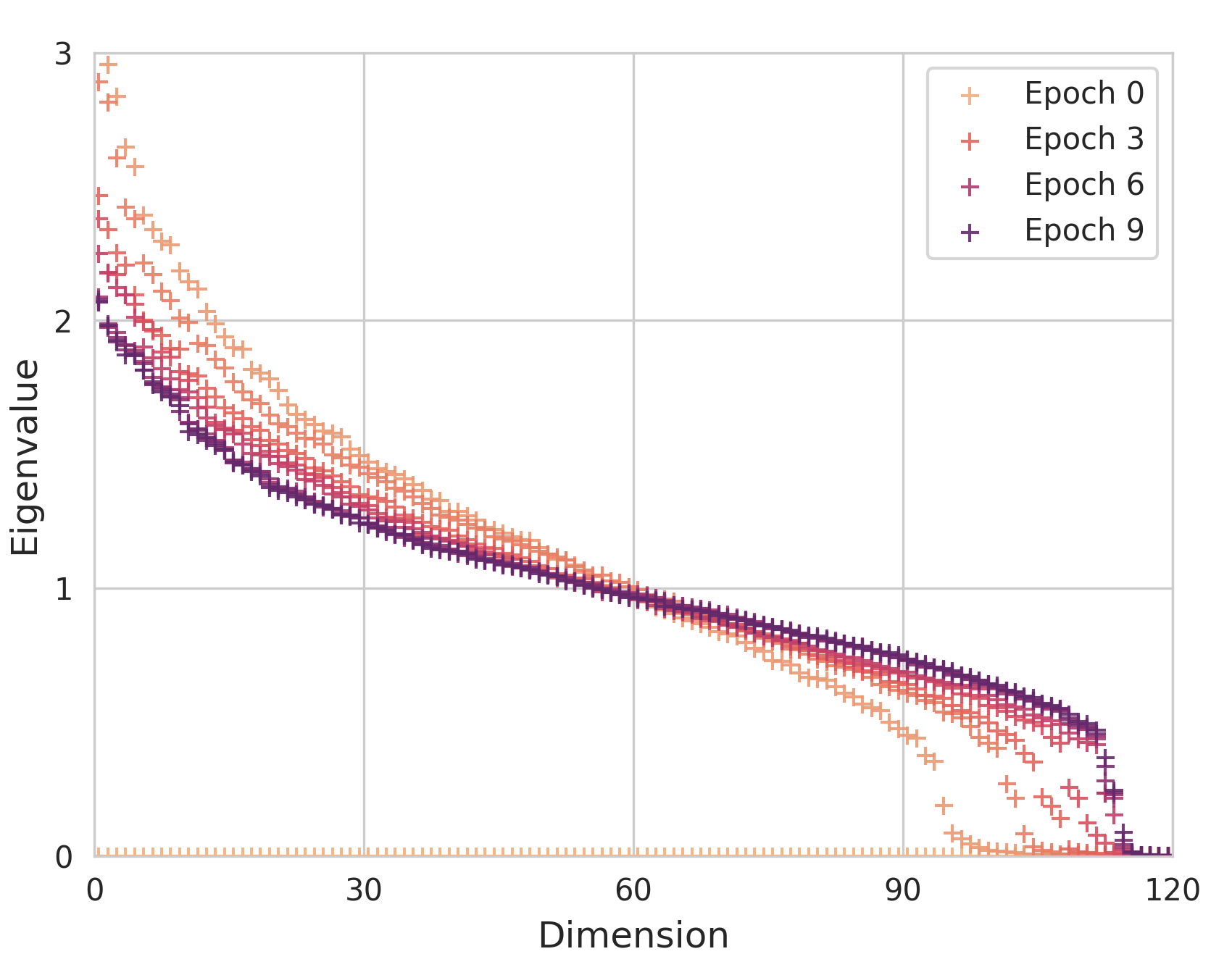}}
	\subfigure[$n=140$]
	{\includegraphics[width=0.32\linewidth]{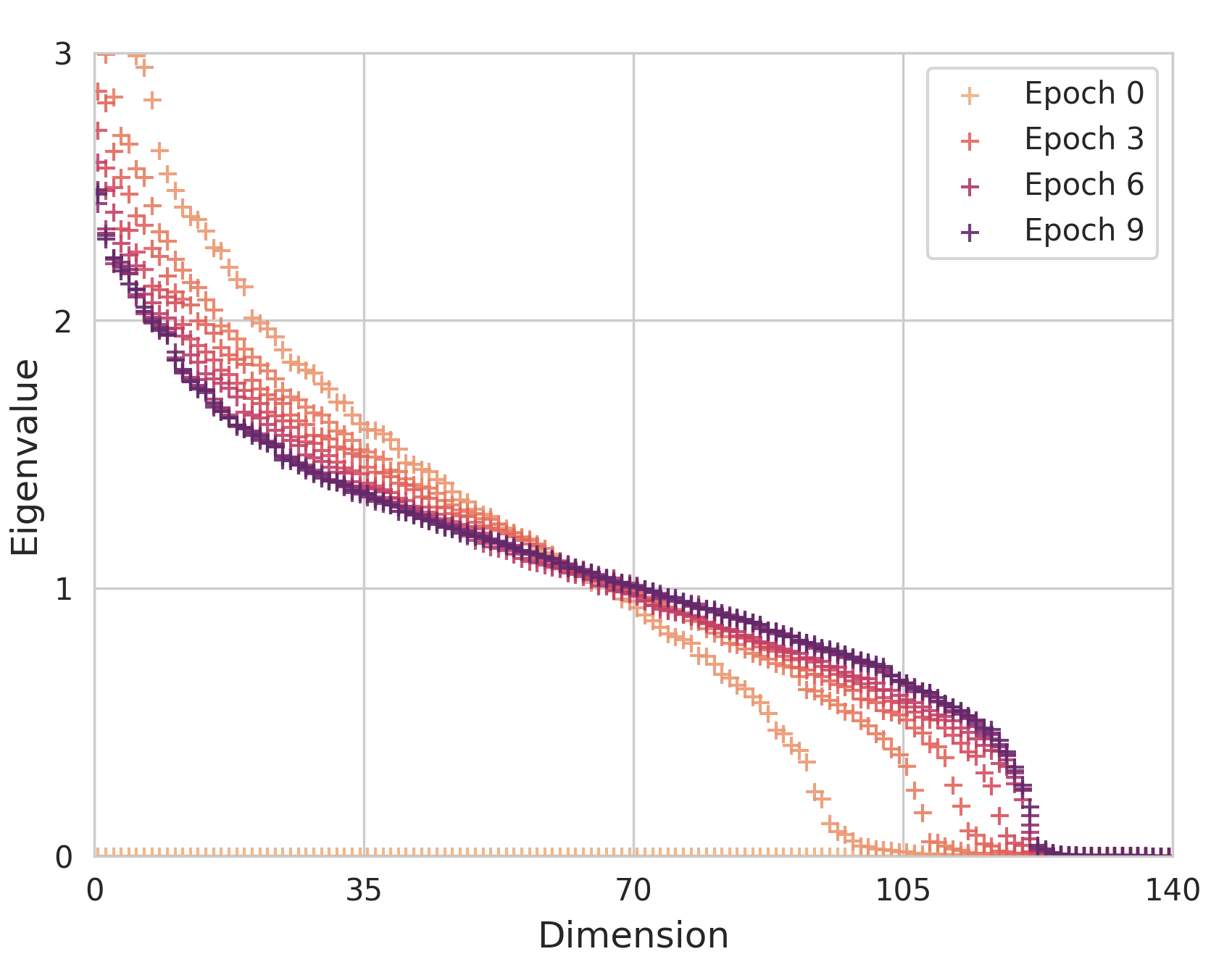}}
	\subfigure[$n=160$]
	{\includegraphics[width=0.32\linewidth]{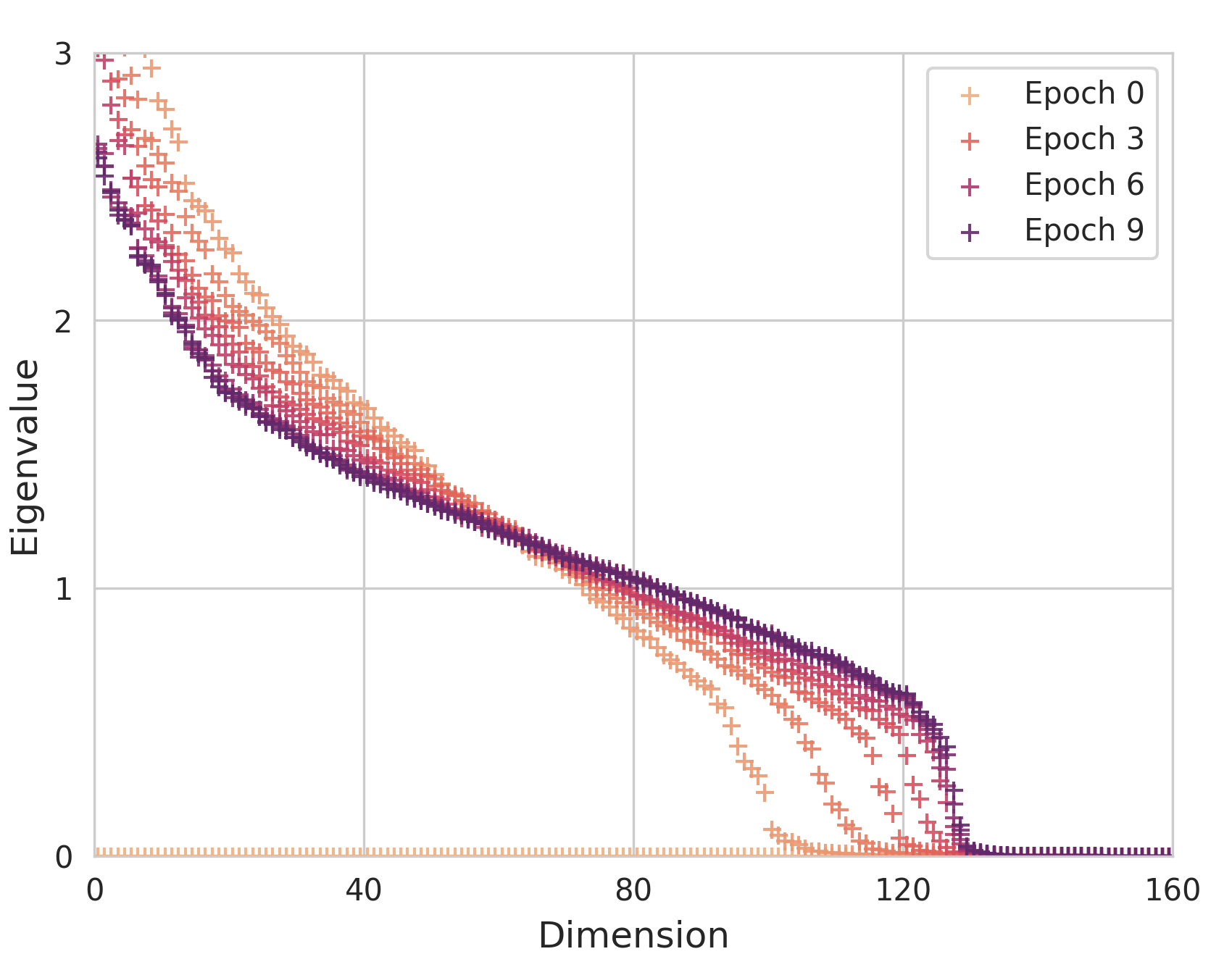}}
	% \subfigure[$n=180$]
	% {\includegraphics[width=0.24\textwidth, trim=23 5 23 20]{fig/180_True_cnn10.png}}
	\caption{\textbf{CNN-$10$}. The sorted eigenvalues of $\mathrm{cov}(\bm{u_i}-\bm{u_j},\bm{u_i}-\bm{u_j}), (i\neq j)$ 
		are plotted for different choices of the embedding dimension $n$ for the CNN-$10$ embedding network. 
		When $n$ is small, the eigenvalues are distributed around $1$, as in (a)--(c). 
		Increasing the embedding dimension $n$, the sorted eigenvalues decrease to $0$ after some dimension, as in (d)--(f).
	}
	\label{fig:esd-cnn10}
\end{figure}
\begin{figure}[htb!]
	\centering
	\subfigure[AE$_g$]
	{\includegraphics[width=0.47\linewidth]{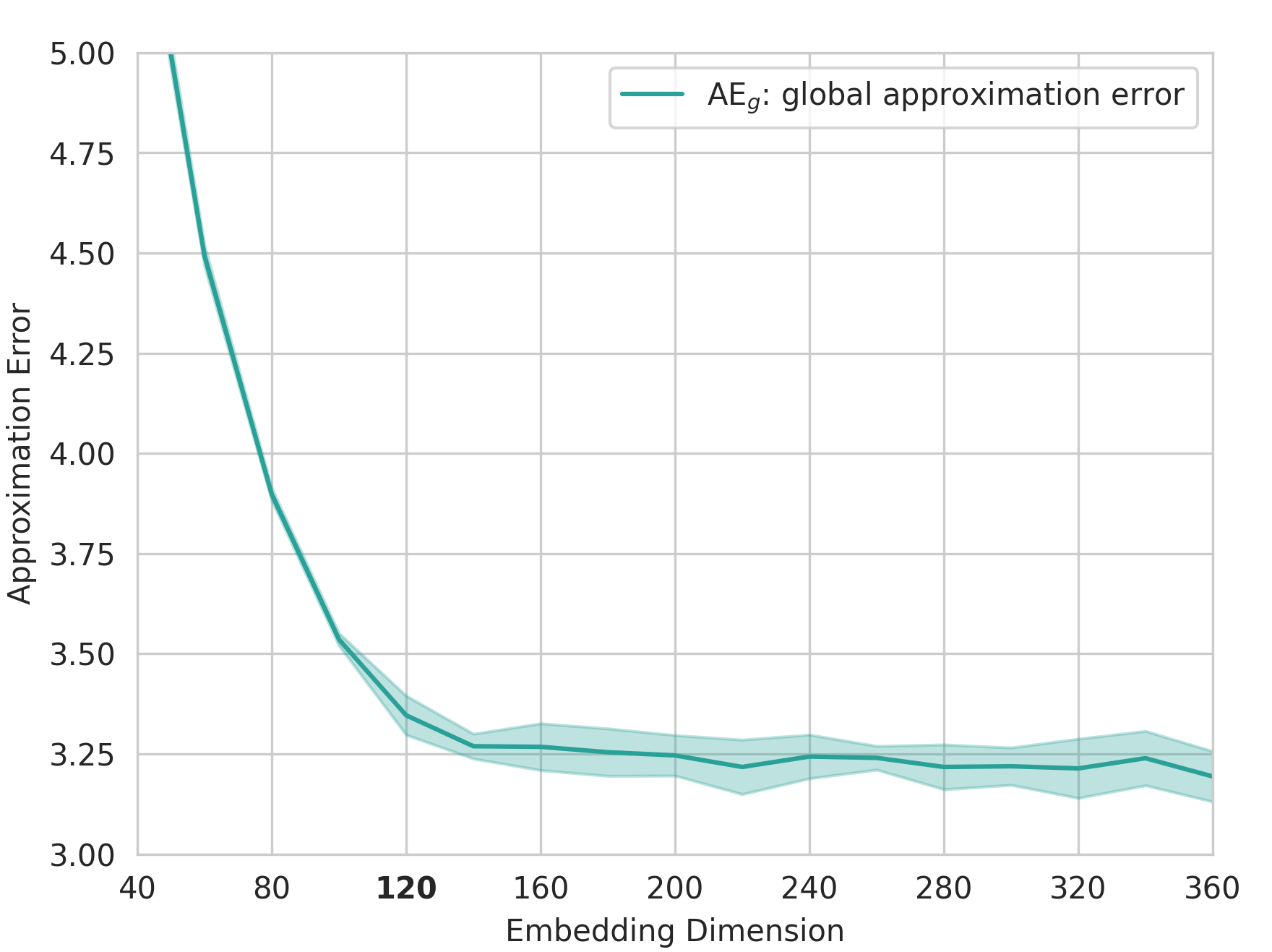}}
	\subfigure[AE$_h$]
	{\includegraphics[width=0.47\linewidth]{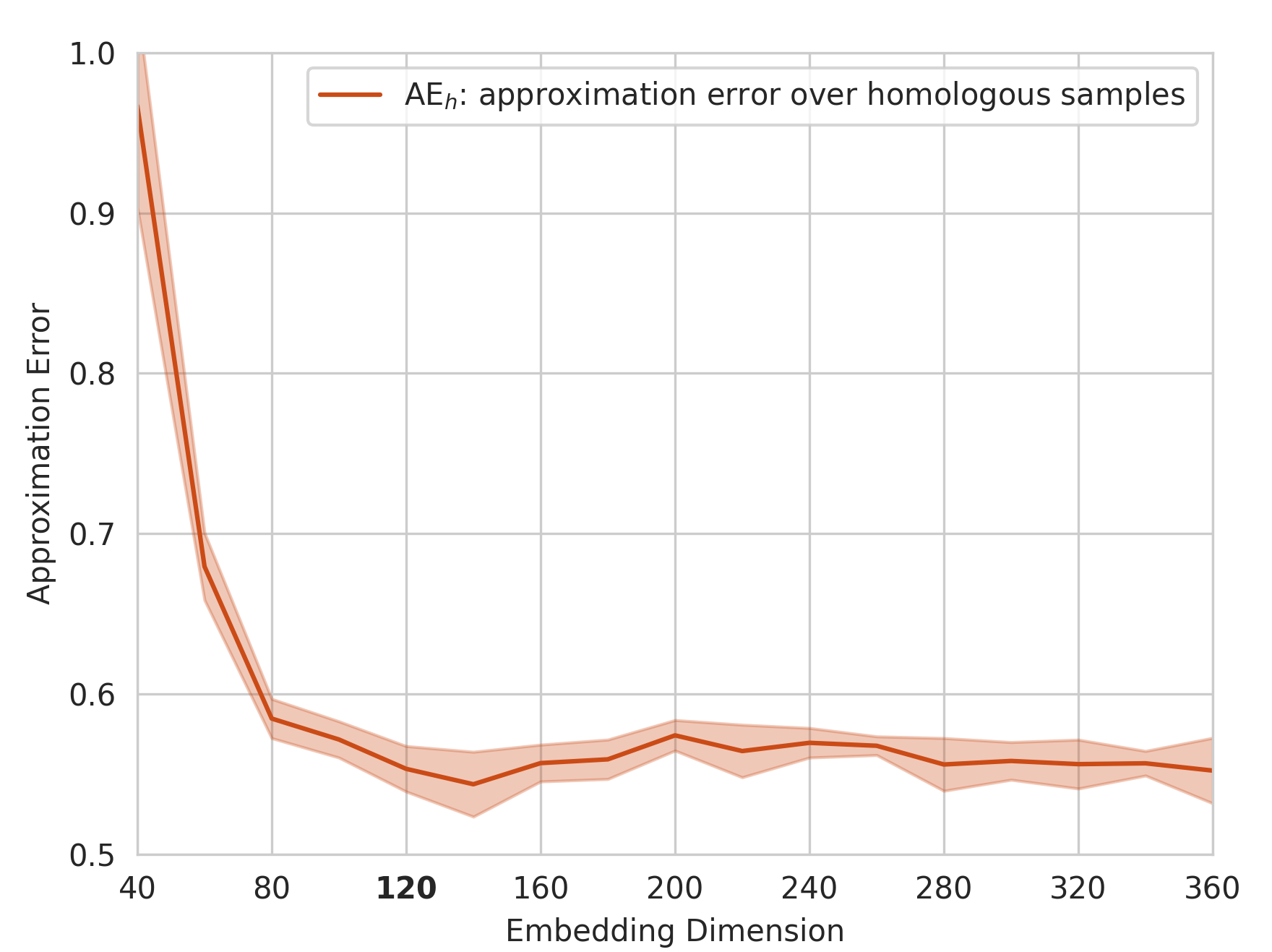}}
	\caption{\textbf{CNN-$10$}. 
	The AE$_g$ and AE$_h$ are shown against the embedding dimension in (a) and (b), respectively. 
	The curves are plotted based on the mean and standard deviation over 5 runs. 
	The approximation errors decrease alongwith increase the embedding dimension $n$ until the ESD $n_0$, 
	which is $120$ for CNN-$10$. 
	When $n>n_0$, there is no gain of the performance on a larger $n$.}
	\label{fig:esd-performance-cnn10}
\end{figure}

\textbf{CNN-$10$}. For the embedding network CNN-$10$, the corresponding figures, namely \cref{fig:esd-cnn10,fig:esd-performance-cnn10}, 
exhibit similar patterns to those observed for CNN-$5$ in \cref{fig:esd,fig:esd-performance}. 
The estimated ESD is about $n_0=120$. 
The approximation error decreases as the embedding dimension increases until reaching the ESD $n_0$. 
The only difference between CNN-$10$ and CNN-$5$ is that the embedding network of CNN-$10$ demonstrates a lower
standard deviation across different runs. 
Hence, we may infer that the embedding network of CNN-$10$ is more stable during the training phase compared to CNN-$5$. 

\begin{figure}[htb!]
	\centering
	% \subfigure[$n=40$]
	% {\includegraphics[width=0.24\textwidth, trim=23 5 23 20]{fig/40_True_fatcnn5.png}}
	% \subfigure[$n=60$]
	% {\includegraphics[width=0.32\linewidth, trim=23 5 23 20]{fig/60_True_fatcnn5.png}}
	\subfigure[$n=80$]
	{\includegraphics[width=0.32\linewidth,]{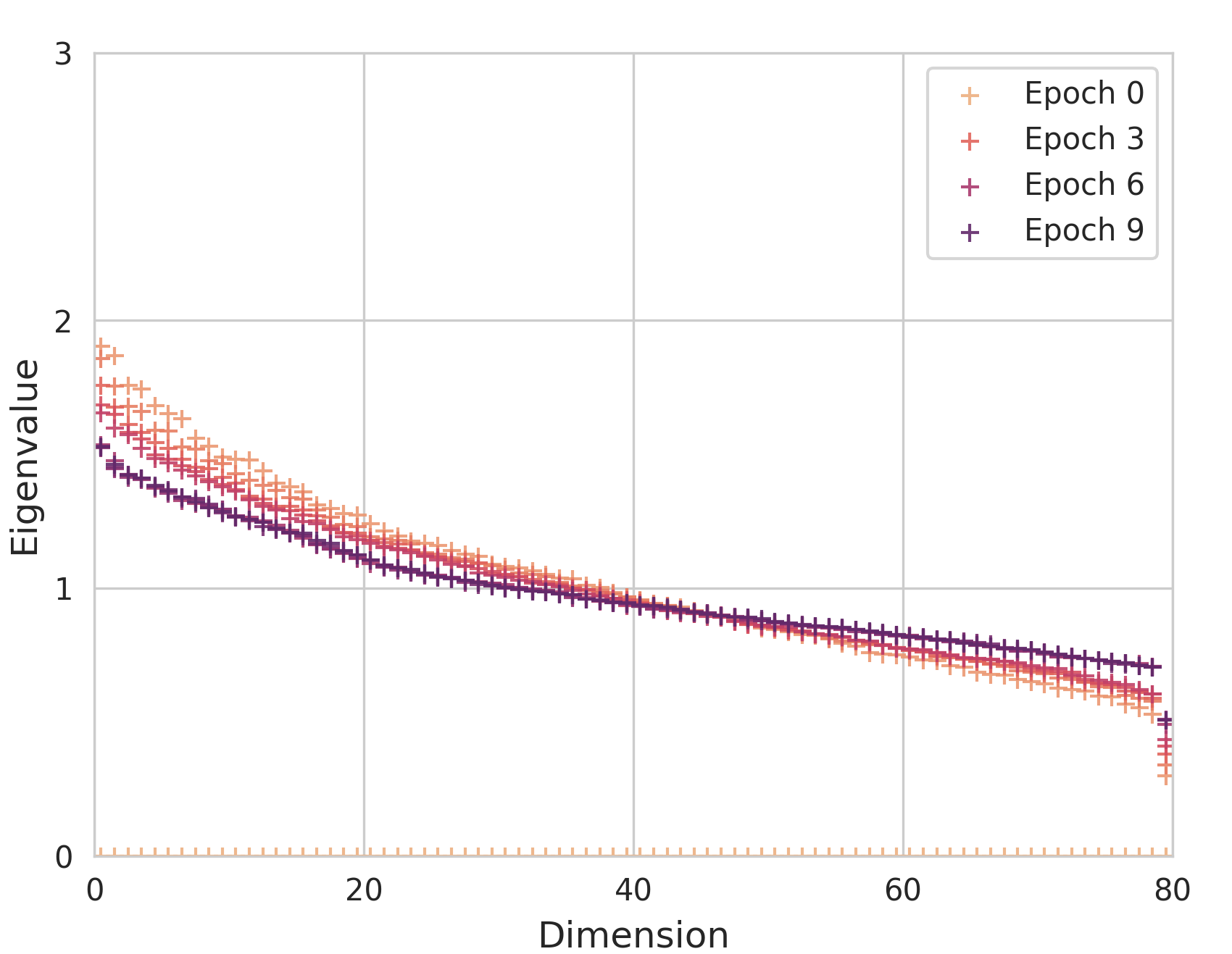}}
	\subfigure[$n=100$]
	{\includegraphics[width=0.32\linewidth,]{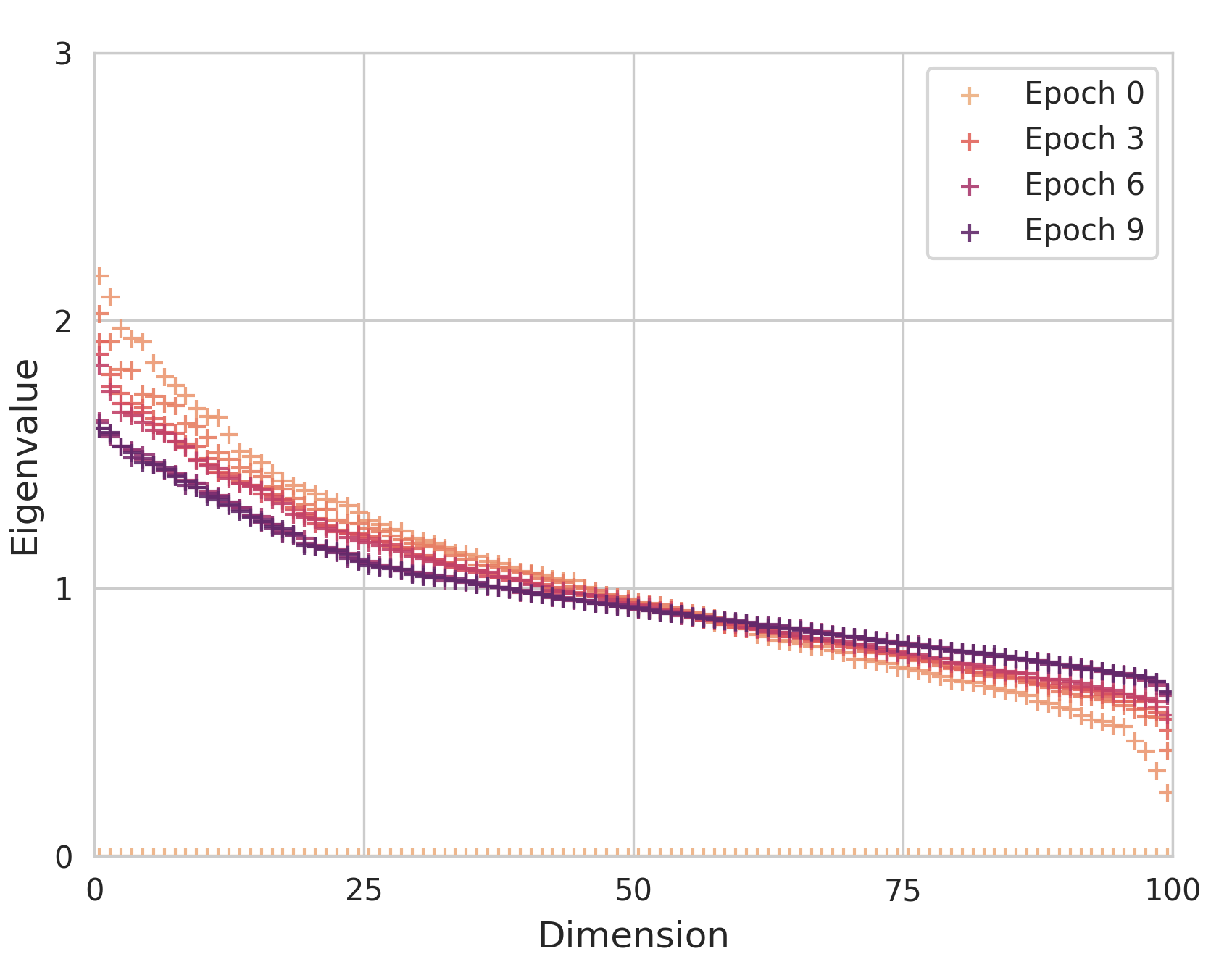}}
	\subfigure[$n=120$] 
	{\includegraphics[width=0.32\linewidth,]{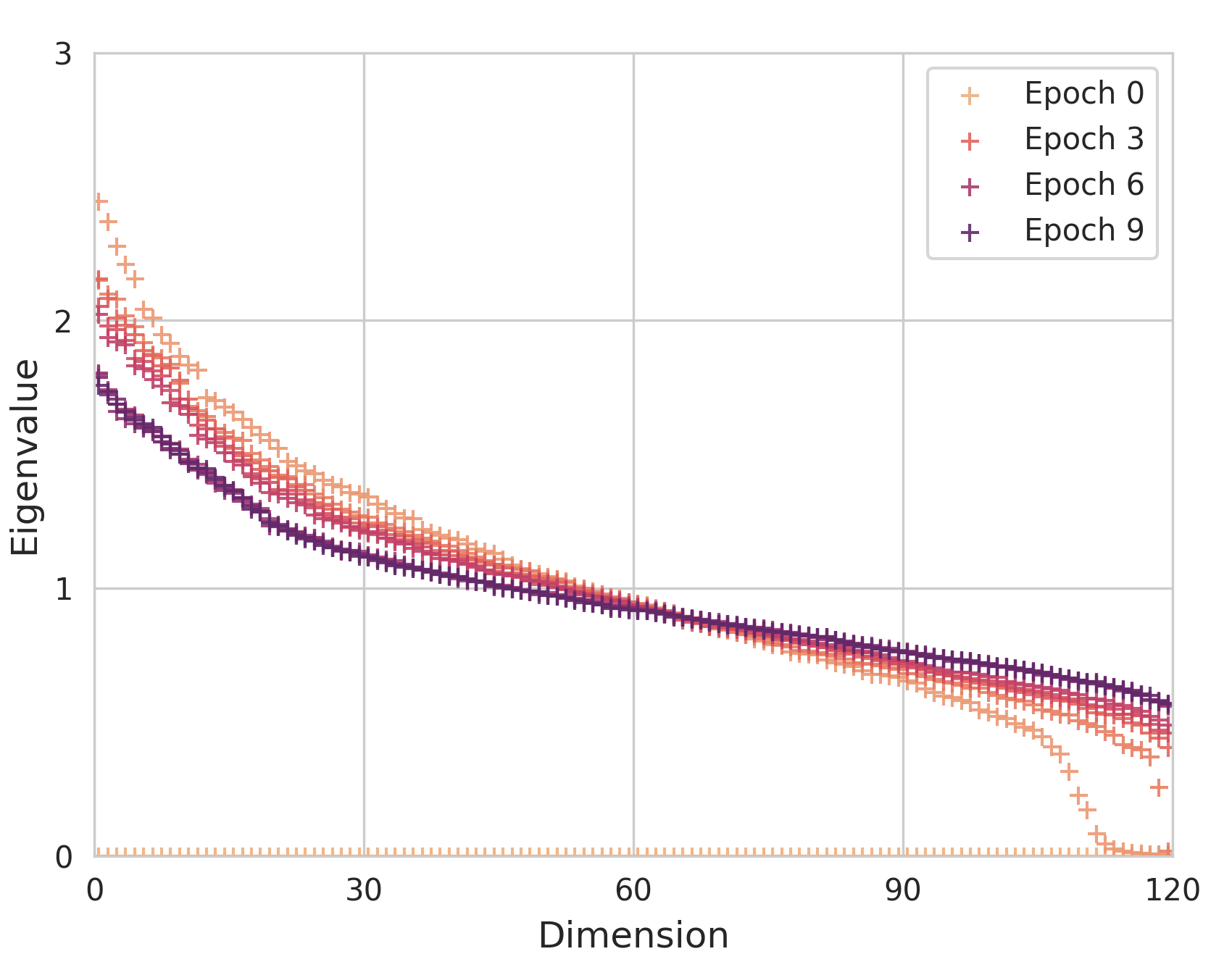}}
	\subfigure[$n=140$]
	{\includegraphics[width=0.32\linewidth,]{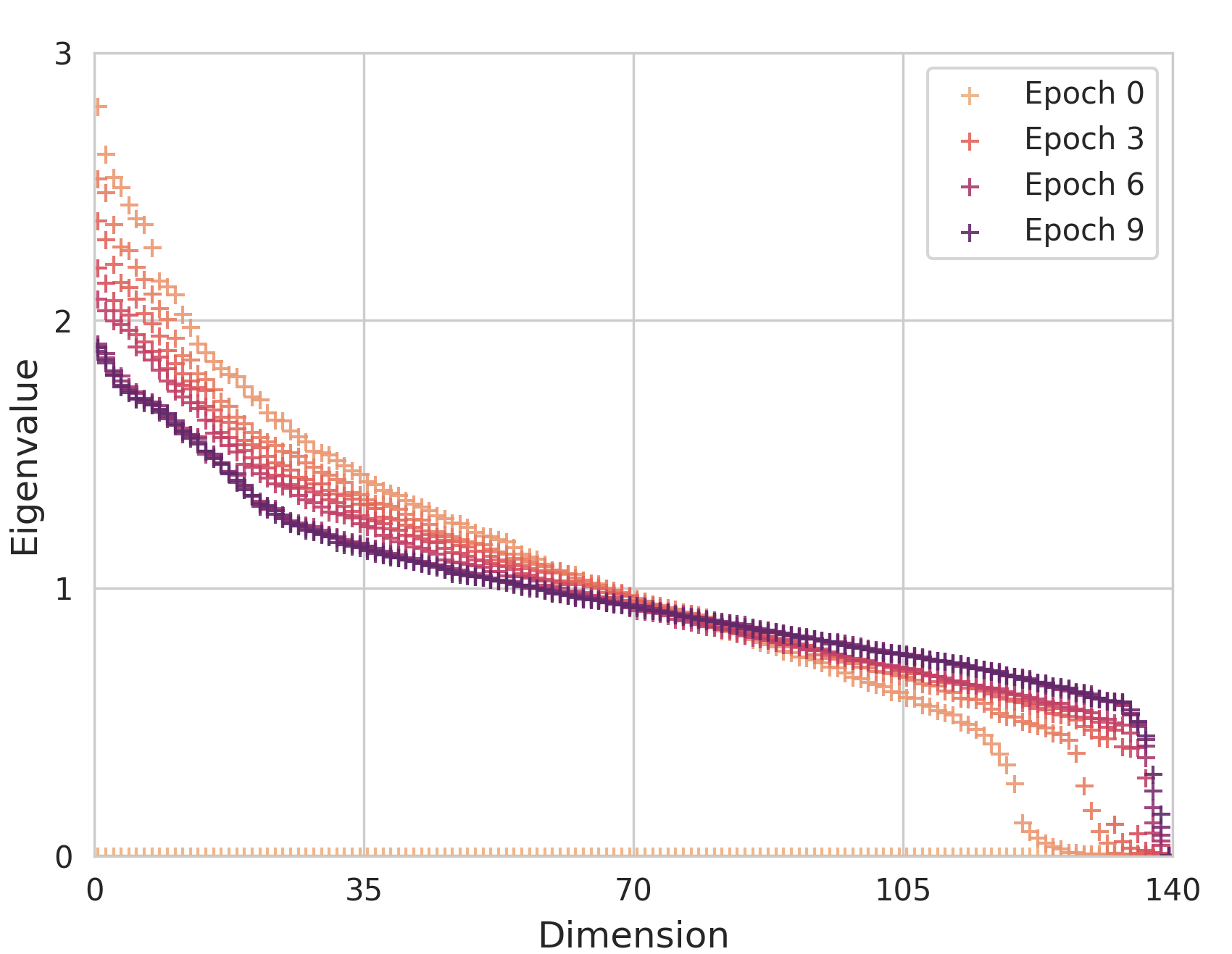}}
	\subfigure[$n=160$]
	{\includegraphics[width=0.32\linewidth,]{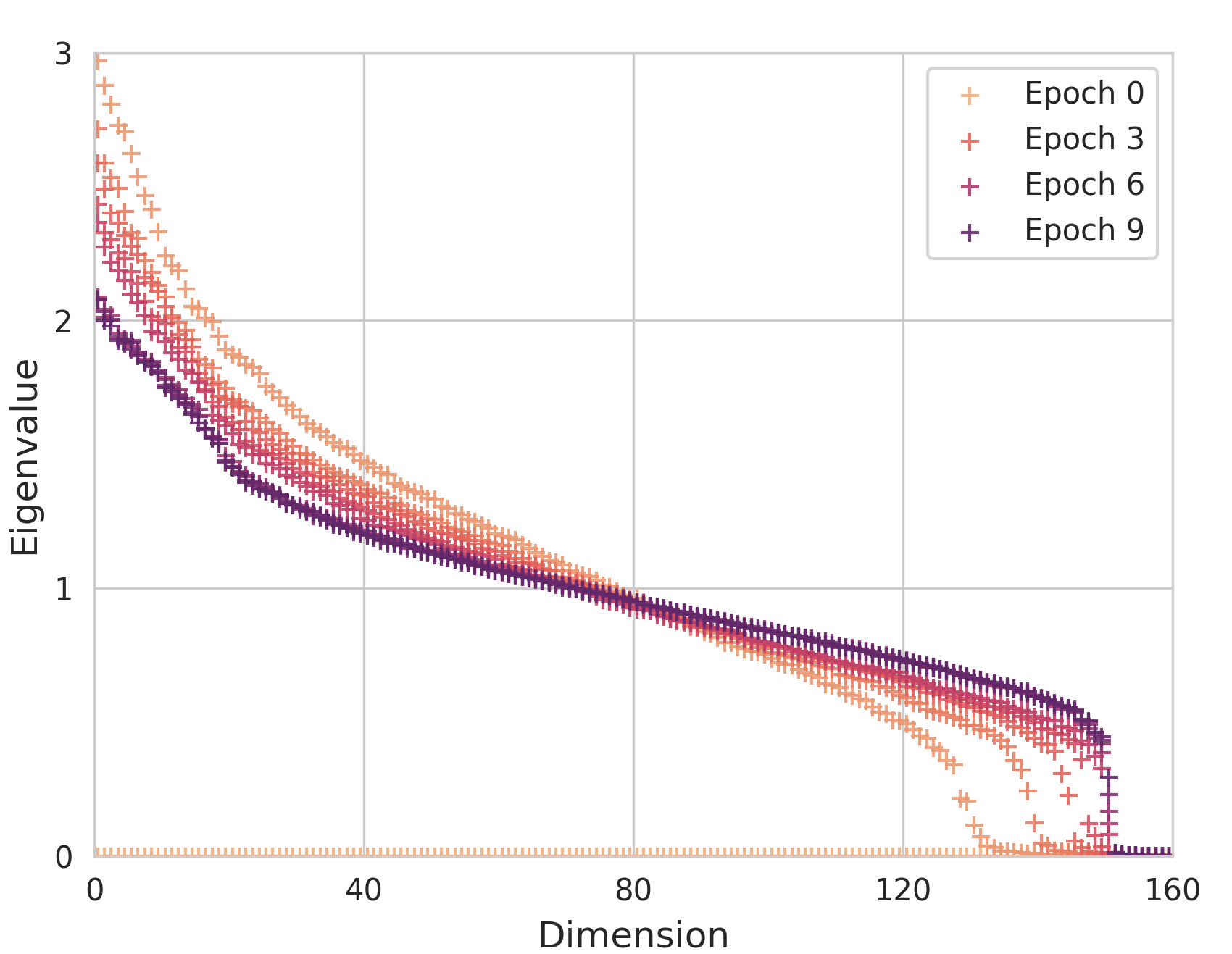}}
	\subfigure[$n=180$]
	{\includegraphics[width=0.32\linewidth,]{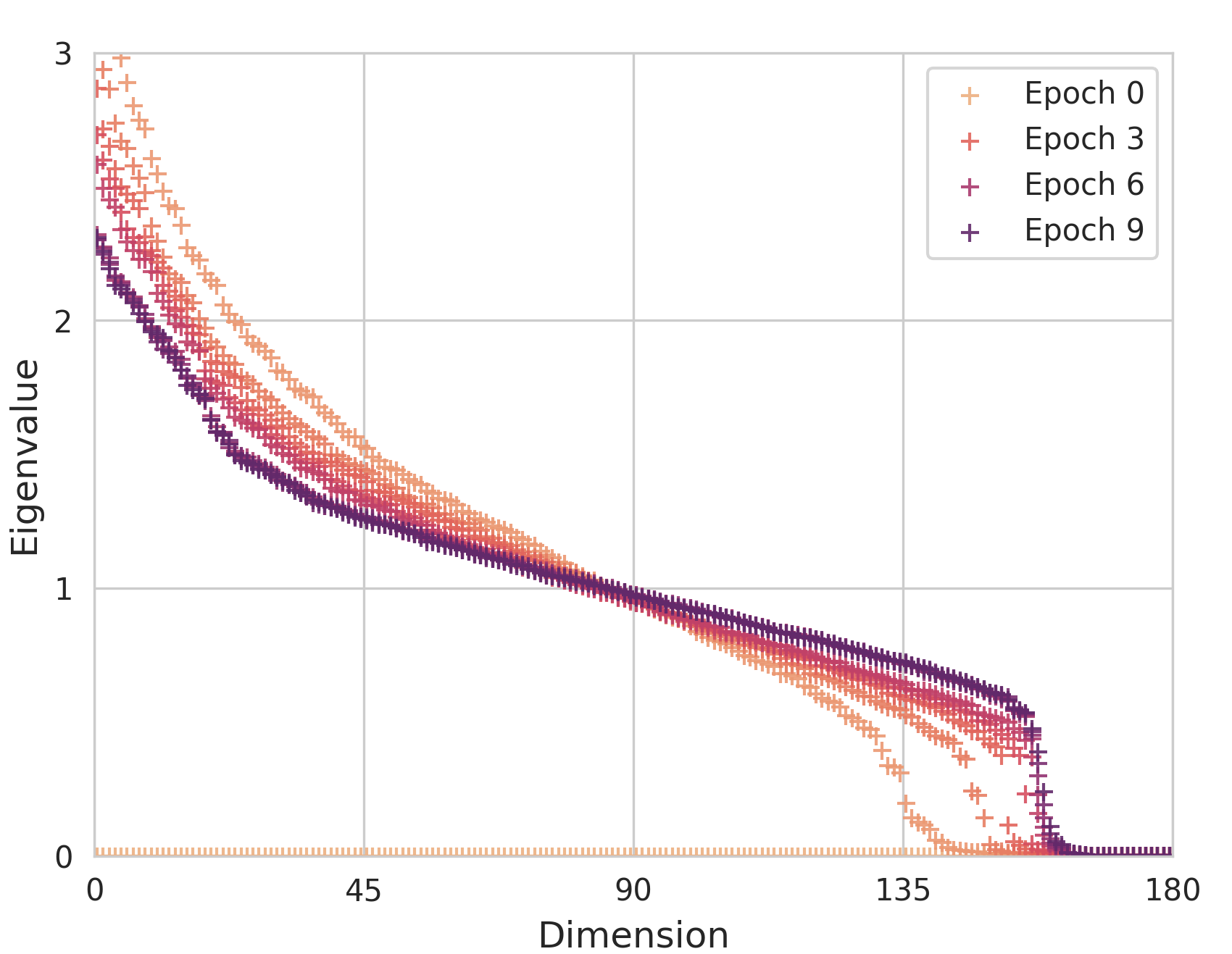}}
	\caption{\textbf{CNN-$5$-w}. The sorted eigenvalues of $\mathrm{cov}(\bm{u_i}-\bm{u_j},\bm{u_i}-\bm{u_j}), (i\neq j)$ 
		are plotted for different choices of the embedding dimension $n$ for the CNN-$5$-w embedding network. 
		When $n$ is small, the eigenvalues are distributed around $1$, as in (a)--(c). 
		Increasing the embedding dimension $n$, the sorted eigenvalues decrease to $0$ after some dimension, as in (d)--(f).
	}
	\label{fig:esd-fatcnn5}
\end{figure} 
\begin{figure}[htb!]
	\centering
	\subfigure[AE$_g$]
	{\includegraphics[width=0.47\linewidth]{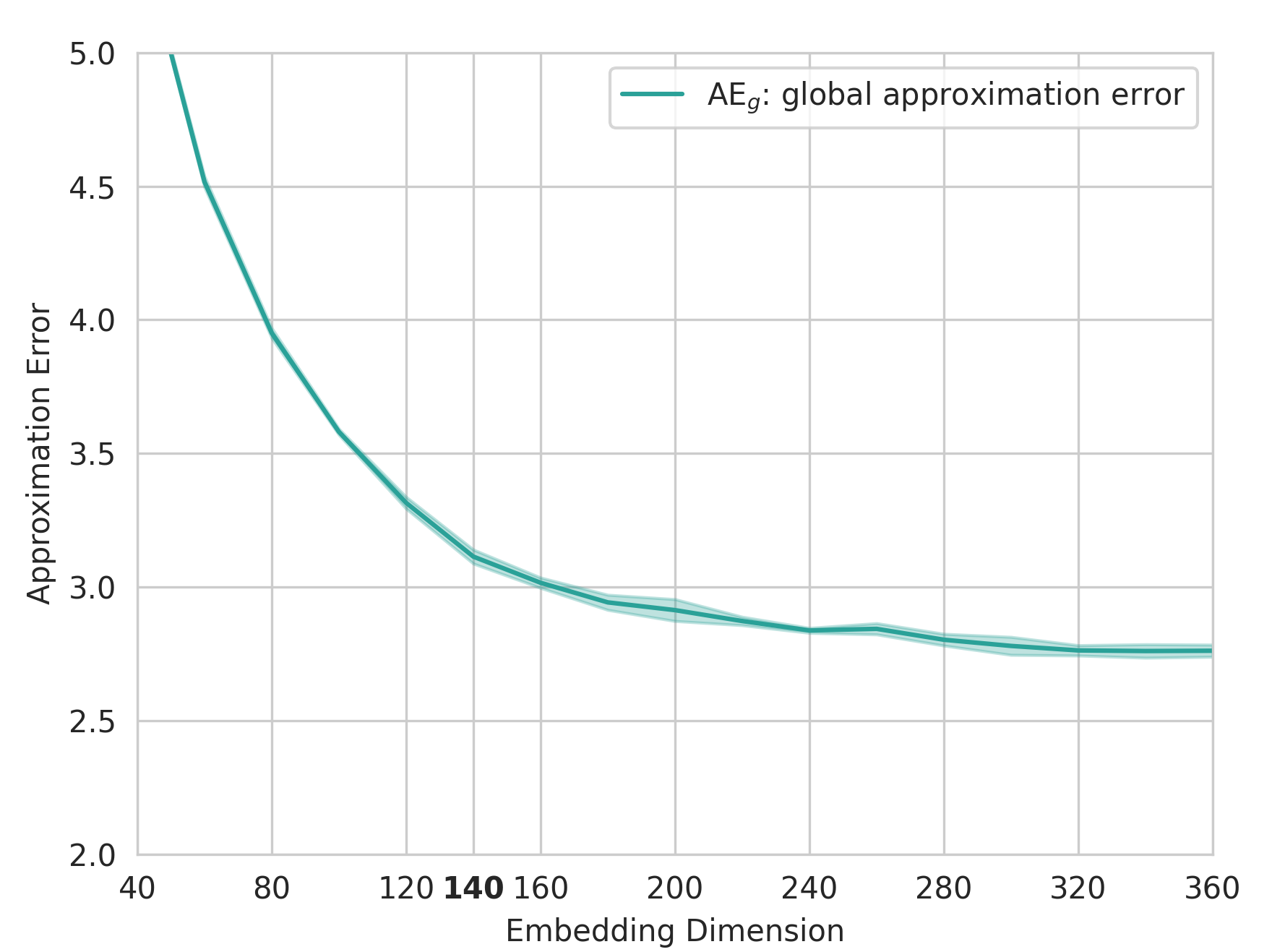}}
	\subfigure[AE$_h$]
	{\includegraphics[width=0.47\linewidth]{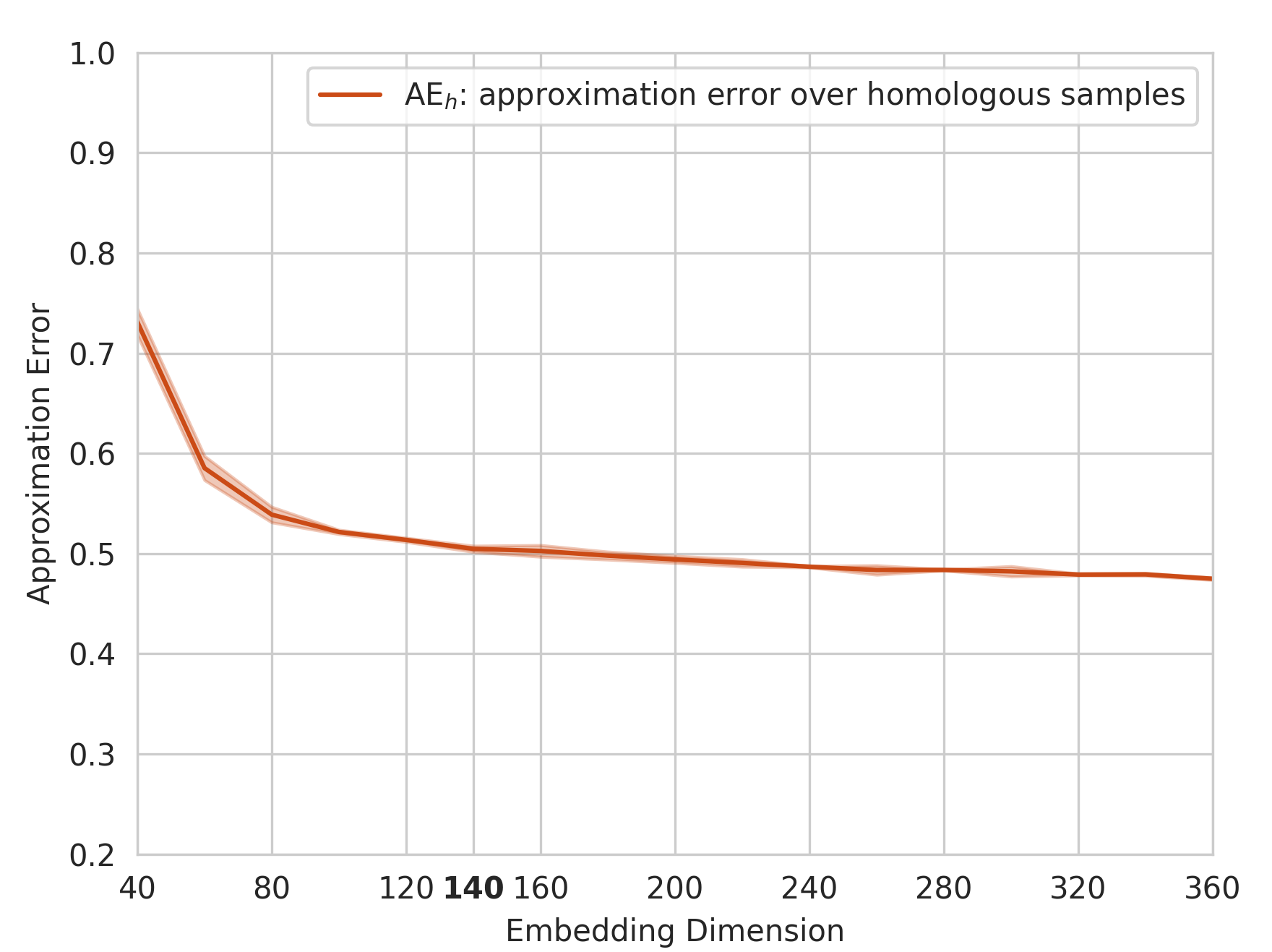}}
	\caption{\textbf{CNN-$5$-w}. The AE$_g$ and AE$_h$ are shown against the embedding dimension in (a) and (b), respectively. 
	The curves are plotted based on the mean and standard deviation over 5 runs. 
		The approximation errors decrease alongwith increase the embedding dimension $n$ until the ESD $n_0$, 
		which is $140$ for CNN-$5$-w. 
		When $n>n_0$, the improvement in performance with a larger $n$ is not significant, particularly for AE$_h$.}
	\label{fig:esd-performance-fatcnn5}
\end{figure}

\textbf{CNN-$5$-w}. The results obtained from CNN-$5$-w are depicted in \cref{fig:esd-fatcnn5,fig:esd-performance-fatcnn5}. 
In comparison to \cref{fig:esd}, it is observed that the zero eigenvalues of 
$\mathrm{cov}(\bm{u_i}-\bm{u_j},\bm{u_i}-\bm{u_j}), (i\neq j)$ 
starts to appear when the embedding dimension reaches $140$ rather than $120$. 
Based on this, the estimated ESD is about $n_0=140$. 
In \cref{fig:esd-performance-fatcnn5}, the AE$_g$ continues to decrease slightly after $n_0$, 
while the AE$_h$ changes negligibly after $n_0$. 

\begin{figure}[htb!]
	\centering
	% \subfigure[$n=40$]
	% {\includegraphics[width=0.24\textwidth, trim=23 5 23 20]{fig/40_True_fatcnn10.png}}
	% \subfigure[$n=60$]
	% {\includegraphics[width=0.24\textwidth, trim=23 5 23 20]{fig/60_True_fatcnn10.png}}
	\subfigure[$n=80$]
	{\includegraphics[width=0.32\linewidth]{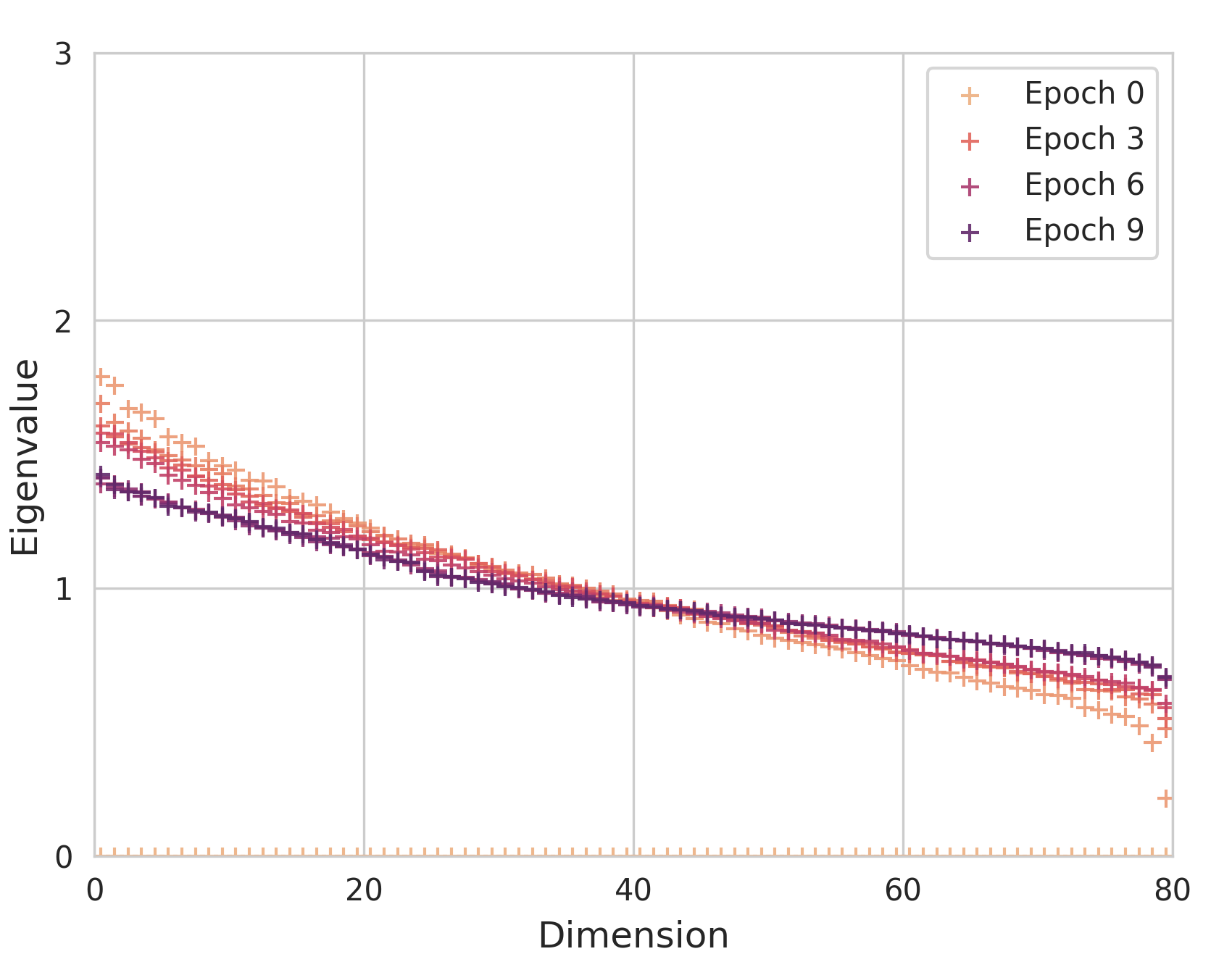}}
	\subfigure[$n=100$]
	{\includegraphics[width=0.32\linewidth]{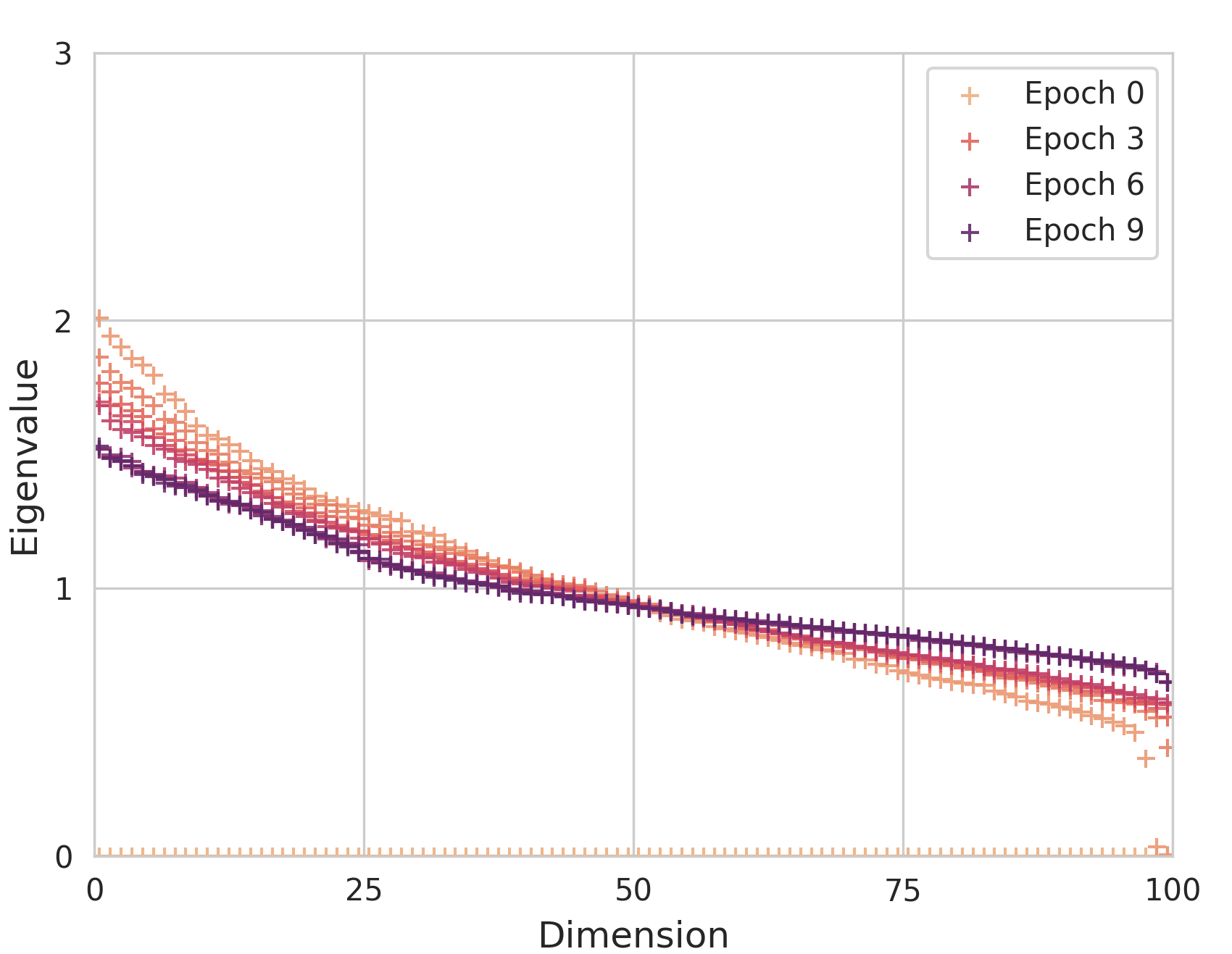}}
	\subfigure[$n=120$] 
	{\includegraphics[width=0.32\linewidth]{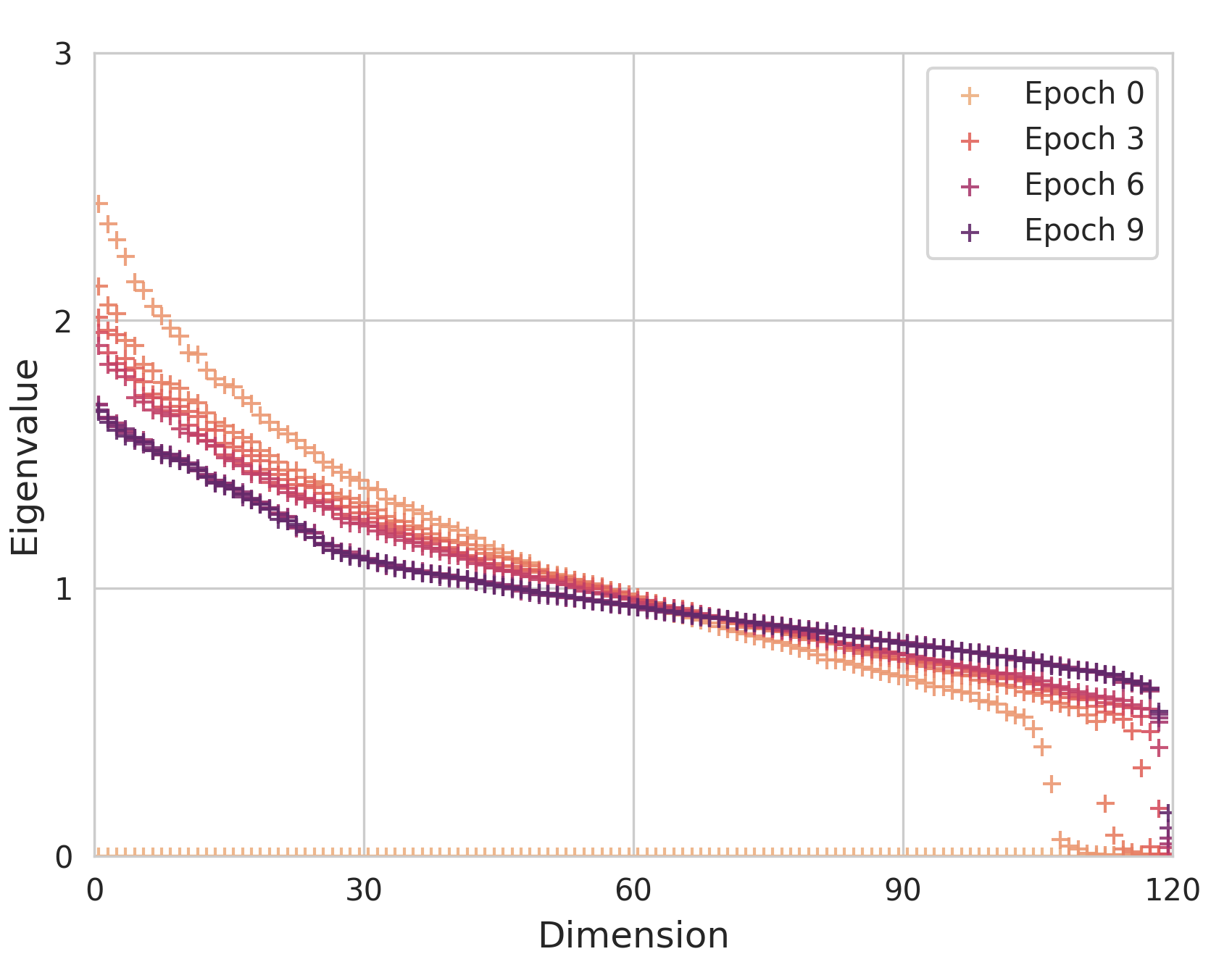}}
	\subfigure[$n=140$]
	{\includegraphics[width=0.32\linewidth]{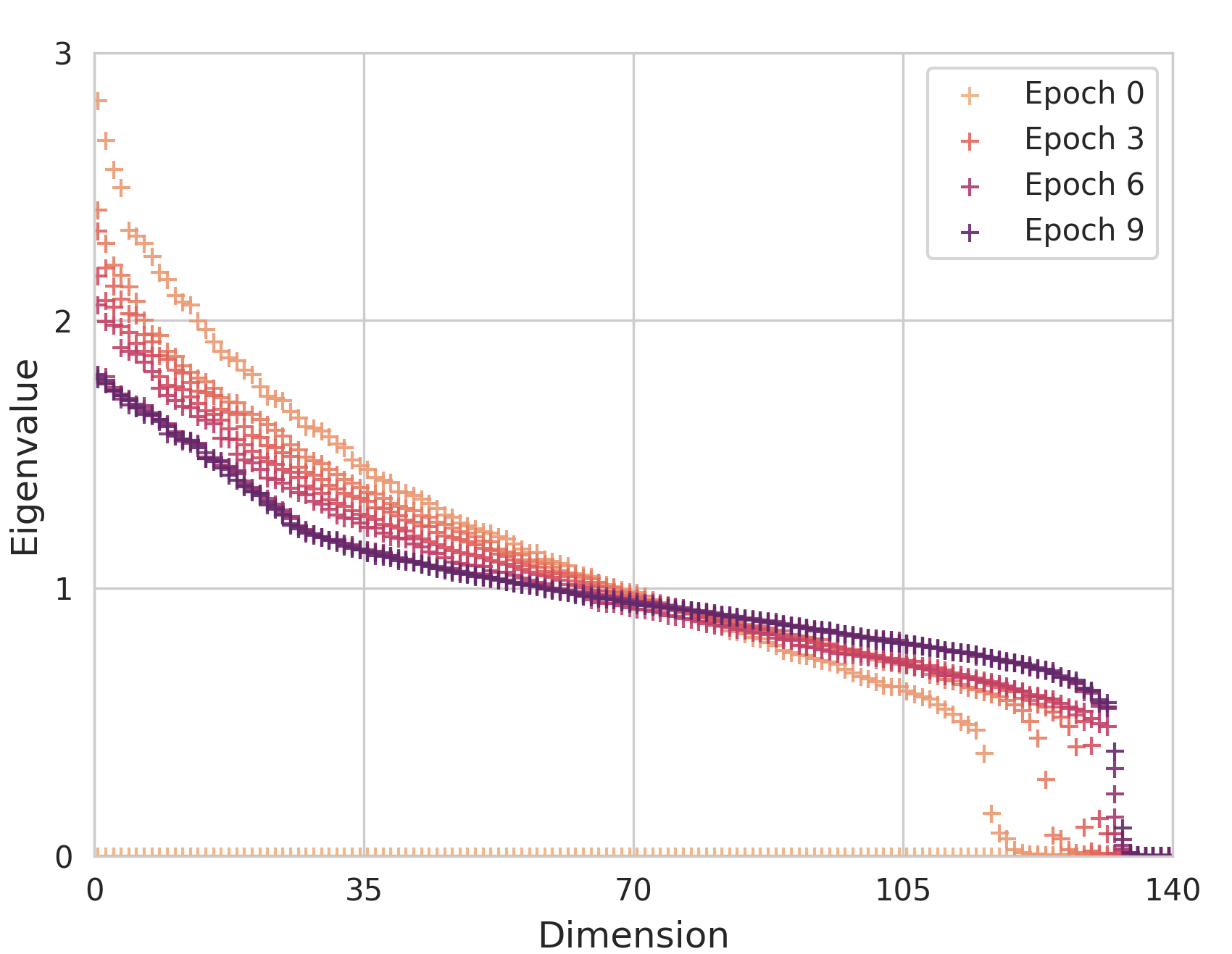}}
	\subfigure[$n=160$]
	{\includegraphics[width=0.32\linewidth]{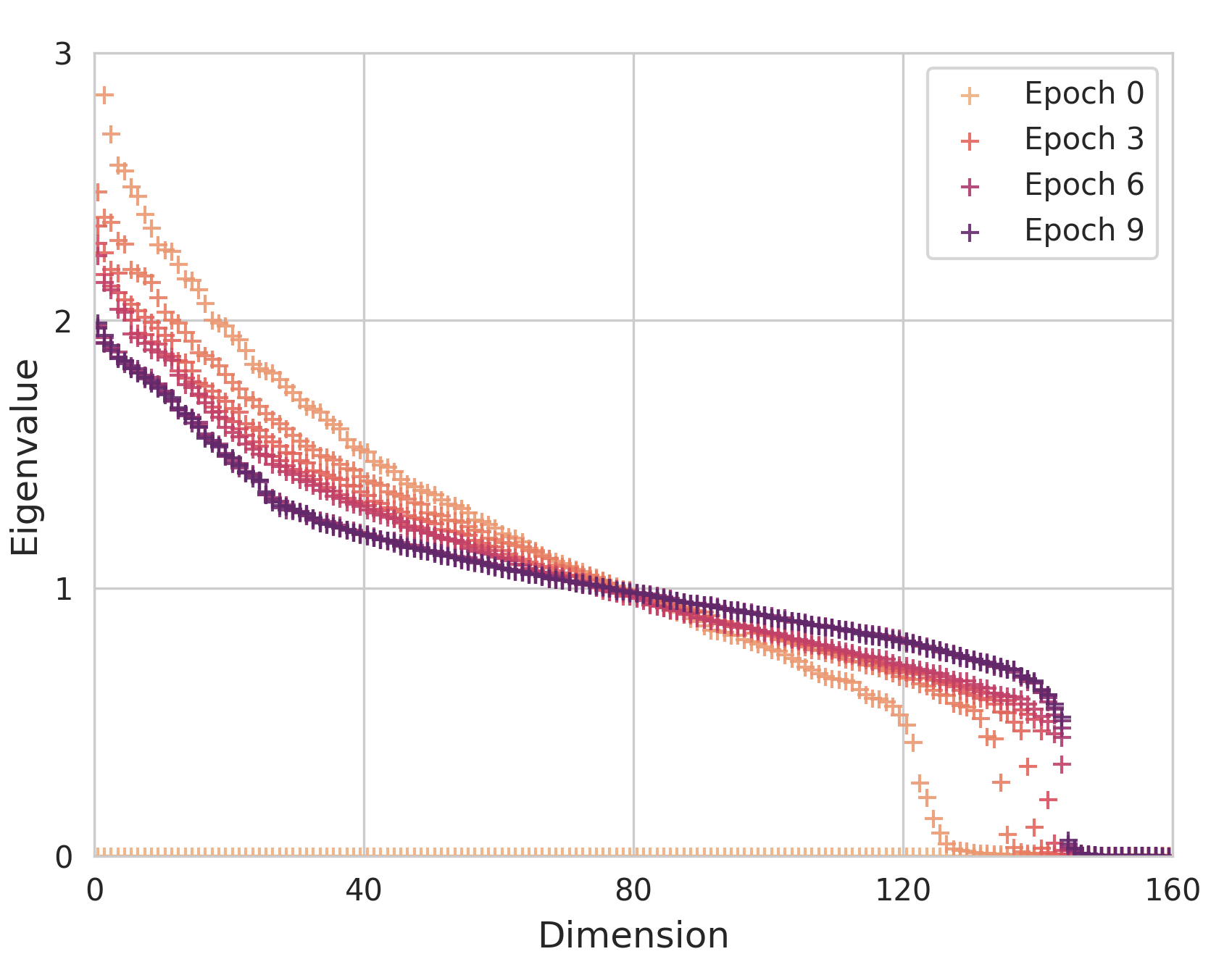}}
	\subfigure[$n=180$]
	{\includegraphics[width=0.32\linewidth]{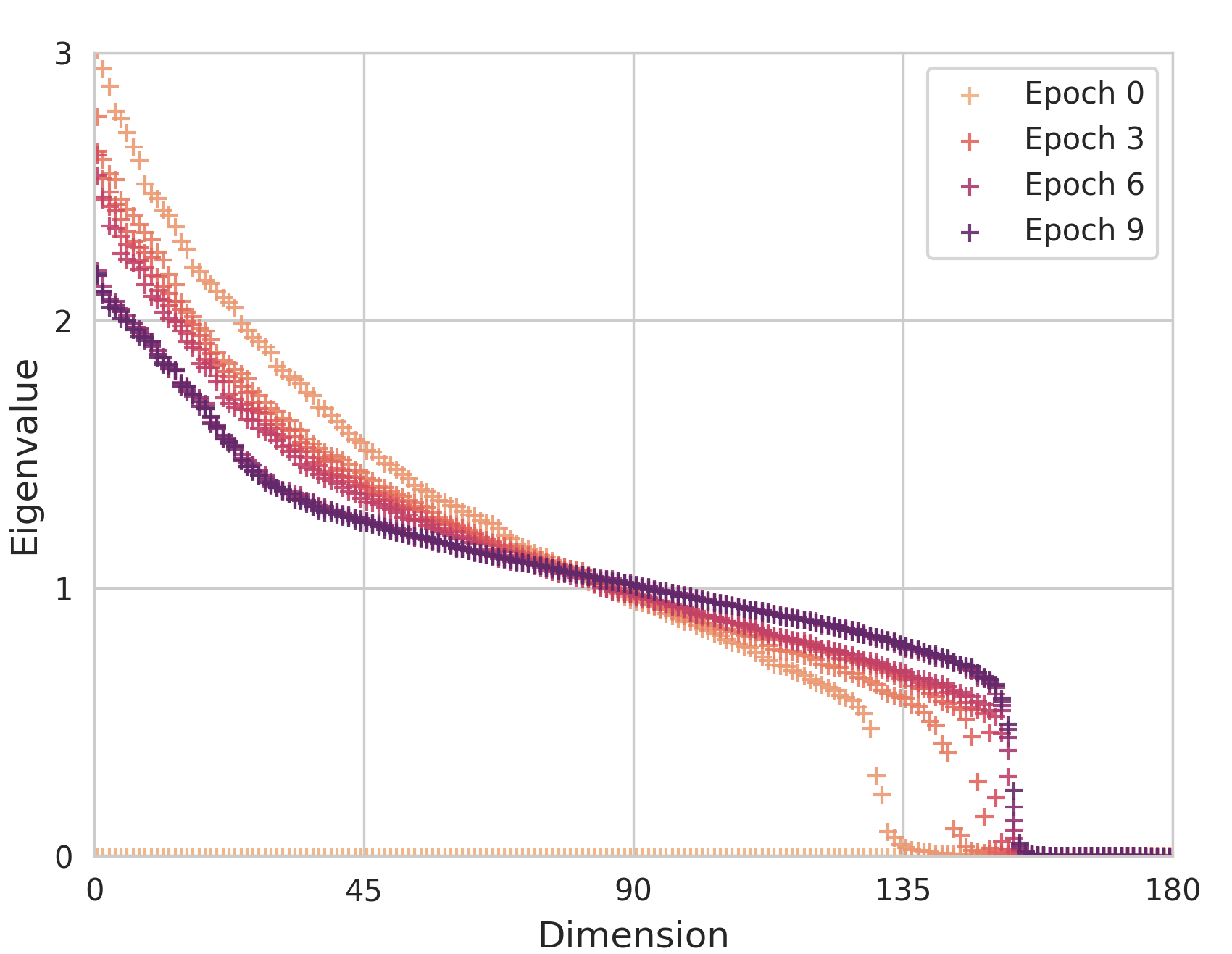}}
	\caption{\textbf{CNN-$10$-w}. The sorted eigenvalues of $\mathrm{cov}(\bm{u_i}-\bm{u_j},\bm{u_i}-\bm{u_j}), (i\neq j)$ 
		are plotted for different choices of the embedding dimension $n$ for the CNN-$10$-w embedding network. 
		When $n$ is small, the eigenvalues are distributed around $1$, as in (a)--(c). 
		Increasing the embedding dimension $n$, the sorted eigenvalues decrease to $0$ after some dimension, as in (d)--(f).
	}
	\label{fig:esd-fatcnn10}
\end{figure}

\begin{figure}[htb!]
	\centering
	\subfigure[AE$_g$]
	{\includegraphics[width=0.47\linewidth]{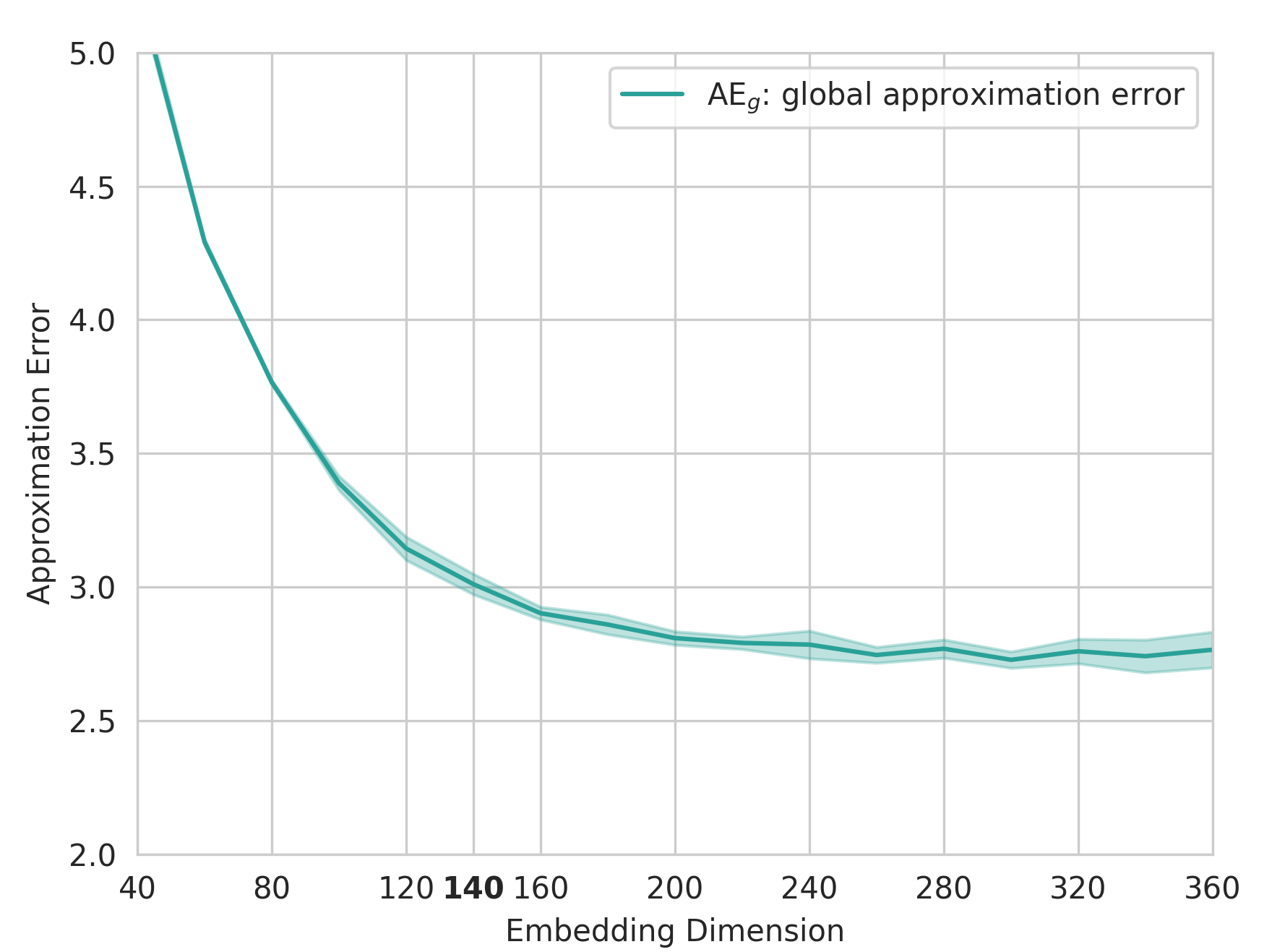}}
	\subfigure[AE$_h$]
	{\includegraphics[width=0.47\linewidth]{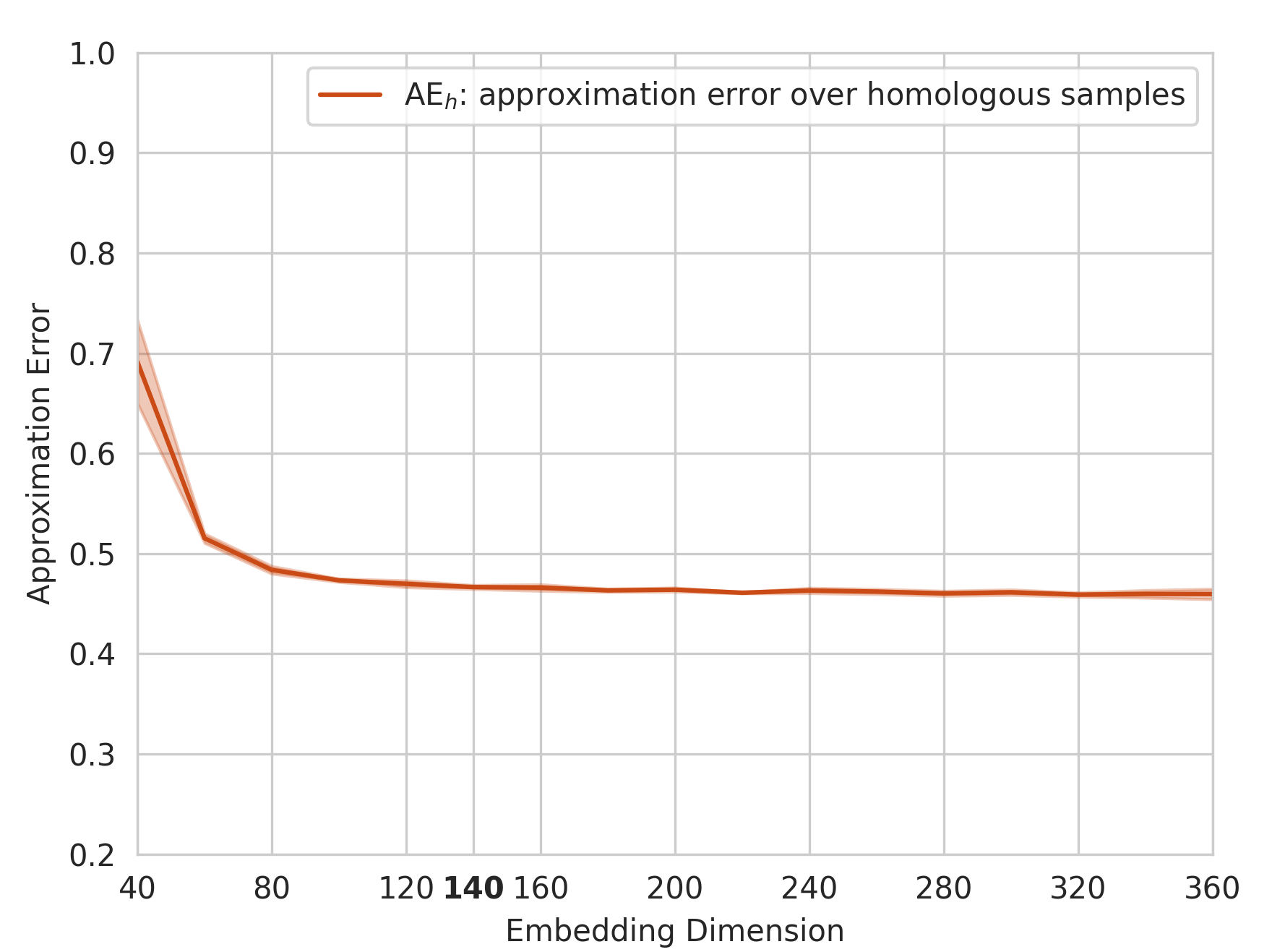}}
	\caption{\textbf{CNN-$10$-w}. The global approximation error and homologous approximation error are shown against the embedding dimension in (a) and (b), respectively. 
	The curves are plotted based on the mean and standard deviation over 5 runs. 
		The approximation errors decrease alongwith increase the embedding dimension $n$ until the ESD $n_0$, 
		which is $140$ for CNN-$10$-w. 
		When $n>n_0$, the improvement in performance with a larger $n$ is not significant, particularly for AE$_h$.}
	\label{fig:esd-performance-fatcnn10}
\end{figure}

\textbf{CNN-$10$-w}. As suggested in \cref{fig:esd-fatcnn10,fig:esd-performance-fatcnn10}, 
the estimated ESD for the CNN-$10$-w engaged in this paper 
is approximately $n_0 = 140$. 
Similar to the difference between CNN-$5$ and CNN-$5$-w, 
the ESD $n_0$ of embedding network CNN-$10$-w exceeds that of the plain CNN-$10$.  

\begin{figure}[htb!]
	\centering
	% \subfigure[$n=40$]
	% {\includegraphics[width=0.24\textwidth, trim=23 5 23 20]{fig/40_True_gru.png}}
	% \subfigure[$n=60$]
	% {\includegraphics[width=0.24\textwidth, trim=23 5 23 20]{fig/60_True_gru.png}}
	\subfigure[$n=80$]
	{\includegraphics[width=0.32\linewidth]{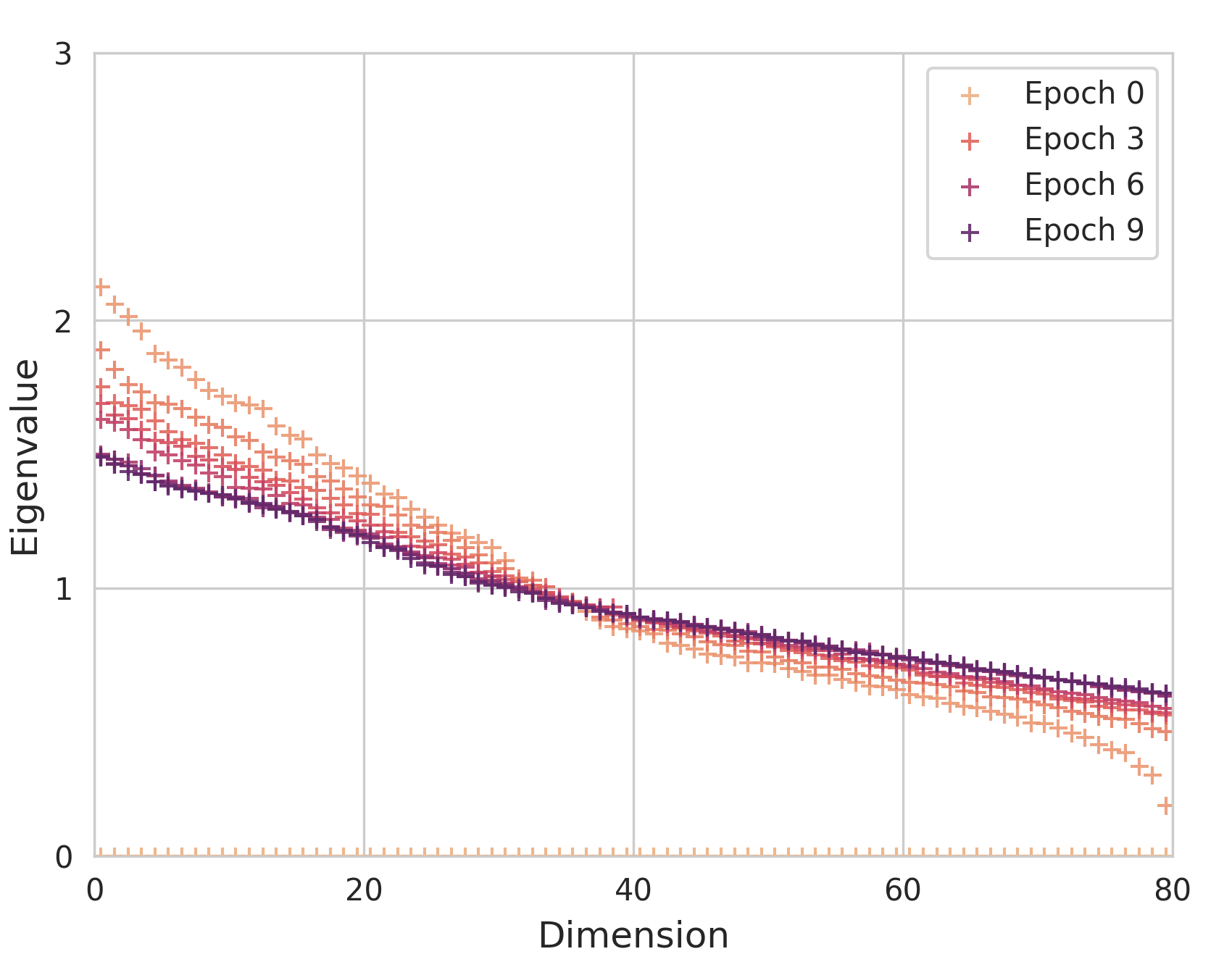}}
	\subfigure[$n=100$]
	{\includegraphics[width=0.32\linewidth]{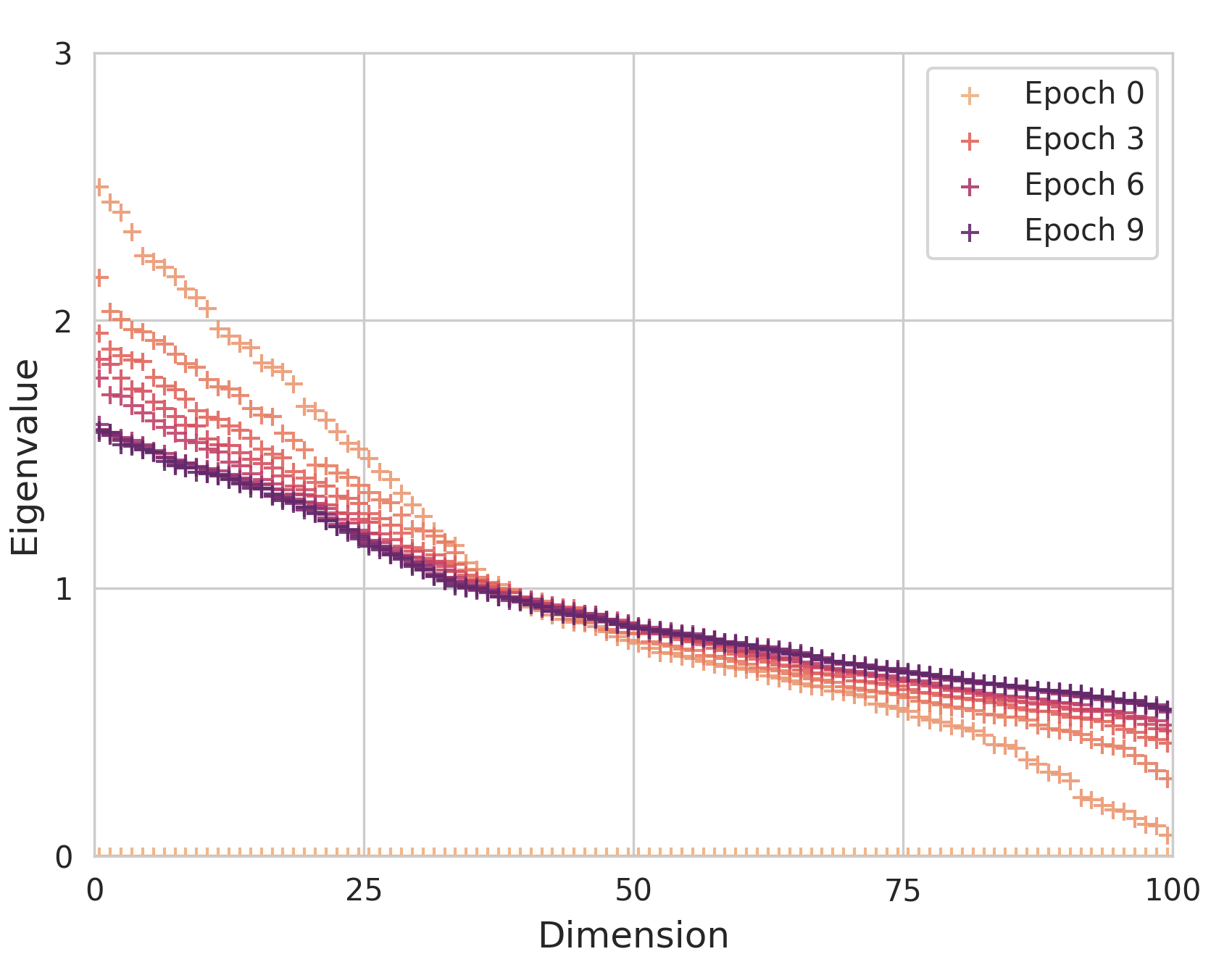}}
	\subfigure[$n=120$] 
	{\includegraphics[width=0.32\linewidth]{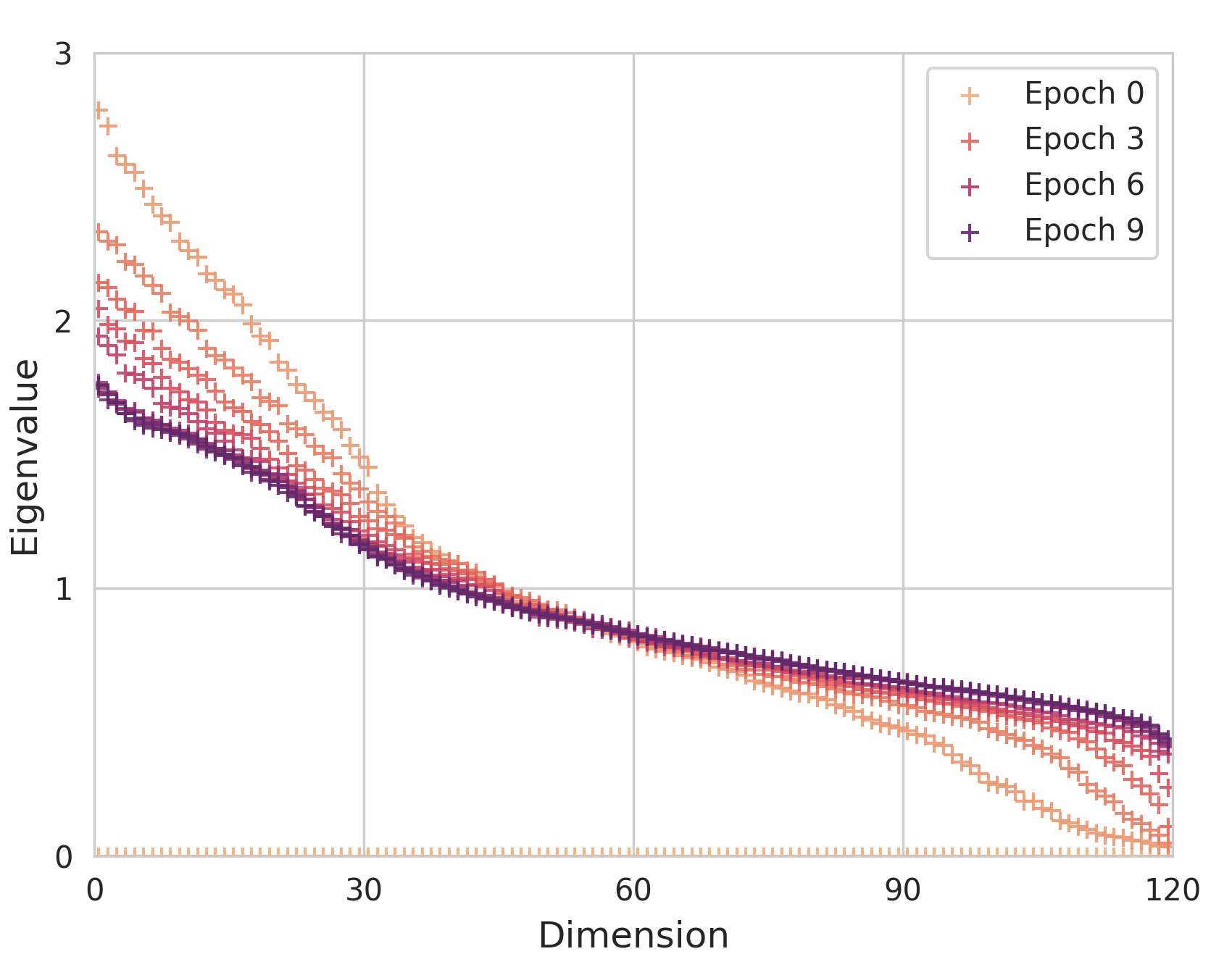}}
	\subfigure[$n=140$]
	{\includegraphics[width=0.32\linewidth]{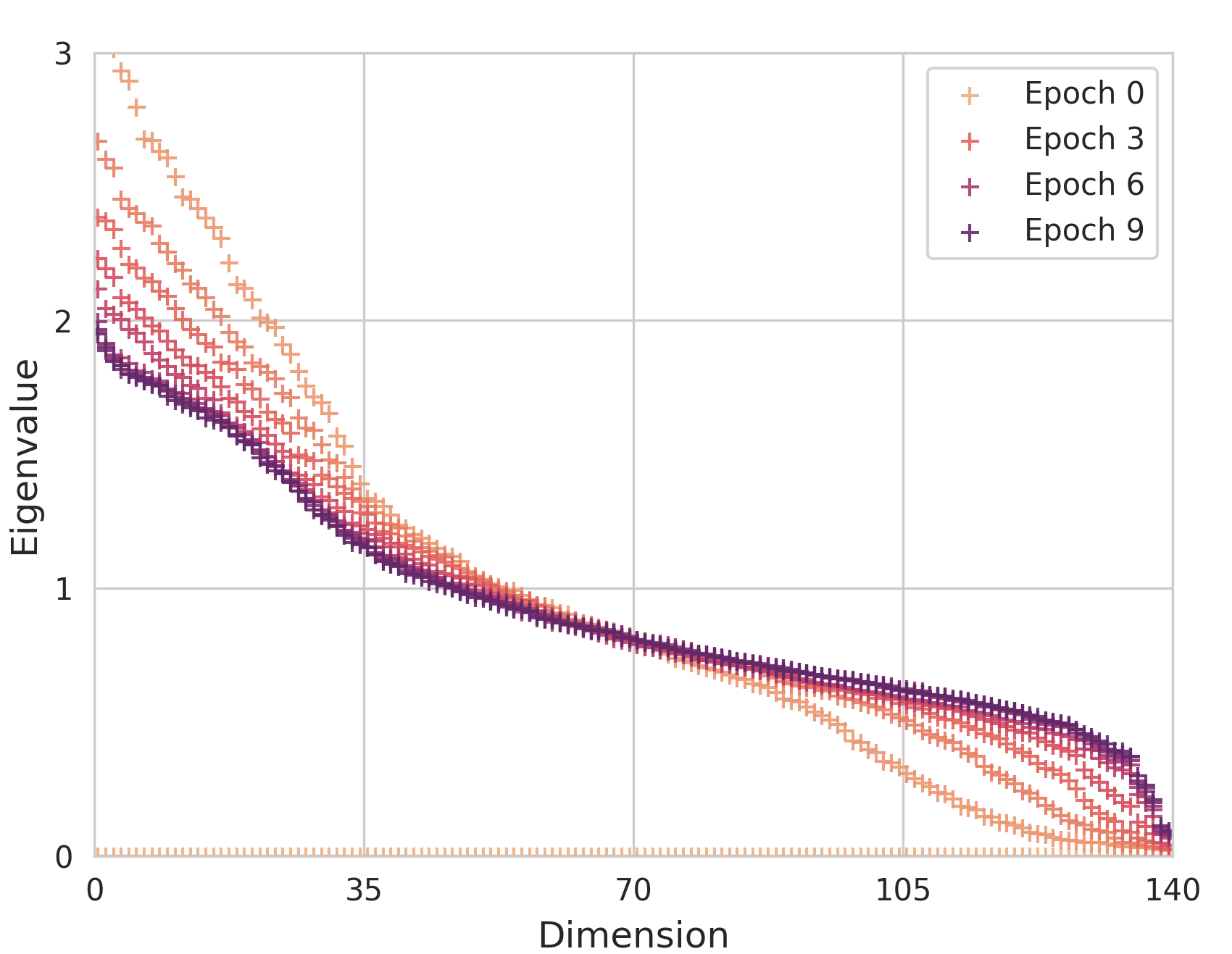}}
	\subfigure[$n=160$]
	{\includegraphics[width=0.32\linewidth]{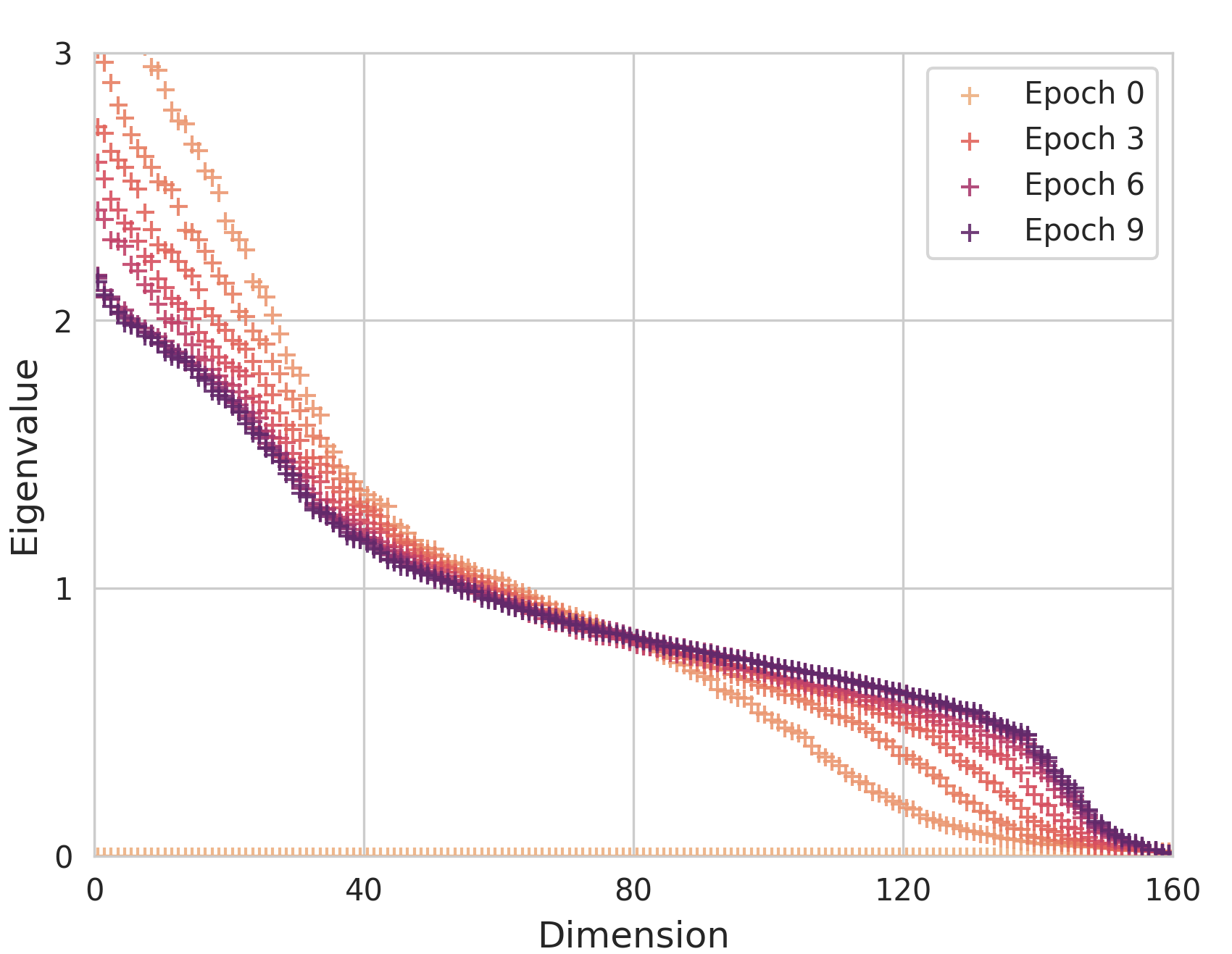}}
	\subfigure[$n=180$]
	{\includegraphics[width=0.32\linewidth]{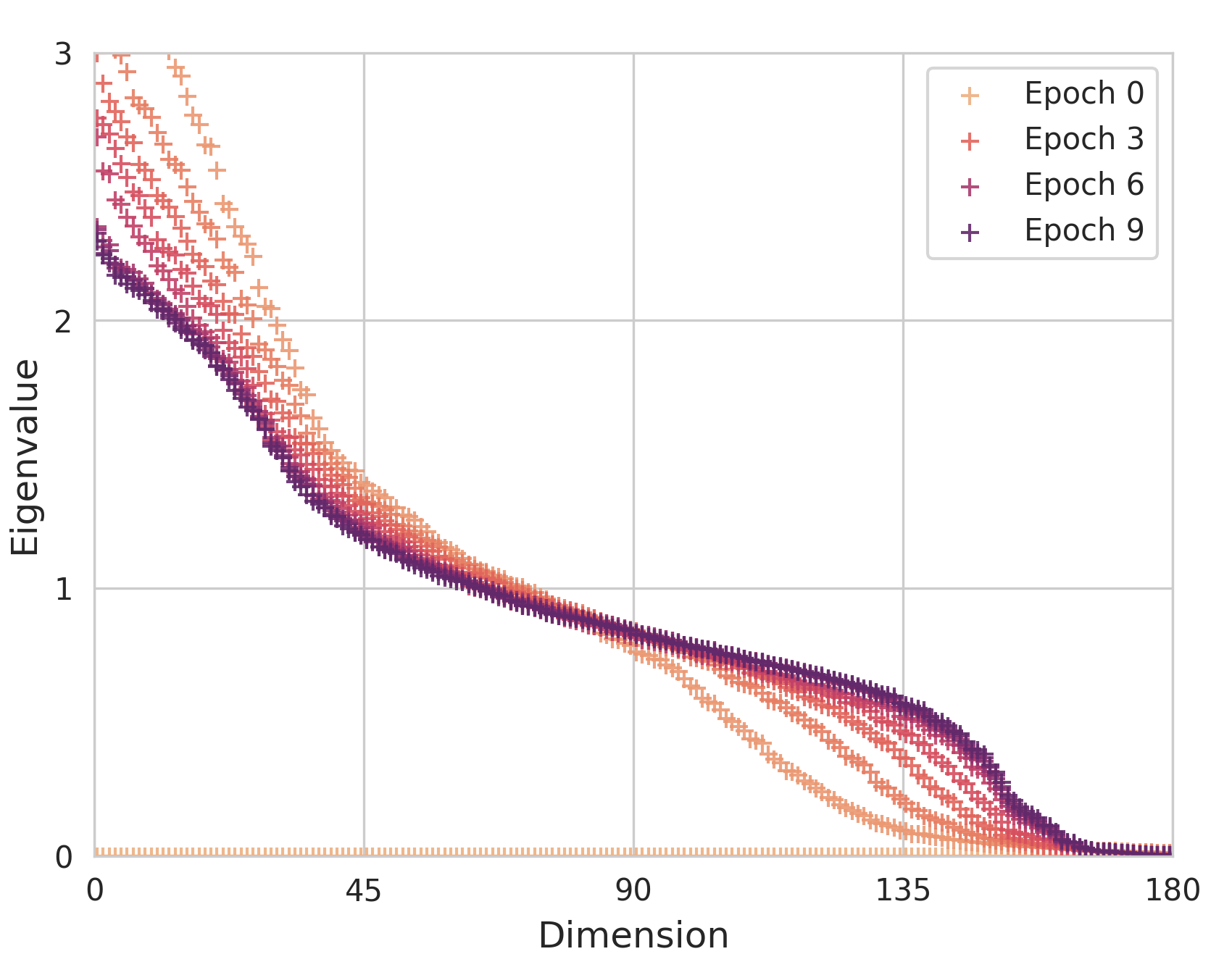}}
	\caption{\textbf{GRU}. The sorted eigenvalues of $\mathrm{cov}(\bm{u_i}-\bm{u_j},\bm{u_i}-\bm{u_j}), (i\neq j)$ 
		are plotted for different choices of the embedding dimension $n$ for the GRU embedding network. 
		When $n$ is small, the eigenvalues are distributed around $1$, as in (a)--(c). 
		Increasing the embedding dimension $n$, the sorted eigenvalues decrease to $0$ after some dimension, as in (d)--(f).
	}
	\label{fig:esd-gru}
\end{figure}
\begin{figure}[htb!]
	\centering
	\subfigure[AE$_g$]
	{\includegraphics[width=0.47\linewidth]{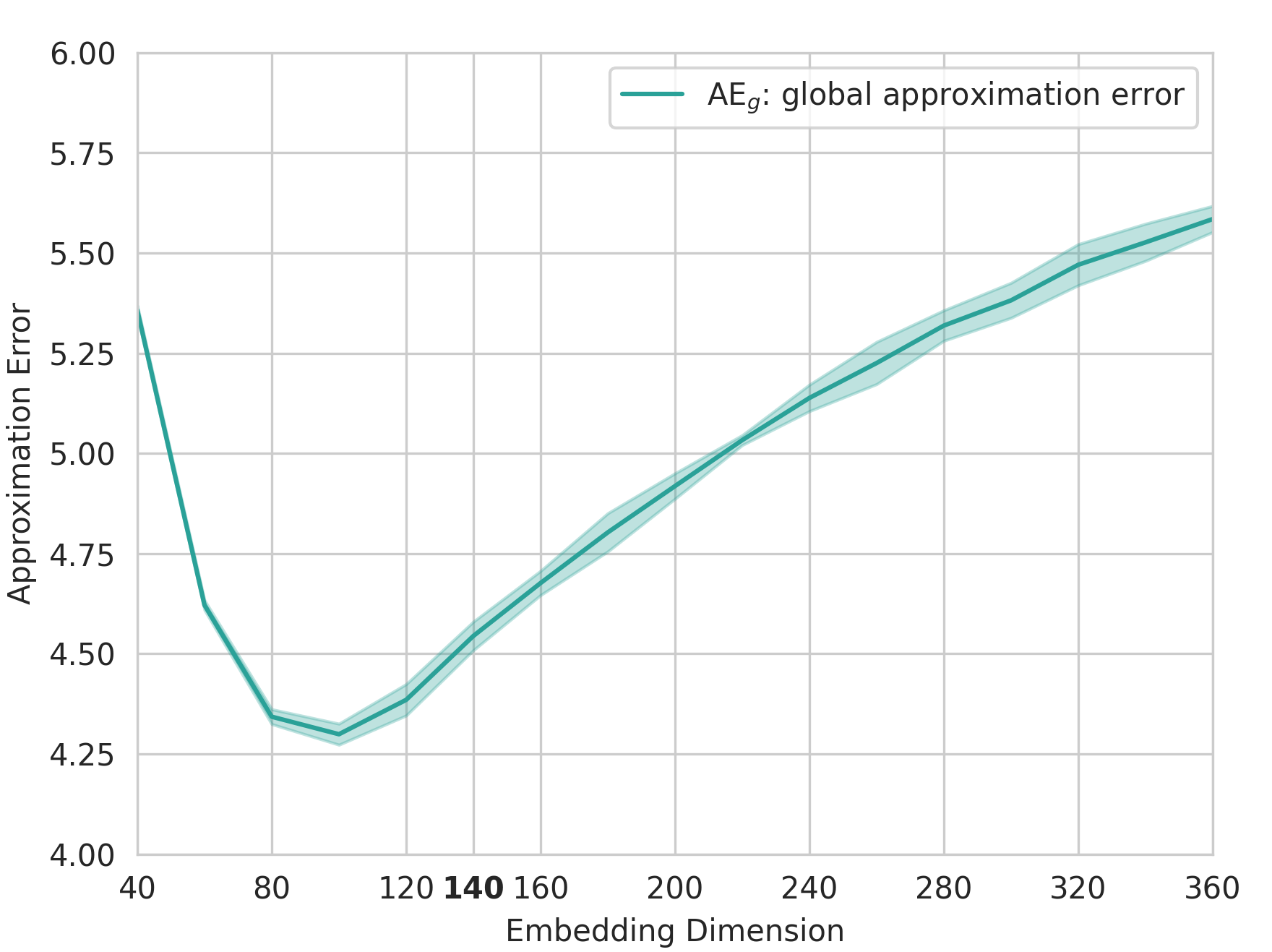}}
	\subfigure[AE$_h$]
	{\includegraphics[width=0.47\linewidth]{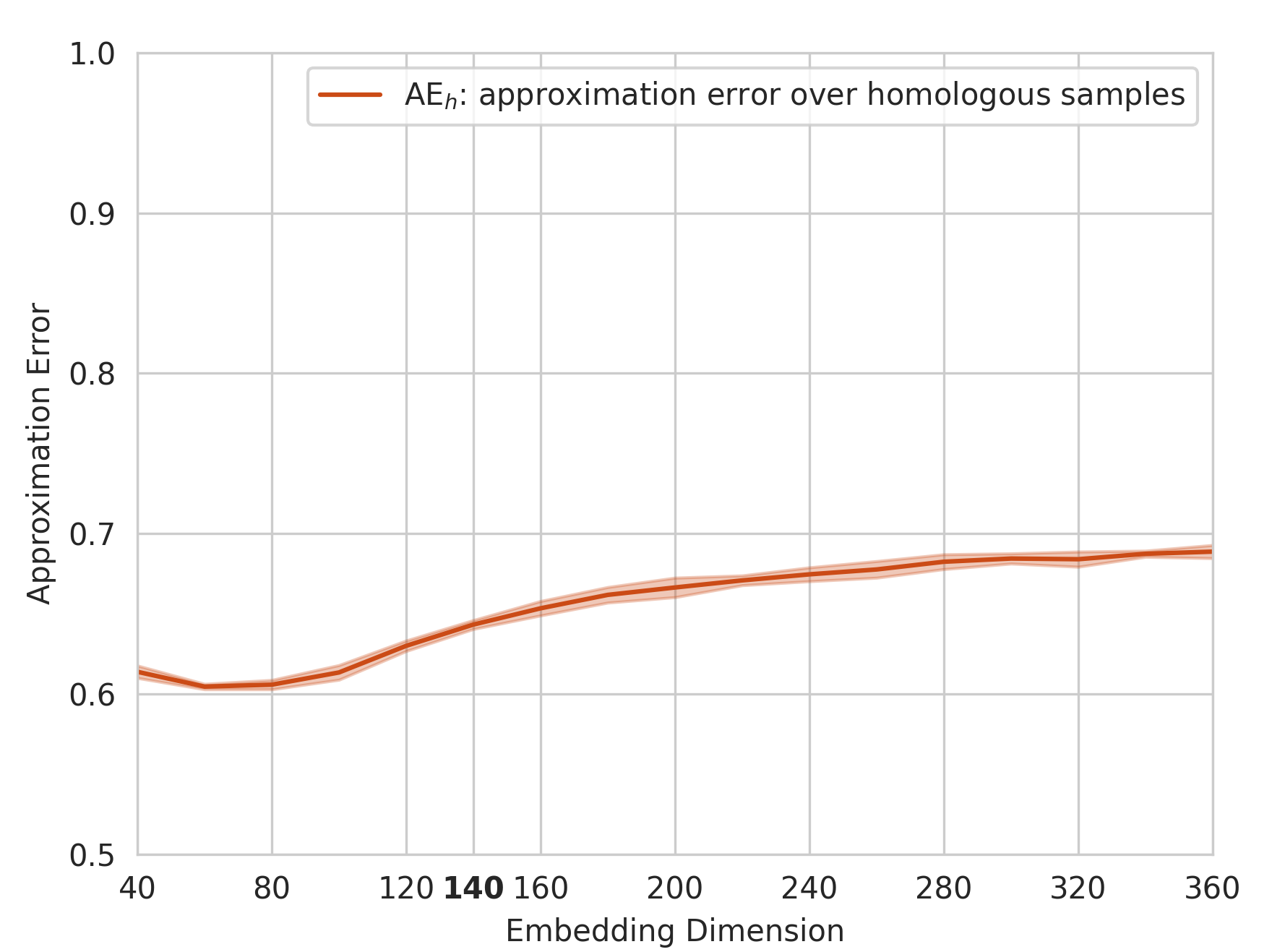}}
	\caption{\textbf{GRU}. 
		The AE$_g$ and AE$_h$ are shown against the embedding dimension in (a) and (b), respectively. 
	The curves are plotted based on the mean and standard deviation over 5 runs. 
		The global approximation error reaches its minimal around embedding dimension $n=100$ which is less than the ESD $n_0=140$. 
		Meanwhile the homologous approximation error is relative stable with the embedding dimension. }
	\label{fig:esd-performance-gru}
\end{figure}

\textbf{GRU}. As shown in \cref{fig:esd-gru}, the estimated ESD for the GRU embedding network is approximately $n_0=140$. 
However, unlike the sharp decline observed in the sorted eigenvalues of the convolutional networks, 
the sorted eigenvalues produced by the GRU embedding network exhibit a gradual decrease towards $0$. 
This observation suggests that the assumption stated in \ref{ass:bound} may not hold for the GRU embedding network. 
Furthermore, \cref{fig:esd-performance-gru} demonstrates that the global approximation error reaches 
its minimum point before reaching the ESD $n_0=140$, 
while the homologous approximation error remains 
relatively stable with respect to the embedding dimension. 
It is also shown that the GRU does not provide a competitive performance, 
as in \cref{fig:esd-performance-gru} and \cref{tab:results}.
This suggests that the GRU may not be an optimal choice of embedding network. 

When comparing the results of CNN-$5$, CNN-$10$, CNN-$5$-w, and CNN-$10$-w, as shown in \cref{fig:esd,fig:esd-cnn10,fig:esd-fatcnn5,fig:esd-fatcnn10}, 
it is observed that increasing the number of layers in the embedding network has little impact on enlarging the ESD, while 
enlarging the convolutional channels lead to a larger ESD. 
This indicates that a wider embedding network has a greater capacity to capture and express features from its input
in the Levenshtein distance embedding job. 

\section{A glimpse of failed sequences}
\begin{figure}[htb]
	\begin{center}
    \subfigure
	{\includegraphics[width=0.47\linewidth]{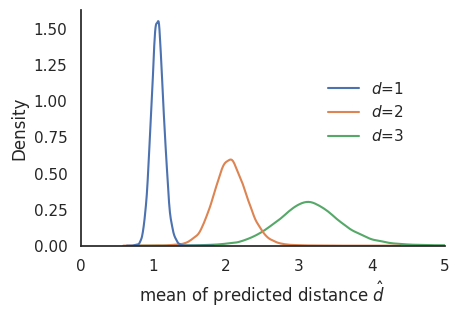}\label{fig:outlier1}}
    \subfigure
	{\includegraphics[width=0.47\linewidth]{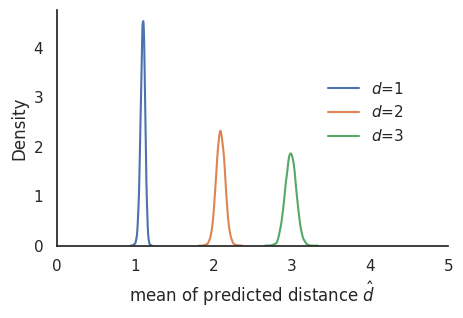}\label{fig:outlier2}}
	\caption{Distribution of $\mathrm{mean}_d(\bm{s})$. 
	(a) the distribution is plotted using the DNA-storage dataset; 
    (b) the distribution is plotted using a random dataset of sequences with length $7$.}
	\end{center}
\end{figure}

Certain kinds of sequences with low cardinality may pose challenges to the model, either due to their inherent property or 
imbalanced training samples. 
To investigate if there are sequences that consistently predict biased distances, 
we calculated the averaged predicted distance for each sequence $\bm{s}$ with a variable sequence $\bm{t}$ as 
$\mathrm{mean}_d(\bm{s}) = \mathrm{mean} \{\hat{d}(\bm{s},\bm{t})|\, d(\bm{s},\bm{t})=d\}$, 
and plotted the distribution of $\mathrm{mean}_d(\bm{s})$ over $\bm{s}$ for $d=1,2,3$, 
as shown in \cref{fig:outlier1}. 
In this figure, we found no evidence of such outlier sequences. 
We also printed and examined sequence pairs $(\bm{s},\bm{t})$ leading to failed approximations.
However, due to the length of the DNA sequences (around $150$ nt) engaged in the experiments, 
identifying clear patterns is challenging. 
To address this, we conducted experiments using shorter random sequences.
The corresponding distribution of $\mathrm{mean}_d(\bm{s})$ on a random dataset of sequence length $7$ 
is also shown in \cref{fig:outlier2}, 
which similarly indicates no evidence of outlier sequences. 
Focusing on sequence pairs that give outlier approximations with Levenshtein distance $1$, we observed a pattern, 
specifically, when insertion or deletion occurs at the beginning of a sequence with an extended homopolymer run (repeated letters), 
the predicted distance tends to be less accurate, as demonstrated in 
\cref{tab:outlier}. 
Intuitively, we speculate that advanced network architectures, positional encoding,
and bidirectional padding may help alleviate this issue. 

\begin{table}[htb]
    \begin{tabular}{cc|cc}
    \toprule
    seq. pair       & $\hat{d}$             & seq. pair       & $\hat{d}$             \\
    \midrule
    \textbf{2} 0 \textsl{3 3 3 3} 1 4 & \multirow{2}{*}{1.54}   & \textbf{2} \textsl{3 3 3 3 3 3} 4 & \multirow{2}{*}{1.52} \\
    0 \textsl{3 3 3 3} 1 4 4 &                                  & \textsl{3 3 3 3 3 3} 4 4 &                       \\
    \midrule
    1 1 \textbf{3} 1 \textsl{3 3 3} 4 & \multirow{2}{*}{1.51}   & \textsl{0 0 0} 3 0 3 3 4 & \multirow{2}{*}{1.60} \\
    1 1 1 \textsl{3 3 3} 4 4 &                                  & \textbf{0} \textsl{0 0 0} 3 3 4 4 &                       \\
    \midrule
    \textbf{2} 1 \textsl{3 3 3 3 3} 4 & \multirow{2}{*}{1.54}   & 0 \textsl{2 2 2 2 2 2} 4 & \multirow{2}{*}{1.55} \\
    1 \textsl{3 3 3 3 3} 4 4 &                                  & \textbf{1} 0 \textsl{2 2 2 2 2 2} &                       \\
    \midrule
    \textbf{0} 2 \textsl{0 0 0 0} 2 4 & \multirow{2}{*}{1.52}   & \textbf{2} 1 \textsl{3 3 3 3} 2 4 & \multirow{2}{*}{1.55} \\
    2 \textsl{0 0 0 0} 2 4 4 &                                  & 1 \textsl{3 3 3 3} 2 4 4 &                       \\
    \midrule
    \textbf{2} 0 \textsl{3 3 3 3} 0 4 & \multirow{2}{*}{1.57}   & \textbf{2} 0 \textsl{2 2 2 2 2} 4 & \multirow{2}{*}{1.52} \\
    0 \textsl{3 3 3 3} 0 4 4 &                                  & 0 \textsl{2 2 2 2 2} 4 4 &                       \\
    \midrule
    \textbf{0} 1 \textsl{0 0 0 0} 1 4 & \multirow{2}{*}{1.56}   & 1 \textsl{3 3 3 3} 0 0 4 & \multirow{2}{*}{1.52} \\
    1 \textsl{0 0 0 0} 1 4 4 &                                  & \textbf{2} 1 \textsl{3 3 3 3} 0 0 &                       \\
    \midrule
    \textbf{2} 3 \textsl{1 1 1 1 1} 4 & \multirow{2}{*}{1.54} &                 & \multirow{2}{*}{ } \\
    3 \textsl{1 1 1 1 1} 4 4 &                       &                 &                       \\
    \bottomrule
    \end{tabular}
    \centering
    \caption{Outlier pairs with Levenshtein distance $1$ from a random dataset of sequences with length $7$. 
    The boldface letters indicate the inserted or deleted letters; the italic letters show the extended homopolymer runs; 
    the letter `4' is for the padding letter. }\label{tab:outlier}
\end{table}

\section{Assumption verification}
Primarily concerned assumptions are the assumptions: A1, A2, and A4. 
Among these, the A4 is a revision of A2. 
In \cref{fig:esd,fig:esd-cnn10,fig:esd-fatcnn5,fig:esd-fatcnn10}, the validation of A4 is confirmed. 
When the embedding dimension is below the early stop dimension $n_0$, the covariance matrices have eigenvalues close to $1$, 
which indicates the embedding elements tend to be independent; 
when the embedding dimension exceeds $n_0$, the sorted eigenvalues exhibit a sharp decline to $0$. 
This phenomenon verifies the A4 to some extent. 

\begin{figure}[htb!]
	\centering
	\subfigure[$n=40$]
	{\includegraphics[width=0.19\linewidth]{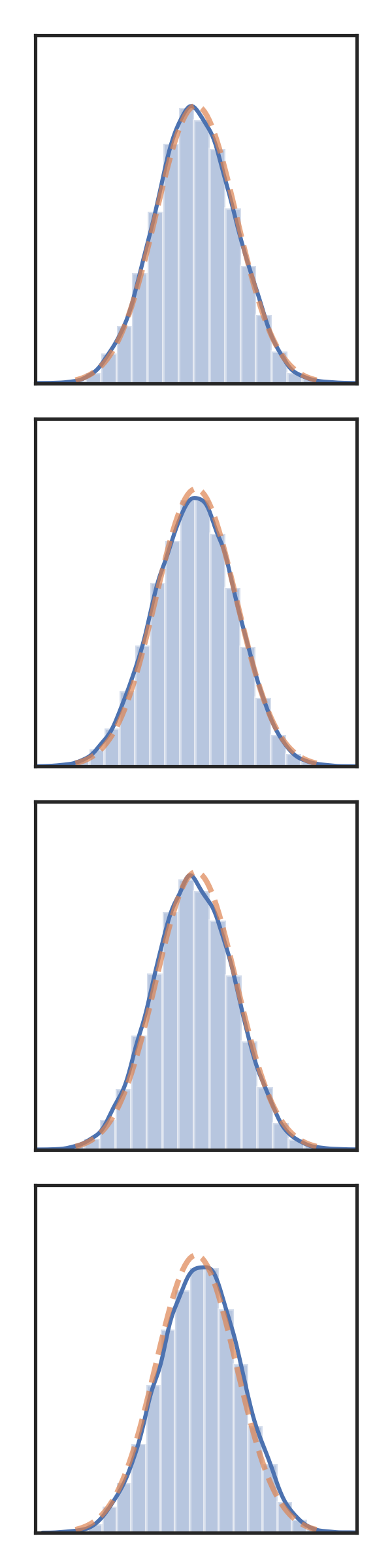}}
	\subfigure[$n=80$]
	{\includegraphics[width=0.19\linewidth]{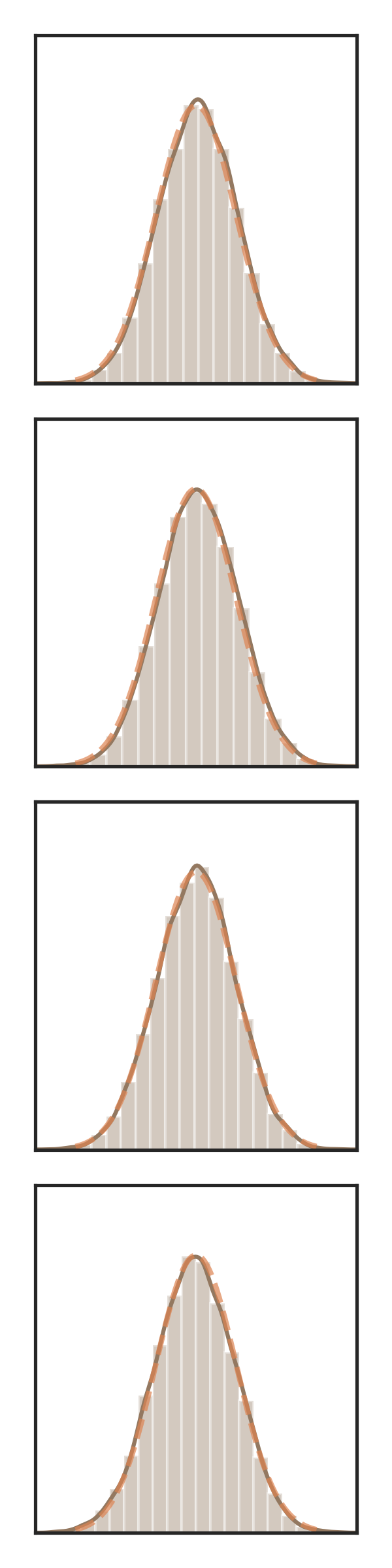}}
    \subfigure[$n=120$]
	{\includegraphics[width=0.19\linewidth]{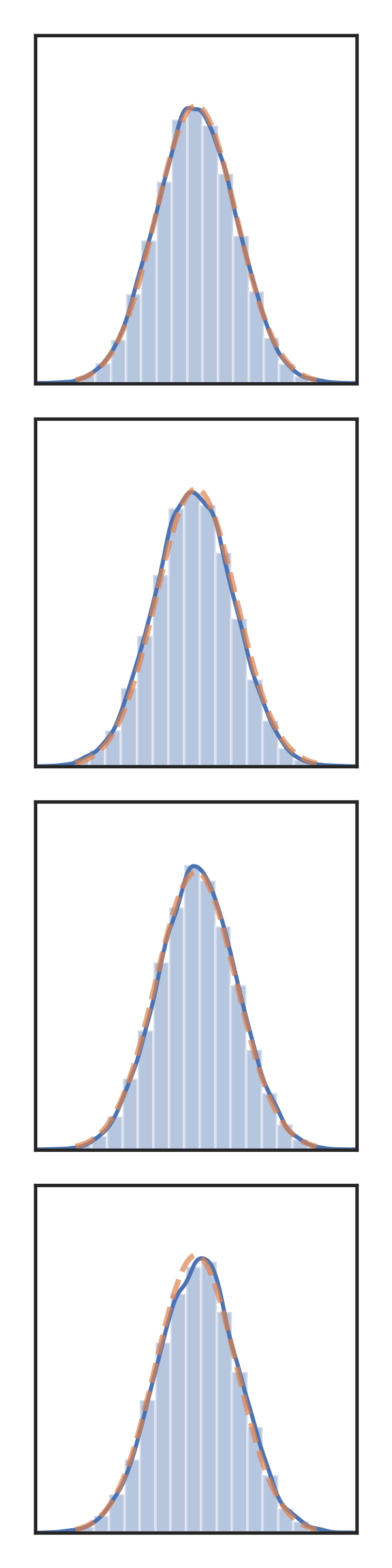}}
    \subfigure[$n=160$]
	{\includegraphics[width=0.19\linewidth]{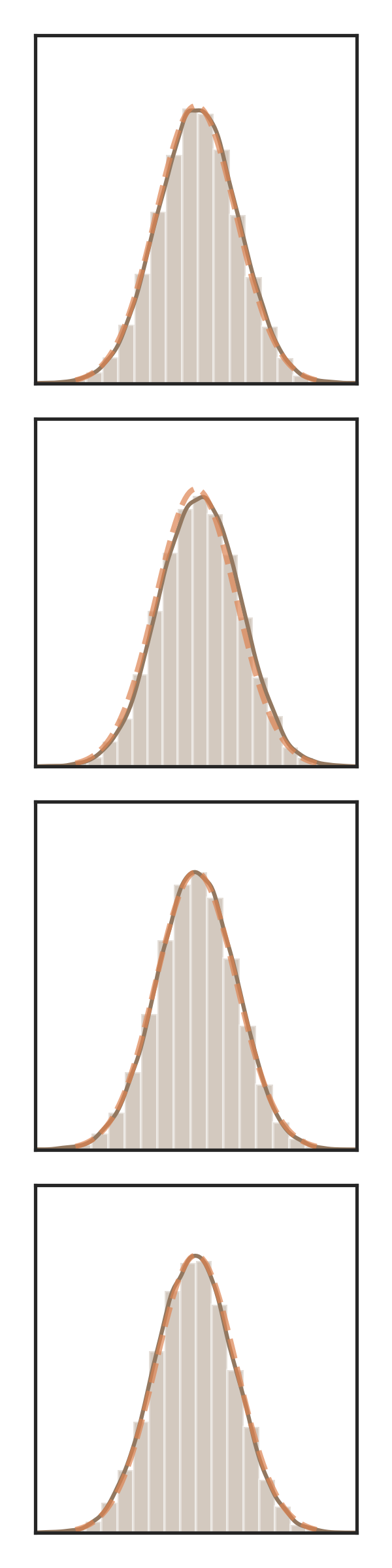}}
    \subfigure[$n=200$]
	{\includegraphics[width=0.19\linewidth]{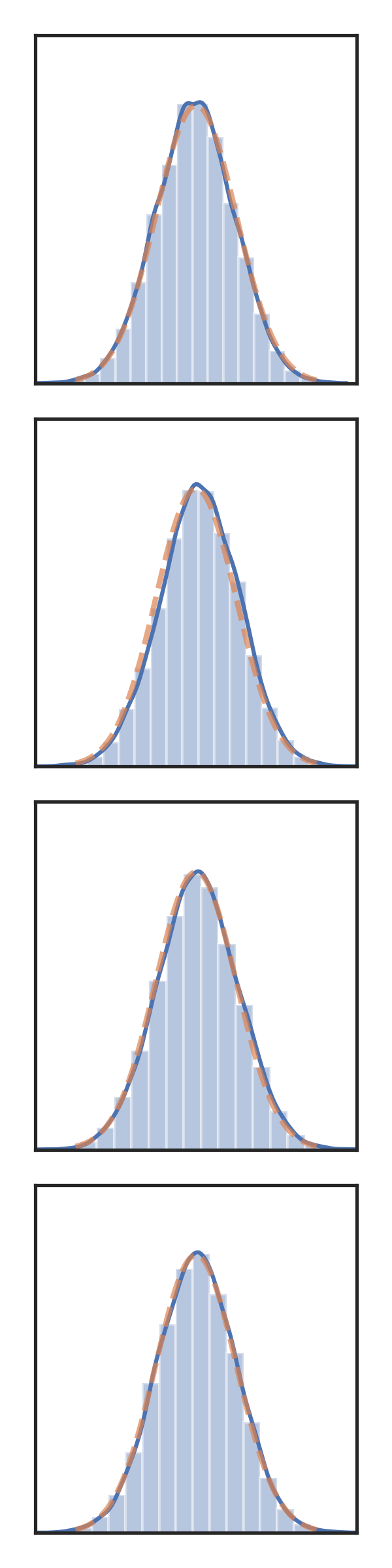}}
	\caption{The distribution of the first $4$ embedding elements by CNN-$10$-w. 
    (a)-(e) correspond to different choices of embedding dimension from $n=40$ to $n=200$. }\label{fig:element-distribution}
\end{figure}

To show whether A1 holds when the embedding dimension varies, we plotted the statistical distributions of 
the first several embedding elements, as illustrated in \cref{fig:element-distribution}. 
This figure suggested that the embedding dimension has minimal impact on the normality of embedding elements $\bm{u}_i$.
\end{document}